\documentclass[10pt,twocolumn,letterpaper]{article}

\usepackage{cvpr}              

\usepackage[dvipsnames]{xcolor}

\def\paperID{272} 
\def\confName{3DV\xspace}
\def\confYear{2024\xspace}

\definecolor{cvprblue}{rgb}{0.21,0.49,0.74}
\usepackage[pagebackref,breaklinks,colorlinks,citecolor=cvprblue]{hyperref}
\usepackage{times}
\usepackage{epsfig}
\usepackage{graphicx}
\usepackage{amsmath}
\usepackage{amssymb}
\usepackage{booktabs}
\usepackage{multicol}
\usepackage{colortbl} 
\usepackage[dvipsnames]{xcolor}
\newcommand*\rot{\rotatebox{60}}
\newcommand*\ang{\rotatebox{50}}

\usepackage{textcomp}
\usepackage{stfloats}
\usepackage{url}
\usepackage{verbatim}
\usepackage{float}
\usepackage{algorithm}
\usepackage{lipsum}
\usepackage{dirtytalk}
\usepackage{adjustbox}
\usepackage{multirow}
\usepackage{pifont}

\title{A Benchmark Grocery Dataset of Realworld Point Clouds From Single View}

\author{Shivanand Venkanna Sheshappanavar, Tejas Anvekar, Shivanand Kundargi, \\
Yufan Wang and Chandra Kambhamettu\\
Video/Image Modeling and Synthesis (VIMS) Lab., Dept. of Computer and Information Sciences\\
University of Delaware,
Newark, DE, USA 19716\\
{\tt\small \{ssheshap, tanvekar, leff, chandrak\}@udel.edu}
}

\begin{document}
\maketitle
\begin{abstract}
Fine-grained grocery object recognition is an important computer vision problem with broad applications in automatic checkout, in-store robotic navigation, and assistive technologies for the visually impaired. Existing datasets on groceries are mainly 2D images. Models trained on these datasets are limited to learning features from the regular 2D grids. While portable 3D sensors such as Kinect were commonly available for mobile phones, sensors such as LiDAR and TrueDepth, have recently been integrated into mobile phones. Despite the availability of mobile 3D sensors, there are currently no dedicated real-world large-scale benchmark 3D datasets for grocery. In addition, existing 3D datasets lack fine-grained grocery categories and have limited training samples. Furthermore, collecting data by going around the object versus the traditional photo capture makes data collection cumbersome. Thus, we introduce a large-scale grocery dataset called 3DGrocery100. It constitutes 100 classes, with a total of 87,898 3D point clouds created from 10,755 RGB-D single-view images. We benchmark our dataset on six recent state-of-the-art 3D point cloud classification models. Additionally, we also benchmark the dataset on few-shot and continual learning point cloud classification tasks. Project Page: \url{https://bigdatavision.org/3DGrocery100/}.

\end{abstract}    
\section{Introduction}
\label{sec:intro}

\begin{figure}[!h]
\centering
\includegraphics[width=0.9\linewidth]{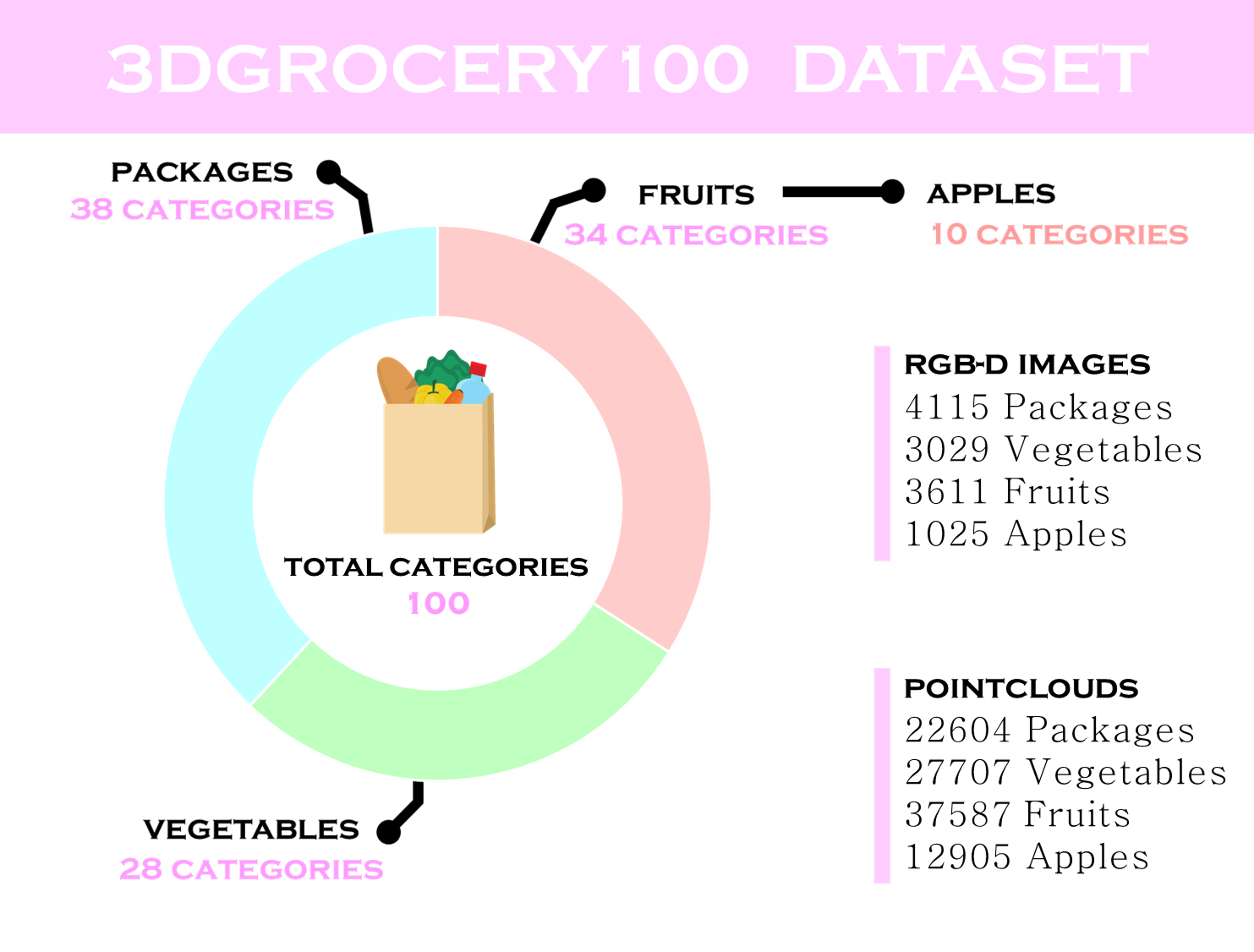}
\caption{3DGrocery100 Dataset Statistics. The dataset constitutes 10,755 RGB-D images and 87,898 point clouds spread across 100 classes. At a high level, the groceries are categorized into Fruits (10 apple and 24 non-apple classes), Vegetables (28), and Packages (38). Note: Apples are a subset of Fruits. Non-apples fruit RGB-D image count: 2,586; point cloud count: 24,682.}
\label{Fig1}
\end{figure}

3D Computer vision is an active research area with broad applications in autonomous navigation~\cite{sales20123d}, healthcare~\cite{singh20203d}, and augmented/virtual reality~\cite {krichenbauer2017augmented}. Among these applications, autonomous robotic navigation~\cite{aggarwal2015data} and assistance to the visually impaired~\cite{georgiadis2019computer,massiceti2021orbit} require recognizing and localizing objects in real-world scenarios. Grocery recognition~\cite{geng2018fine,leo2021systematic} in stores is a challenging problem. Real-world grocery recognition helps recognize misplaced products, identify product depletion~\cite{ruiz2019aim3s}, automatic grocery checkout, and assistive technologies for the visually impaired for a comfortable grocery experience. Developing robust grocery recognition systems would demand larger datasets to effectively train and deploy deep learning models. 

The past two decades have witnessed real-world grocery datasets growth (see Table~\ref{table1}) with varying sample sizes and image resolutions for retail product recognition. 2D representations inherently suffer from the loss of shape information and mainly cater to the needs of deep learning models~\cite{he2016deep} that rely on 2D images. Whereas with the availability of shape information in our grocery dataset, intrinsic characteristics of grocery objects would be captured which aids the recognition applications. Furthermore, unlike the 2D image-based counterpart, 3D point cloud-based recognition systems~\cite{yu2022lfpnet} suffer due to the lack of large-sized benchmark datasets further limiting our ability to build, evaluate, and compare the strengths of different methods, especially the recent data-hungry deep learning techniques. 
\newcommand{\xmark}{\ding{55}}%
\newcommand{\cmark}{\ding{51}}%

\begin{table*}[!ht]
\caption{Details of grocery datasets. OW - Open World, S - Studio, Acc - Accuracy, AP - Average Precision (\textit{italicized}). ``-" not available.} \label{table1}
\centering
\begin{adjustbox}{width=1.\linewidth}
\begin{tabular}{r|ccccccl}
\hline
\multirow{2}{*}{\textbf{Dataset}}&\textbf{Classes}&\textbf{Samples}&\textbf{Train}&\textbf{Test}&\textbf{OW}&\textbf{Acc/\textit{AP}}&\textbf{Model/Paper}\\ \cline{2-8}
& \multicolumn{7}{c}{2D Datasets}\\ \hline
\textbf{SOIL-47}~\cite{koubaroulis2002evaluating}(2002)&47&987&-&-& \xmark &71.0&MNS~\cite{matas2000colour}\\
\textbf{GroZi-120}~\cite{merler2007recognizing}(2007)&120&11194&676&11194&\xmark&18&SIFT~\cite{lowe2004distinctive}\\
\textbf{Supermarket}~\cite{rocha2010automatic}(2010)&15&2633&-&-&\cmark&98&SVM-Fusion~\cite{rocha2010automatic}\\
\textbf{GroZi-3.2K}~\cite{george2014recognizing}(2014)&80&9101&8421&680&\cmark&\textit{23.49}&Exemplar-based MLIC~\cite{george2014recognizing}\\
\textbf{Grocery}~\cite{varol2015toward}(2015)&10&354&-&-&\xmark&92.3&SVM~\cite{varol2015toward}\\
\textbf{Freiburg}~\cite{jund2016freiburg}(2016)&25&4947&4000&1000&\cmark&78.9&CaffeNet\cite{jia2014caffe}\\
\textbf{MVTec D2S}~\cite{follmann2018mvtec}(2018)&60& 21000& 4380 & 13020 &\xmark&\textit{79.9}&Mask R-CNN~\cite{he2017mask}\\
\textbf{Grocery Store}~\cite{klasson2019hierarchical}(2019)&81&5125&2640&2485&\cmark&84.0&DenseNet-169~\cite{huang2017densely}\\
\textbf{RPC}~\cite{wei2019rpc}(2019)& 200& \textbf{83739}&-&-&\xmark&56.68&Syn~\cite{hu2016detecting,krahenbuhl2011efficient}+Render~\cite{zhu2017unpaired}\\
\textbf{Magdeburg}~\cite{filax2019data} (2019)&942& 65315 &\textbf{23360}&\textbf{41955}&\xmark&91.83&VGG-16~\cite{simonyan2014very}\\
\textbf{TGFS}~\cite{hao2019take}(2019)&24&30000&22815&15212&\cmark&65.5&FCIOD~\cite{hao2019take}\\
\textbf{SKU-110K}~\cite{goldman2019precise}(2019)&\textbf{110712}&11762&8233+588&2941&\cmark&\textit{49.2}&Deep IoU Detection~\cite{goldman2019precise}\\
\textbf{RP2K}~\cite{peng2020rp2k}(2021)&2388&10385&-&-&\cmark&\textbf{95.18}&ResNet-34~\cite{he2016deep}\\ \hline
& \multicolumn{7}{c}{3D Datasets}\\ \hline
\textbf{BigBird}~\cite{singh2014bigbird}(2014)& 100 & 600 &-&-&\xmark&-&-\\
\textbf{HOPE}~\cite{tyree20226}(2022)&28&238+914&-&-&Toy& -& -\\
\textbf{Object-Verse}~\cite{deitke2023objaverse}(2023)&  \textbf{21,000} & \textbf{818,000} & \textbf{-} & \textbf{-} & \xmark & \textit{28.3} & GOL+3DCP~\cite{deitke2023objaverse}\\  
\textbf{3DGrocery63 (Ours)}&  63 & 87898 & 66032 & 21866 & \cmark & - & -\\
\textbf{3DGrocery (Ours)}&  100 & 87898 & 66032 & 21866 & \cmark & \textbf{50.50} & LocalFeatures~\cite{sheshappanavar2023local, sheshappanavar2023learning}\\ \hline
\end{tabular}\end{adjustbox}
\end{table*}

3D representations such as RGB-D and point clouds are richer in geometric information that closely represents the real world. The recent development of 3D deep learning models~\cite{qi2017pointnet,qi2017pointnet++,wang2019dynamic,guo2021pct,ma2022rethinking,PointNeXt} exploits the representational power of 3D data. However, these models are often trained and evaluated on limited datasets. Typically a CAD-based synthetic dataset, i.e., ModelNet40~\cite{wu20153d} with 12,311 point cloud samples spread across 40 classes, and a real-world dataset with variants, i.e., ScanObjectNN~\cite{uy2019revisiting} with 14,510 samples spread across 15 classes. Most recently, very large 3D datasets such as Objaverse~\cite{deitke2023objaverse} and Omni3D~\cite{brazil2023omni3d} are released, but they either artist-curated or do not contain sufficient classes of grocery, respectively.


Technological innovations over the past decade in 3D cameras/sensors, such as Light Detection And Ranging (LiDAR\cite{li2020lidar}), TrueDepth\cite{rajan2003simultaneous}, Kinect\cite{6190806}, etc., have increased the availability and accessibility of 3D sensors resulting in improved data acquisition. Additionally, the availability of high computational devices for 3D data processing has paved the way for many real-time 3D applications solving complex 3D problems. Furthermore, the availability of 3D sensors in handheld devices has opened new avenues of research that can address high-impact problems. However, despite the availability of 3D sensors in mobile devices, real-world 3D datasets are scarce. Notably, as listed in Table~\ref{table1}, the availability of 3D datasets for grocery recognition is limited. Collecting 3D data using mobile devices requires going around the object and scanning it to create a 3D point cloud. It is often not practical in a grocery store to go around objects as they are placed on racks/shelves. These issues encourage us to explore and capture the single-view RGB and Depth images which can be processed into point cloud instances of groceries (dataset statistics in Figure~\ref{Fig1}). Though these point clouds are incomplete, they carry geometric details of the grocery objects. Single-view RGB-D image acquisition is more straightforward and close to real-world scenarios mimicking photo capturing but provides both objects' geometric structures and point colors. 3D point clouds are a set of XYZ coordinates in 3D space with properties of permutation invariance.  

From the advent of PointNet~\cite{qi2017pointnet} to the most recent PointNeXt~\cite{PointNeXt}, a few hundred novel networks have been proposed for point cloud classification. Deep learning models inspired by these methods process raw point clouds and are among the popular choices for deployment among self-driving applications~\cite{fernandes2021point}. The performance of deep learning models depends heavily on the quality and quantity of the datasets used in training them. However, as stated in PatchAugment~\cite{sheshappanavar2021patchaugment}, most of the available 3D datasets are synthetic. Only a handful of real-world 3D point cloud datasets~\cite{uy2019revisiting} are available for researchers to train deep learning models for real applications. Furthermore, real-world benchmark 3D datasets, especially those collected using mobile devices, are limited. We present the largest 3D point cloud grocery dataset from RGB-D images collected using mobile phones (iPhone 12 Pro and Pro Max). 

Fresh produce grocery items, though packaged, are subject to pricing errors and usually require touching/picking by customers, which is discouraged during pandemics. Large items, such as watermelons, pose additional challenges for barcode scanning due to their size. These issues complicate grocery recognition tasks, highlighting the need for no-contact recognition methods. Additionally, the random orientation of produce and often hidden barcode labels further challenge effective recognition. Besides, image-based object recognition~\cite{fernandcz2017image} relies heavily on texture, color, and appearance cues. 2D images miss 3D geometric details, making 3D datasets essential for utilizing the 3D features of groceries in classification. In this regard, we present the 3DGrocery100 point cloud dataset obtained from RGB-D images as shown in Figure~\ref{Fig2}. Our large-scale 3D grocery dataset will enable the grocery recognition community to apply, develop, and adapt various deep learning techniques for 3D grocery classification. The key contributions of our paper are four-fold:
\begin{itemize}
    \item \textbf{3DGrocery100:} We present a novel benchmark 3D point cloud grocery dataset with 87,898 instances obtained from 10,755 RGB-D images across 100 categories. 
    \item \textbf{Classification Evaluation:} We benchmark our 3D dataset and its five subsets, each with two variants (color and no color), on six recent state-of-the-art (SOTA) models for 3D point cloud classification tasks. We also test and compare the recent SOTA models for robustness.
    \item \textbf{Few-shot Evaluation:} We merge similar shape classes to create a 63-class variant of the dataset, i.e., 3DGrocery63, for Few-shot evaluation and propose a strong baseline that evaluates the true generalization power of meta-learners on point cloud few-shot classification.
    \item \textbf{Class Incremental Learning Evaluation:}We set a 3D grocery benchmark using a class-incremental learning baseline. We extend 2D LWF~\cite{mccloskey1989catastrophic} with a dynamic multi-head classifier to benchmark on our 3D grocery dataset.
\end{itemize}

\section{Related Works}
\subsection{2D Grocery Datasets} 

Grocery datasets in the 2D domain (images) have been available since the mid-90s; veggievision~\cite{bolle1996veggievision} consists of 5000 images and 150 categories. Over the next decade, 2D grocery datasets~\cite{koubaroulis2002evaluating,merler2007recognizing} mainly improved the RGB images' resolution. The improvement in computing hardware during the past 15 years has paved the way to larger datasets~\cite{rocha2010automatic,george2014recognizing,jund2016freiburg,follmann2018mvtec,klasson2019hierarchical,wei2019rpc,filax2019data,hao2019take,goldman2019precise,peng2020rp2k}, gradually increasing the number of categories and images per category. The Supermarket Produce~\cite{rocha2010automatic} dataset features varied product view angles but uses a white canvas background, unlike real grocery store settings. Our dataset captures products in actual store environments, preserving the natural variations in view angles, elevations, and distances to objects.

The most recent dataset, RP2K~\cite{peng2020rp2k}, offers 2388 product categories, but it is limited to an average of only 37 RGB images per class. Like earlier datasets with challenges from specular reflections in packaged products, our images also feature classes with reflective packaging, complicating 3D analysis as discussed later. Recently, ~\cite{georgiadis2019computer} carefully considered the navigation context and divided the dataset into four abstraction levels, i.e., product, shelf, trail, and others. Our dataset supports the first two abstraction levels of grocery recognition. The comprehensive review work~\cite{santra2019comprehensive} on retail grocery categorization while listing the absence of RGB-D images for groceries encourages the exploration of RGB-D to address critical problems in grocery recognition. 

\subsection{3D Grocery Datasets}
While there are several 2D datasets in the grocery domain, the 3D grocery datasets among the representations such as RGB-D, point cloud, mesh, and voxels are rare~\cite{tyree20226,lai2011large,singh2014bigbird}. Among these datasets, though~\cite{lai2011large}, are RGB-D images of objects, it is not entirely a grocery dataset but only contains a few classes that belong to the grocery. For example, the HOPE~\cite{tyree20226} dataset provides the RGB-D video recordings of 28 categories of toy grocery objects in different lighting conditions. However, these are toy grocery items and do not overlap with any of the existing real-world brands. Further, the toy grocery items are placed in a non-store setting limiting the models trained in demonstrating in a real-store scenario.  BigBird~\cite{singh2014bigbird} provides 3D point clouds along with RGB-D images but is a small dataset with only 600 images. 

\subsection{3D Point Cloud Classification}
Point cloud analysis has gained traction since the pioneering work of PointNet~\cite{qi2017pointnet} that directly processes raw points through Multi-Layer Perceptrons (MLPs). However, PointNet loses valuable local geometric information while aggregating global features using max-pooling. Overcoming this limitation, PointNet++~\cite{qi2017pointnet++} captures local neighborhoods to learn local semantic information. DGCNN~\cite{ wang2019dynamic} introduced EdgeConv to capture local geometry by generating edge features to distinguish a point from its neighbors. 

PCT~\cite{guo2021pct} learns features through the attention mechanism of Transformers~\cite{vaswani2017attention}. PointMLP~\cite{ma2022rethinking} employs a simple feed-forward residual MLP network aggregating hierarchically extracted local features. PointNeXt~\cite{PointNeXt} improves PointNet++~\cite{qi2017pointnet++} by using better training strategies and an inverted residual bottleneck design with separable MLPs. These methods use  ModelNet40~\cite{wu20153d} (synthetic) and ScanObjectNN~\cite{uy2019revisiting} (real-world) datasets with 40 and 15 classes, respectively. Despite groceries being 3D objects, there are no benchmark 3D grocery datasets for training point-based deep learning models. Our dataset enables the training of models to classify groceries in real-world stores.

\subsection{Point Cloud Few Shot Learning}
Recent progress in Few-Shot Learning (FSL) for 2D image processing has led to two main approaches: metric-based methods, which improve class discrimination through feature space, and optimization-based methods, which focus on model adaptability to new classes. One such metric-based approach, Prototypical Net~\cite{snell2017prototypical}, introduces class-prototype, which is the mean of the support features of each class. Meanwhile, optimization-based FSL techniques aim for quick adaptation to new tasks through minimal gradient updates, with ongoing research enhancing 2D FSL. Meta-baseline~\cite{chen2021meta} introduces a cosine metric classifier with learnable weights, demonstrating improved performance compared to the Squared Euclidean Distance. SimpleTrans~\cite{luo2022channel} proposes a straightforward transform function to adjust the weights of different channels, mitigating the channel bias problem in FSL. In this paper, we follow GPr-Net~\cite{gpr_tejas} for point-cloud FSL using ProtoNet and observe the grocery dataset as a strong generalization benchmark.

\subsection{Point Cloud Class-Incremental Learning}
Many essential robotic perception applications heavily rely on real-world data. However, real-world data is dynamic and arrives in a continuous stream. Incremental learning strategies have been developed to learn from this incoming data effectively. These strategies can be broadly classified into Task-Incremental~\cite{taskil1,taskil2,taskil3,tasksil4}, Domain-Incremental~\cite{dil1,dil2}, and Class-Incremental learning (CIL)~\cite{cl1,cl2,cl3}. CIL closely emulates the dynamic arrival of data in real-world scenarios. Baseline approaches in CIL like~\cite{mccloskey1989catastrophic,ewc} address the problem of catastrophic forgetting in neural networks, which is essential when dealing with incremental real-world data. \cite{chowdhury2021learning} explore learning without forgetting on-point clouds using semantic representation. Advancements in depth sensor technology have spurred growth in 3D and 2.5D data applications. CIL tackles the challenge of processing novel streaming data without task identifiers at test time, reflecting real-world conditions. This makes incremental learning on new data increasingly important.


\begin{figure*}[ht]
\centering
\includegraphics[width=0.84\linewidth]{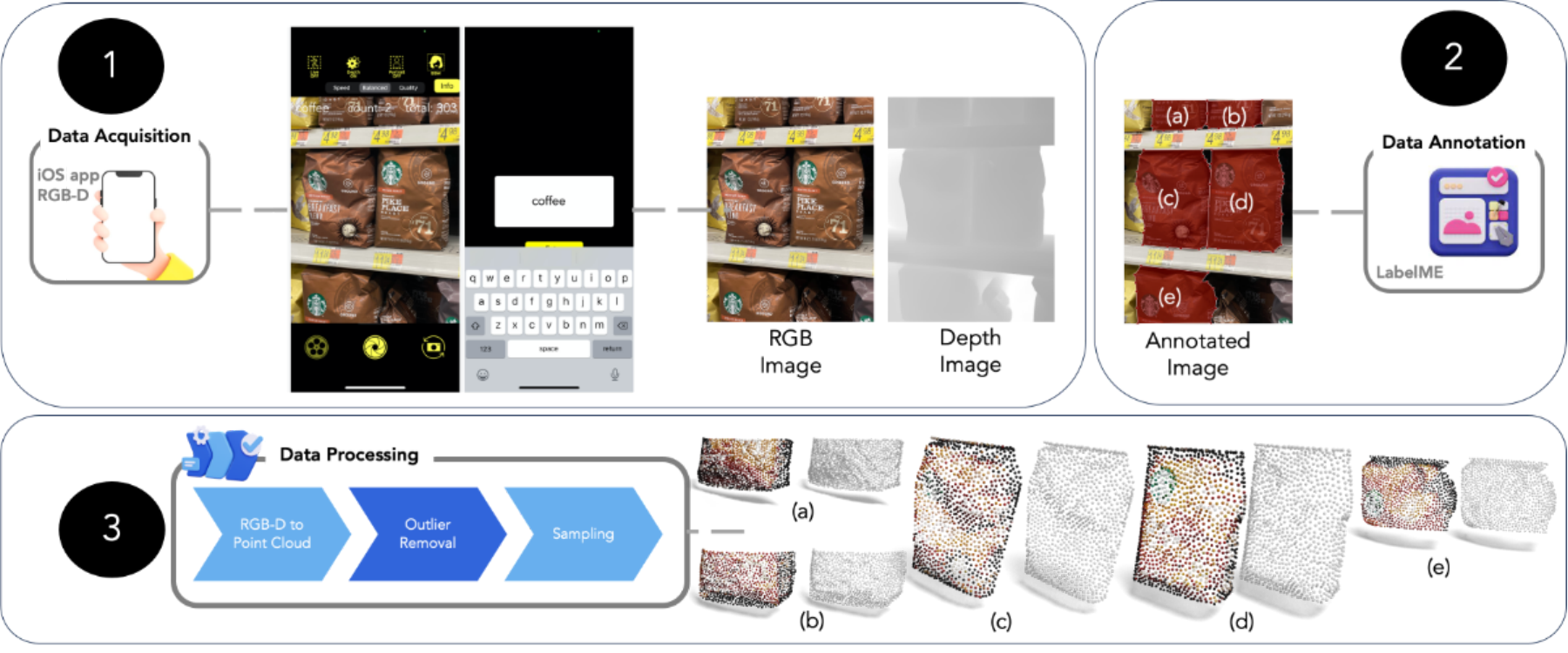}
\caption{\textbf{(1)} iOS app to capture an RGB image and a Depth Image (darker regions nearer to the camera) for a class \textbf{(2)} annotations on the RGB image - coffee instances numbered [a-e], \textbf{(3)} [a-e] 1024 Farthest point sampled 3D points of 5 coffee objects with and without colors.}
\label{Fig2}
\end{figure*}

\section{Dataset creation}

This section outlines our dataset creation process, as depicted in Figure~\ref{Fig2}, highlighting three main steps.
\subsection{Data collection} 

Data was collected over four months from 18 different local grocery stores. Based on the availability of items in the stores, the collection spanned from a few days to weeks. Grocery stores are well-lit environments with similar arrangements for each class. For example, most packages are placed on shelves on both sides of an aisle, while a few vegetables and fruits are arranged on a table.

\subsubsection{iOS app for collection} Recently, Apple iPhones have offered multiple ways to capture 3D objects. The latest iPhones Pro and Pro Max (version numbers 12, 13, 14, and 15) come with a LiDAR sensor, a go-to sensor to capture the scene's depth in a range of up to 5 meters. The LiDAR camera gives a low resolution (192 $\times$ 256) depth map. The RGB image resolution is 3024 $\times$ 4032. In the case of LiDAR, the lower resolutions of depth maps result in sparse point clouds losing significant 3D information. Another way to capture depth is stereo vision using the built-in back-facing dual camera. The depth map and camera calibration metrics are saved as a 32-bit floating-point array. The resolution of these depth maps is 576 $\times$ 768. The point clouds generated with stereo depth preserve the local geometry of the object, making it look more realistic (refer to Figure~\ref{Fig8} and its description in the supplementary). The better quality of 3D point clouds from the depth captured using the stereo vision built-in back-facing dual camera~\cite{app} encouraged us to adopt the back-facing dual camera to capture depth data along with RGB images in our customized iOS app for data collection. 

\subsubsection{Data Hierarchy and RGB-D dataset}
\begin{figure}[!b]
\centering
\includegraphics[width=0.96\linewidth]{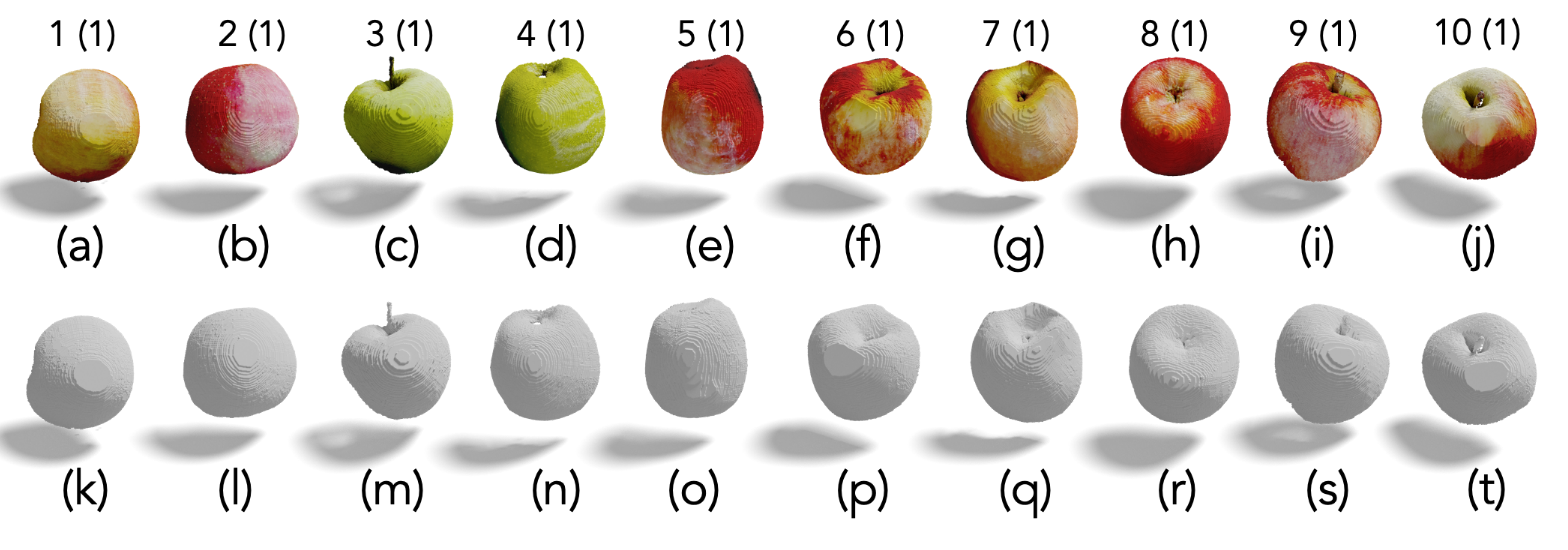}
\caption{(a-j) point clouds of apple-evercrisp, apple-fuji, apple-golden-delicious, apple-granny-smith, apple-honeycrisp, apple-pazazz, apple-pink-lady, apple-red-delicious, apple-royal-gala, and apple-wild-twist with color. (k-t) point clouds in the same order without color. Colors play a significant role in distinguishing objects from different classes. Without color, all ten classes of apples appear very similar in the 3D point cloud representation.}
\label{Fig3}
\end{figure}

Following~\cite{klasson2019hierarchical}, Figure~\ref{Fig1} shows the high-level hierarchy as three main categories of groceries, i.e., Fruits, Vegetables, and Packages, along with the count of 2D RGB-D images and 3D point clouds from them. The Fruits category is divided into ten classes of apples and 24 classes of non-apple fruits. Vegetables and Packages have 28 and 38 classes, respectively. We combine these categories to form the ``Full" dataset of 100 classes. Figures~\ref{Fig3} and~\ref{Fig4} show visual examples of classes from Apple10 and (Fruits, Vegetables, and Packages) respectively. We provide two variants of the point clouds i.e., with and without colors. These grocery items were positioned in racks (most packages and some vegetables), on tables (most fruits and a few vegetables), inside refrigerators (often packages), and in the natural environment of local grocery stores with similar lighting settings. The resolution of RGB images is 3024$\times$4032, and the resolution of the depth maps is 576$\times$768. During data collection, we used portrait mode, making the height of images larger than the width. In addition, our dataset also has challenging images, which have products with reflective surfaces, images taken at oblique views, and images with darker backgrounds.

\subsubsection{Challenges}

Data collection using mobile phones would often result in shaky and blurred images. Such images are harder to annotate, resulting in heavily distorted or noisy 3D point clouds. We have carefully discarded such images from our dataset. In some stores, a few grocery items, such as milk and eggs, are placed in open racks with sufficient cooling,  while other stores place these items in a refrigerator or a cooler. We have collected and annotated them as separate classes to enable recognition systems to learn the differences between such classes. Although the doors of these refrigerators are slightly reflective, we observed no depth distortion for the milk-in-cooler and eggs-in-cooler classes. However, the colors are slightly different compared to their non-refrigerated counterpart classes. We left a gap of at least 24 hours between the visits to each grocery store to capture variations in the stores' inventory, especially those of fruits and vegetables. A few items take longer to sell, and changes in their inventory often take more time. We planned data collection of such items with longer gaps.

\subsection{Data Annotation}

To annotate such a large dataset of 10755 grocery images, we started annotation in parallel while the data collection was still in progress. We used LabelMe~\cite{labelme} to annotate collected 2D RGB images. We drew a polygon for each object in the picture to approximate its boundary as shown in Figure~\ref{Fig2} (data annotation section). Since the annotated 2D RGB images and depth maps generate the 3D dataset, any extra pixel (due to human error) that does not belong to the object of interest would have a very different depth value leading to background noise or outliers. Therefore, we annotated the polygon carefully on or within the object boundary to avoid additional background noise. As objects in an image tend to get occluded or are partly out of the frame, we annotated objects whose visibility was significant, i.e., approximately at least 25\% visible.

\begin{figure*}[!ht]
\centering
\includegraphics[width=0.98\linewidth]{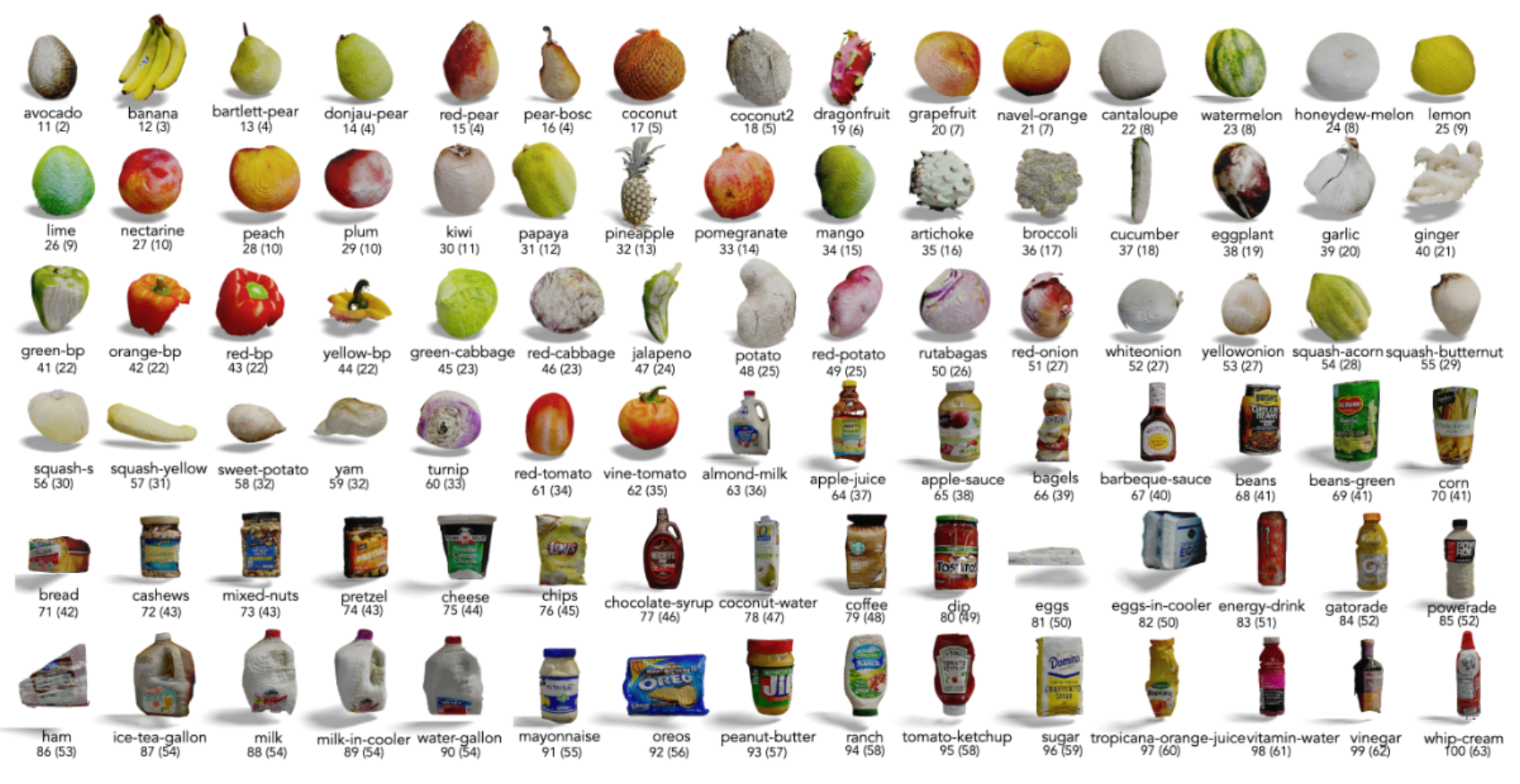}
\caption{3D point cloud representations of 24 (of the 34) classes of fruits (non-apples) with colors. The sets (bartlett-pear, donjau-pear, red-pear, pear-bosc), (cantaloupe, watermelon, honeydew-melon), (lemon, lime), (nectarine, peach, plum), and (grapefruit, navel-orange) are similar in shapes, and these classes get merged in case of 3DGrocery63. 28 classes of Vegetables with colors. 38 classes Packages with colors. *-bp is bell-pepper and *-s is spaghetti. The first number represents the class ID in 3DGrocery100, and the number in the parenthesis represents the class ID in 3DGrocery63 obtained after merging similar-shaped classes in 3DGrocery100.}
\label{Fig4}
\end{figure*}

\subsection{Data Processing}
We used Pinhole Camera Intrinsic Parameters to convert RGB-D images to point clouds using Open3D~\cite{zhou2018open3d} library. Processing RGB-D images for point cloud objects is challenging and produces outliers, especially at the object boundaries, due to minor annotation issues that overstep the background or neighboring objects. We compute the median of the point cloud and use it as a threshold to remove background points from the point cloud. The reflective and transparent surfaces of the grocery items often result in outlier depth values resulting in outlier 3D points. We use PointCleanNet~\cite{rakotosaona2020pointcleannet} for removing outliers from the point clouds (refer to Figure~\ref{Fig9} in supplementary). PointCleanNet cleans point cloud datasets in a two-step process: first, it finds and removes outliers, then denoises the remaining points by estimating a per-point displacement vector. We use their pre-trained outlier-removal model to process our RGB-D point cloud data. This deep learning approach requires less labor and does not make any assumptions about the noise model or object surface.  

We first separate XYZ coordinates from the XYZRGB data. Then, pass the XYZ data into the pre-trained PointCleanNet~\cite{rakotosaona2020pointcleannet}, to estimate the per-point probabilities to check if a point is an outlier. In our case, a relatively lower threshold of 0.4  yields better results when compared to PointCleanNet~\cite{rakotosaona2020pointcleannet}, which uses a threshold of 0.5. Finally, we combine the RGB information from the original data and XYZ coordinates based on the estimated probability values to produce ``cleaned" data in XYZRGB format. These XYZ values are normalized to the range [-1,1].

Even though we selectively annotated significantly visible objects, many of them ended up with only a few points in their 3D point cloud representations due to their size or placement within the image. We retained objects with at least 10,000 points and discarded the rest. After applying PointCleanNet~\cite{rakotosaona2020pointcleannet} for outlier removal, point clouds from each class are similar to those shown in Figures~\ref{Fig3} and~\ref{Fig4}. We used the farthest point sampling (FPS) method to sample 1024 points (with and without colors as in Figure~\ref{Fig2}) for our experiments. Visual examples of five samples per class with 1024 FPS points are shown in the supplementary. Table~\ref{tab2} shows the train and test split of all the subsets of our dataset. We divide the RGBD images into 75\% training and 25\% testing and use all the point clouds obtained from the 75-25 split as train and test point clouds, respectively.

\section{Applications}
Our dataset constitutes three categories, i.e., Fruits, Vegetables, and Packages with 34, 28, and 38 classes, respectively. The Fruits category is further divided into ten classes of apples (Apple10) and 24 non-apple classes. In this section, we analyze our experimental results on the complete dataset (Full) and four subsets (Apple10, Fruits, Vegetables, and Packages). Table~\ref{tab2} shows the number of images and the corresponding point clouds created from the images.

\subsection{Real-World Point Cloud Classification}
We consider six representative works on 3D object classification namely, PointNet~\cite{qi2017pointnet}, PointNet++\cite{qi2017pointnet++}, DGCNN\cite{wang2019dynamic}, PCT~\cite{guo2021pct}, PointMLP\cite{ma2022rethinking}, and PointNeXt\cite{PointNeXt} to benchmark our dataset. We retain the training parameters, such as learning rate, optimizer, number of epochs, batch size, and weight decay for each of these methods as per the original papers. While PointNet~\cite{qi2017pointnet} and PointNet++~\cite{qi2017pointnet} had shown poor performance on the real-world dataset ScanObjectNN~\cite{uy2019revisiting}, the most recent deep learning models PointMLP~\cite{ma2022rethinking} and  PointNeXt~\cite{PointNeXt} have achieved SOTA classification results. However, the hardest variant of ScanObjectNN~\cite{uy2019revisiting} dataset has less than 15,000 samples and 15 classes. Table~\ref{tab2} shows the overall accuracy on both variants of the four subsets, and the Full dataset.

\begin{table}[!ht]
\centering
\caption{3D point cloud classification results on 3DGrocery100. Params in M (million) and FLOPs in G (giga).}
\label{tab2}
\resizebox{\linewidth}{!}{%
\begin{tabular}{rccccccc}
\hline 
\textbf{Models} &
  \textbf{Apples} &
  \textbf{Fruits} &
  \textbf{Vegges} &
  \textbf{Packages} &
  \textbf{Full} &
  \textbf{\#Params} &
  \textbf{FLOPs} \\ \hline
\#Classes &
  \cellcolor[HTML]{EFEFEF}10 &
  \cellcolor[HTML]{EFEFEF}24 &
  \cellcolor[HTML]{EFEFEF}28 &
  \cellcolor[HTML]{EFEFEF}38 &
  \cellcolor[HTML]{EFEFEF}100 &
   &
   \\
\#Images &
  \cellcolor[HTML]{DCDCDC}1025 &
  \cellcolor[HTML]{DCDCDC}2586 &
  \cellcolor[HTML]{DCDCDC}3029 &
  \cellcolor[HTML]{DCDCDC}4115 &
  \cellcolor[HTML]{DCDCDC}10755 &
   &
   \\
Train &
  772 &
  1944 &
  2264 &
  3103 &
  8083 &
   &
   \\
Test &
  253 &
  642 &
  765 &
  1012 &
  2672 &
   &
   \\
\#Patches &
  \cellcolor[HTML]{EFEFEF}12905 &
  \cellcolor[HTML]{EFEFEF}24682 &
  \cellcolor[HTML]{EFEFEF}27707 &
  \cellcolor[HTML]{EFEFEF}22604 &
  \cellcolor[HTML]{EFEFEF}87898 &
   &
   \\
Train &
  9706 &
  18406 &
  20720 &
  17214 &
  66032 &
   &
   \\
Test &
  3199 &
  6276 &
  6987 &
  5390 &
  21866 &
   &
   \\ \hline
\multicolumn{8}{c}{\cellcolor[HTML]{DEEEFF}\textbf{INPUT: RGB Image}} \\
\textbf{Swin T~\cite{liu2021swin}} &
  85.10 &
  98.50 &
  98.40 &
  98.3 &
  98.5 &
   28.00 &
  04.50  \\
\multicolumn{1}{l}{} &
  \multicolumn{1}{l}{} &
  \multicolumn{1}{l}{} &
  \multicolumn{1}{l}{} &
  \multicolumn{1}{l}{} &
  \multicolumn{1}{l}{} &
  \multicolumn{1}{l}{} &
  \multicolumn{1}{l}{} \\
\multicolumn{8}{c}{\cellcolor[HTML]{DBFFC8}\textbf{INPUT: 1024 POINTS + COLORS}} \\
\textbf{PointNet~\cite{qi2017pointnet}} &
  78.35 &
  96.42 &
  97.00 &
  97.85 &
  92.62 &
  00.70 &
  00.31 \\
\textbf{PointNet++~\cite{qi2017pointnet++}} &
  84.00 &
  97.58 &
  \textbf{98.25} &
  \underline{98.46} &
  95.21 &
  01.48 &
  01.71 \\
\textbf{DGCNN~\cite{wang2019dynamic}} &
  83.59 &
  \textbf{98.01} &
  \underline{97.88} &
  \textbf{98.53} &
  \underline{96.20} &
  01.81 &
  05.39 \\
\textbf{PCT~\cite{guo2021pct}} &
  82.14 &
  \underline{97.89} &
  97.63 &
  97.97 &
  94.19 &
  02.88 &
  04.34 \\
\textbf{PointMLP~\cite{ma2022rethinking}} &
  \textbf{90.18} &
  97.75 &
  97.68 &
  98.38 &
  \textbf{96.82} &
  13.24 &
  31.35 \\
\textbf{PointNext~\cite{PointNeXt}} &
  \underline{84.81} &
  97.36 &
  97.37 &
  98.13 &
  94.44 &
  04.52 &
  06.49 \\
\multicolumn{1}{l}{} &
  \multicolumn{1}{l}{} &
  \multicolumn{1}{l}{} &
  \multicolumn{1}{l}{} &
  \multicolumn{1}{l}{} &
  \multicolumn{1}{l}{} &
  \multicolumn{1}{l}{} &
  \multicolumn{1}{l}{} \\
\multicolumn{8}{c}{\cellcolor[HTML]{E8E8E8}\textbf{INPUT: ONLY 1024 POINTS}} \\
\textbf{PointNet~\cite{qi2017pointnet}} &
  18.93 &
  33.16 &
  35.15 &
  71.61 &
  37.61 &
  00.70 &
  00.31 \\
\textbf{PointNet++~\cite{qi2017pointnet++}} &
  \textbf{26.20} &
  \textbf{47.98} &
  \textbf{54.11} &
  \textbf{82.70} &
  \textbf{48.65} &
  01.48 &
  01.71 \\
\textbf{DGCNN~\cite{wang2019dynamic}} &
  19.04 &
  37.01 &
  42.16 &
  79.44 &
  44.19 &
  01.81 &
  05.39 \\
\textbf{PCT~\cite{guo2021pct}} &
  18.95 &
  32.39 &
  43.42 &
  75.34 &
  39.67 &
  02.88 &
  04.34 \\
\textbf{PointMLP~\cite{ma2022rethinking}} &
  19.16 &
  36.38 &
  47.29 &
  79.94 &
  \underline{45.72} &
  13.24 &
  31.35 \\
\textbf{PointNext~\cite{PointNeXt}} &
  \underline{21.63} &
  \underline{40.58} &
  \underline{48.40} &
  \underline{81.35} &
  43.73 &
  04.52 &
  06.49 \\ \hline 
\end{tabular}%
}
\end{table}

\subsection{Benchmarking Real-World Scenarios with Minimal Data}
\label{sec:FSL}


Building upon insights from Meta-Dataset~\cite{triantafillou2019meta}, existing 3D few-shot classification benchmarks~\cite{sharma2020self,feng2022enrich, ye2022makes, ye2023closer} primarily emphasize intra-dataset generalization. However, our objective is to benchmark models for effective generalization across entirely new distributions, even unseen datasets. To address this challenge, we introduce cross-domain benchmarking strategies for Meta-learners, spotlighting their generalization capabilities on 3D datasets.

The significance of 3D Grocery arises from its unique attributes. Comprising real-world 2.5D point clouds, the proposed 3DGrocery dataset is meticulously refined to eliminate inter-class intersections across categories, yielding 63 distinct classes as illustrated in Figure \ref{Fig4}. This dataset remains distinct from well-known point cloud datasets like ShapeNet~\cite{wu20153d}, ModelNet40~\cite{wu20153d}, and ScanObjectNN~\cite{uy2019revisiting}, exhibiting no overlaps. The absence of overlapping classes and differentiation from other popular datasets positions it perfectly for robust generalization benchmarks for Meta-Learners. Through 3D Grocery, we comprehensively assess how well Meta-Learning models can adapt and generalize to novel and distinct data distributions, offering insights into their genuine resilience in real-world challenges.

\begin{table}[!h]\centering
  \caption{Quantitative analysis for Few-shot 3D point cloud classification on 3DGrocery63 dataset. All models are pre-trained on ModelNet40\cite{wu20153d}.}
  \begin{adjustbox}{width=\linewidth}
  \begin{tabular}{rc|cccc|cccc}
    \hline 
   \multirow{2}{*}{\textbf{Model}}&\multirow{2}{*}{\textbf{Weight Init}}&\multicolumn{4}{c|}{\textbf{5-ways}}&\multicolumn{4}{c}{\textbf{10-ways}}\\
   \cline{3-10} 
    &&\multicolumn{2}{c}{\textbf{10-shots}}&\multicolumn{2}{c|}{\textbf{20-shots}}&\multicolumn{2}{c}{\textbf{10-shots}}&\multicolumn{2}{c}{\textbf{20-shots}}\\
   \hline 
    \multirow{2}{*}{\textbf{PointNet}\cite{qi2017pointnet}}&
    Random&54.30&$\pm$09.51&59.40&$\pm$09.59&39.23&$\pm$05.67&43.95&$\pm$05.67\\
    &MN40&56.26&$\pm$09.81&61.58&$\pm$10.22&42.20&$\pm$05.97&46.29&$\pm$06.04\\
    
    \multirow{2}{*}{\textbf{PointNet++}\cite{qi2017pointnet++}}&
    Random&55.18&$\pm$09.37&60.80&$\pm$10.42&40.52&$\pm$05.78&45.22&$\pm$06.37\\
    &MN40&61.60&$\pm$09.23&68.76&$\pm$09.05&46.13&$\pm$04.99&52.04&$\pm$06.38\\
    
    \multirow{2}{*}{\textbf{DGCNN}\cite{wang2019dynamic}}&
    Random&55.20&$\pm$10.26&63.02&$\pm$09.43&41.41&$\pm$06.23&47.13&$\pm$06.38\\
    &MN40&\textbf{65.44}&$\pm$\textbf{10.64}&\textbf{72.18}&$\pm$\textbf{09.59}&\textbf{50.99}&$\pm$\textbf{05.89}&\textbf{57.30}&$\pm$\textbf{07.50}\\
    
    \multirow{2}{*}{\textbf{PointMLP}\cite{ma2022rethinking}}&
    Random&40.34&$\pm$06.17&47.14&$\pm$07.97&27.04&$\pm$04.35&30.95&$\pm$04.55\\
    &MN40&63.96&$\pm$11.75&69.18&$\pm$09.74&48.81&$\pm$06.57&54.35&$\pm$06.59\\

    \multirow{2}{*}{\textbf{PCT}\cite{guo2021pct}}&
    Random&57.36&$\pm$09.12&53.86&$\pm$08.73&44.23&$\pm$05.15&54.09&$\pm$06.23\\
    &MN40&62.45&$\pm$09.64&66.52&$\pm$09.55&44.39&$\pm$05.26&51.17&$\pm$05.92\\
    
    \multirow{2}{*}{\textbf{PointNeXt}\cite{PointNeXt}}&
    Random&55.82&$\pm$08.48&62.38&$\pm$09.46&41.73&$\pm$05.99&46.24&$\pm$06.20\\
    &MN40&60.64&$\pm$10.60&65.12&$\pm$09.46&46.04&$\pm$06.21&51.51&$\pm$05.96\\
    \hline
  \end{tabular}
  \end{adjustbox}
  \label{table3}
\end{table}

\begin{table*}[!ht]\centering
  \caption{Quantitative analysis for Meta-Learning on few-shot 3D point cloud classification. Avg Weak is the average of accuracies in \% from three splits in a weak generalization task with 15 novel classes (~\cite{uy2019revisiting}) and Strong is a strong generalization task with 63 novel classes.}
  \begin{adjustbox}{width=\linewidth}
  \begin{tabular}{r|cccccccc|cccccccc}
    \hline 
    &\multicolumn{8}{c|}{\textbf{Avg Weak}}&\multicolumn{8}{c}{\textbf{Strong}}\\
    \cline{2-17}
    \textbf{Models}&\multicolumn{4}{c|}{\textbf{5-ways}}&\multicolumn{4}{c|}{\textbf{10-ways}}&\multicolumn{4}{c|}{\textbf{5-ways}}&\multicolumn{4}{c}{\textbf{10-ways}}\\
    \cline{2-17}
    &\multicolumn{2}{c|}{\textbf{5-shots}}&\multicolumn{2}{c|}{\textbf{10-shots}}&\multicolumn{2}{c|}{\textbf{5-shots}}&\multicolumn{2}{c|}{\textbf{10-shots}}&\multicolumn{2}{c|}{\textbf{5-shots}}&\multicolumn{2}{c|}{\textbf{10-shots}}&\multicolumn{2}{c|}{\textbf{5-shots}}&\multicolumn{2}{c}{\textbf{10-shots}}\\
    \hline
    
   \textbf{PointNet}~\cite{qi2017pointnet}&53.50&$\pm$0.63&56.36&$\pm$0.60&36.40&$\pm$0.37&40.11&$\pm$0.37&27.38&$\pm$0.41&28.91&$\pm$0.43&14.99&$\pm$0.23&17.10&$\pm$0.24\\
   \textbf{PointNet++}~\cite{qi2017pointnet++}&50.16&$\pm$0.60&55.92&$\pm$0.62&37.03&$\pm$0.36&39.73&$\pm$0.36&24.11&$\pm$0.37&29.39&$\pm$0.43&15.64&$\pm$0.23&18.15&$\pm$0.25\\
   \textbf{DGCNN}~\cite{wang2019dynamic}&51.30&$\pm$0.62&56.47&$\pm$0.61&36.80&$\pm$0.37&40.08&$\pm$0.37&\textbf{28.39}&$\pm$\textbf{0.43}&\textbf{30.56}&$\pm$\textbf{0.43}&\textbf{17.35}&$\pm$\textbf{0.24}&18.24&$\pm$0.25\\
   \textbf{PCT}~\cite{guo2021pct}&57.05&$\pm$0.64&56.58&$\pm$0.64&39.66&$\pm$0.37&43.24&$\pm$0.37&27.57&$\pm$0.41&29.22&$\pm$0.43&16.58&$\pm$0.25&\textbf{18.59}&$\pm$\textbf{0.26}\\
   \textbf{PointMLP}~\cite{ma2022rethinking}&46.13&$\pm$0.62&50.14&$\pm$0.59&32.76&$\pm$0.35&40.45&$\pm$0.37&24.70&$\pm$0.38&25.52&$\pm$0.39&13.45&$\pm$0.20&14.76&$\pm$0.21\\
    \textbf{PointNeXt}~\cite{PointNeXt}&\textbf{58.63}&$\pm$\textbf{0.64}&\textbf{58.94}&$\pm$\textbf{0.65}&\textbf{41.70}&$\pm$\textbf{0.39}&\textbf{44.86}&$\pm$\textbf{0.38}&28.23&$\pm$0.45&28.95&$\pm$0.42&16.29&$\pm$0.23&17.81&$\pm$0.24\\
    
    \hline 
  \end{tabular}
  \end{adjustbox}
  \label{table4}
\end{table*}

\noindent \textbf{Experimental Setup:} We conducted two experiments \textit{Baseline Few shot Evaluation} and \textit{Few shot Meta-Learning} to demonstrate that our proposed dataset 3DGrocery63 can act as a very strong benchmarking dataset to perform weak vs. strong generalization tasks.
\noindent \textbf{(1) Baseline Few shot Evaluation:} We take pre-trained point cloud classifiers \cite{qi2017pointnet,qi2017pointnet++,wang2019dynamic,ma2022rethinking,guo2021pct} on ModelNet40\cite{wu20153d}, and perform $k$-way, $m$-shot few-shot evaluation using features learnt by the classifier as expressed by authors in~\cite{crosspoint}. The few-shot evaluation results on  ScanObjectNN~\cite{uy2019revisiting} are reported in Table~\ref{table13} in supplementary and similarly results on proposed 3DGrocery63 are reported in Table~\ref{table3}; comparing both the tables, it is evident that classifiers generalize well on ScanObjectNN dataset while failing on 3DGrocery63, the main reason for this phenomenon is a data-inductive bias that is common in ScanObjectNN.
\noindent \textbf{(2) Few shot Meta-Learning:} We further evaluate the aforementioned phenomenon by benchmarking on Few-shot Meta-Learning using ProtoNet~\cite{snell2017prototypical} for point-cloud FSL as described in GPR-Net~\cite{gpr_tejas}. We propose six data splits for this setup: Train, Val, and Test (weak1, weak2, weak3, and strong). For the train, we propose to use ShapeNet55~\cite{wu20153d} and exclude 15 categories that intersect with ScanObjectNN; the remaining are used for Val split. For weak1 $\to$ weak3 test splits, we use ScanobjectNN (ONLY OBJ, OBJ+BG, PB75) dataset and 3DGrocery63 for strong test split. We follow an episodic paradigm to train the ProtoNet version of 3D classifiers. The settings are given in the supplementary. Our findings are reported in Table~\ref{table4} and Table~\ref{table14} (in supplementary), which depicts the curse of data-inductive bias; every model generalizes well on weak generalization tasks but fails on proposed strong generalization.  Finally, we conclude that the proposed 3DGrocery63 is a strong baseline to validate the true generalization of meta-learners on point cloud few-shot classification considering data-inductive biases.

\subsection{Benchmarking Real-World Scenarios of Continual Data}

Our extensive grocery dataset has the largest collection of point clouds captured in the real world. As we anticipate the possibility of new classes being introduced to our real-world data, we are actively investigating CIL methods, leveraging which training from scratch on entire data can be avoided. These methods will serve as a benchmark for our data and enable us to effectively handle the inclusion of novel classes without forgetting the information of the past as our dataset continues to grow. We follow PointCLIMB~\cite{pointclimb_shiv} for developing, the problem setting of Point-cloud CIL on the proposed dataset.

\noindent \textbf{problem setting}: Class-Incremental learning problem $\mathcal{T}$ consists of sequence of $k$ tasks:
        \begin{equation}
            \label{eq:3defcil-setting}
            \mathcal{T} = [(C^{1},D^{1}), (C^{2},D^{2})\text{, ... ,}(C^{k},D^{k})]
        \end{equation}
        
where each task $\mathcal{k}$ consists of a set of classes $C^{k} = \{c^{k}_{1},c^{k}_{2}\text{, ... ,}c^{k}_{m^{k}}\}$ and $D^{k}$ is the training data. The point cloud class-incremental problem in which $D^k = \{(p_{1},y_{1}), (p_{2},y_{2})\text{, ... ,}(p_{l^{k}},y_{l^{k}})\}$, where $p$ is point cloud with $n$ points such that $p \in \mathbb{R}^{n \times 3}$. During training for task $k$, the learner only has access to $C^k, D^k$, whereas, during inference, the evaluation is done for the union of all previous tasks $\bigcup_{i=1}^{k} C^i, D^i$. For instance if we encounter task $k=2$, the learner has access to $(C^2,D^2)$ where as evaluation is done for $\{(C^1,D^1), (C^2,D^2)\}$.   

\noindent \textbf{Class-Incremental learning on 3D-Grocery:} we propose to benchmark a baseline approach known as ``Learning Without Forgetting" (LWF)~\cite{mccloskey1989catastrophic} used in the 2D realm to address the issue of catastrophic forgetting. We extend the methodology used in the 2D realm to 3D by adopting features from SOTA point cloud processing architectures such as PointNet~\cite{qi2017pointnet}, PointNet++~\cite{qi2017pointnet++}, DGCNN~\cite{wang2019dynamic}, PointMLP~\cite{ma2022rethinking}, and PCT~\cite{guo2021pct}, and using a dynamic multi-head classifier which adjusts itself automatically according to the novel classes that arrive in stream.

\begin{table}[!ht]
\centering
\caption{Performance of different backbone's on 3DGrocery63 dataset in a CIL scenario. \textbf{\textit{Joint:}} Upper bound, \textbf{\textit{FT:}} Fine-Tuning.}
\label{tab:CIL}
\resizebox{\linewidth}{!}{%
\begin{tabular}{rcccccc}
\hline 
 &
   &
  \textbf{39} &
  \textbf{6} &
  \textbf{6} &
  \textbf{6} &
  \textbf{6} \\ \cline{3-7} 
\multirow{-2}{*}{\textbf{Backbone}} &
  \multirow{-2}{*}{\textbf{\begin{tabular}[c]{@{}c@{}} \# Classes $\rightarrow$\\  Loss $\downarrow$\end{tabular}}} &
  \textbf{Acc} &
  \textbf{Acc} &
  \textbf{Acc} &
  \textbf{Acc} &
  \textbf{Acc} \\ \hline 
\textbf{} &
  \textit{\textbf{FT}} &
  41.40 &
  07.61 &
  04.62 &
  06.27 &
  03.49 \\
\textbf{PointNet~\cite{qi2017pointnet}} &
  \textit{\textbf{LwF}} &
  41.40 &
  07.55 &
  06.34 &
  06.42 &
  04.77 \\
\textbf{} &
  \cellcolor[HTML]{DAE8FC}\textit{\textbf{Joint}} &
  \cellcolor[HTML]{DAE8FC}41.40 &
  \cellcolor[HTML]{DAE8FC}42.46 &
  \cellcolor[HTML]{DAE8FC}43.50 &
  \cellcolor[HTML]{DAE8FC}43.99 &
  \cellcolor[HTML]{DAE8FC}44.07 \\ \hline
\textbf{} &
  \textit{\textbf{FT}} &
  55.33 &
  07.99 &
  04.89 &
  06.70 &
  04.34 \\
\textbf{PointNet++~\cite{qi2017pointnet++}} &
  \textit{\textbf{LwF}} &
  55.33 &
  13.45 &
  07.17 &
  07.17 &
  05.94 \\
\textbf{} &
  \cellcolor[HTML]{DAE8FC}\textit{\textbf{Joint}} &
  \cellcolor[HTML]{DAE8FC}55.33 &
  \cellcolor[HTML]{DAE8FC}55.87 &
  \cellcolor[HTML]{DAE8FC}56.49 &
  \cellcolor[HTML]{DAE8FC}57.05 &
  \cellcolor[HTML]{DAE8FC}57.65 \\ \hline
\textbf{} &
  \textit{\textbf{FT}} &
  49.02 &
  08.52 &
  06.40 &
  06.61 &
  04.58 \\
\textbf{DGCNN~\cite{wang2019dynamic}} &
  \textit{\textbf{LwF}} &
  49.02 &
  \underline{17.75} &
  \textbf{07.55} &
  \underline{07.33} &
  \textbf{06.25} \\
\textbf{} &
  \cellcolor[HTML]{DAE8FC}\textit{\textbf{Joint}} &
  \cellcolor[HTML]{DAE8FC}49.02 &
  \cellcolor[HTML]{DAE8FC}50.91 &
  \cellcolor[HTML]{DAE8FC}50.10 &
  \cellcolor[HTML]{DAE8FC}51.04 &
  \cellcolor[HTML]{DAE8FC}51.23 \\ \hline
\textbf{} &
  \textit{\textbf{FT}} &
  51.16 &
  07.74 &
  04.85 &
  06.44 &
  04.29 \\
\textbf{PointMLP-E~\cite{ma2022rethinking}} &
  \textit{\textbf{LwF}} &
  51.16 &
  10.79 &
  06.71 &
  06.62 &
  05.56 \\
\textbf{} &
  \cellcolor[HTML]{DAE8FC}\textit{\textbf{Joint}} &
  \cellcolor[HTML]{DAE8FC}51.16 &
  \cellcolor[HTML]{DAE8FC}52.14 &
  \cellcolor[HTML]{DAE8FC}50.69 &
  \cellcolor[HTML]{DAE8FC}51.03 &
  \cellcolor[HTML]{DAE8FC}51.06 \\ \hline
\textbf{} &
  \textit{\textbf{FT}} &
  22.30 &
  04.37 &
  03.10 &
  04.19 &
  01.80 \\
\textbf{PCT~\cite{guo2021pct}} &
  \textit{\textbf{LwF}} &
  22.30 &
  04.96 &
  04.26 &
  04.43 &
  02.58 \\
\textbf{} &
  \cellcolor[HTML]{DAE8FC}\textit{\textbf{Joint}} &
  \cellcolor[HTML]{DAE8FC}22.30 &
  \cellcolor[HTML]{DAE8FC}22.09 &
  \cellcolor[HTML]{DAE8FC}21.26 &
  \cellcolor[HTML]{DAE8FC}22.53 &
  \cellcolor[HTML]{DAE8FC}25.79 \\ \hline
\textbf{} &
  \textit{\textbf{FT}} &
  54.65 &
  07.97 &
  04.84 &
  06.72 &
  03.76 \\
\textbf{PointNeXT~\cite{PointNeXt}} &
  \textit{\textbf{LwF}} &
  54.84 &
  \textbf{21.41} &
  \underline{06.51} &
  \textbf{07.56} &
  \underline{05.87} \\
\textbf{} &
  \cellcolor[HTML]{DAE8FC}\textit{\textbf{Joint}} &
  \cellcolor[HTML]{DAE8FC}54.43 &
  \cellcolor[HTML]{DAE8FC}55.80 &
  \cellcolor[HTML]{DAE8FC}56.15 &
  \cellcolor[HTML]{DAE8FC}57.78 &
  \cellcolor[HTML]{DAE8FC}58.28 \\ \hline 
\end{tabular}%
}
\end{table}

The combination of baseline and these advanced architectures results in our extended method called ``LWF" and is compared with baselines Fine-tuning (FT) / Lower-bound, and Joint-training / Upper-bound. LWF aims to mitigate catastrophic forgetting in point clouds, ensuring our dataset's adaptability to new and evolving classes while preserving the learned knowledge from previous data. For benchmarking on the CIL setting, we split our dataset into five tasks, with the base task involving 39 classes and the remaining tasks with six classes each arriving incrementally. We train a joint head classifier, which adjusts according to the arrived novel classes. We benchmark point cloud CIL with each task trained on 40 epochs. PointNeXt~\cite{PointNeXt} and DGCNN~\cite{wang2019dynamic} perform better on incremental tasks compared to other backbones, as shown in Table~\ref{tab:CIL}.

\section{Conclusion}
This paper introduces the largest real-world 3D dataset on groceries called 3DGrocery100. One of the key contributions of this dataset is its wide and fine-grained variety of grocery categories. It contains 100 classes divided into 10, 24, 28, and 38: Apples, Fruits (non-apples), Vegetables, and Packages. High-resolution 10,755 RGB-D images were collected using mobile phones with 3D sensors and were processed to create the largest real-world point cloud dataset of 87,898 objects. In addition, this dataset is diverse due to the presence of point cloud objects of varying sizes under natural occlusions. We benchmarked six representative state-of-the-art methods on all five subsets and two color-based variants of our dataset. Our dataset stands out with its distinctive fine-grained features, making it an excellent benchmark for few-shot classification tasks, especially in strong generalization. Given the challenge of encountering novel classes, we focus on evaluating class-incremental learning approaches for classification. This evaluation will help us assess the dataset's ability to handle the incorporation of new classes and maintain robust classification performance while building on the knowledge acquired from existing data.
{
    \small
    \bibliography{main}
}
\clearpage
\setcounter{page}{1}
\maketitlesupplementary

We show the use of the packages subset dataset to pre-train the PointNeXt model to achieve the state-of-the-art result on the hardest variant of the ScanObjectNN dataset. We provide counts of RGB-D images for each class. We show five visual samples from each of the 100 classes. We show the confusion matrices from evaluations of our dataset on six state-of-the-art object classification models. 

\section{Packages subset for Pre-training}

Unlike Fruits or Vegetables, Packages' shapes are unique because they do not show much deformation at each instance level. For example, the shape of beans or Gatorade bottles has consistency across multiple instances, whereas different broccoli samples have different shapes. This shape uniqueness of packages and high performance of package classification, even in the absence of colors, make the subset dataset suitable for pretraining methods upon which small datasets can be fine-tuned to achieve a gain in classification accuracy as shown in Table~\ref{table6}.

\begin{table}[H]\centering
\caption{Pre-training PointNeXt~\cite{PointNeXt} model on the packages subset (w/o colors) and fine-tuning with the ScanObjectNN's hardest variant improves classification accuracy by 0.64\%.}\label{table6}
\begin{tabular}{|c|c|}
\hline
  Method & Acc. (\%)\\
\hline
PointNeXt~\cite{PointNeXt} & 87.70\\
PointNeXt w/ pre-training & 88.34 ($\uparrow$ \textbf{0.64}) \\
\hline
\end{tabular}
\end{table}

\section{Number of RGB-D Images}

\begin{figure}[H]
\centering
\includegraphics[width=\linewidth,trim={5mm 5mm 5mm 5mm},clip]{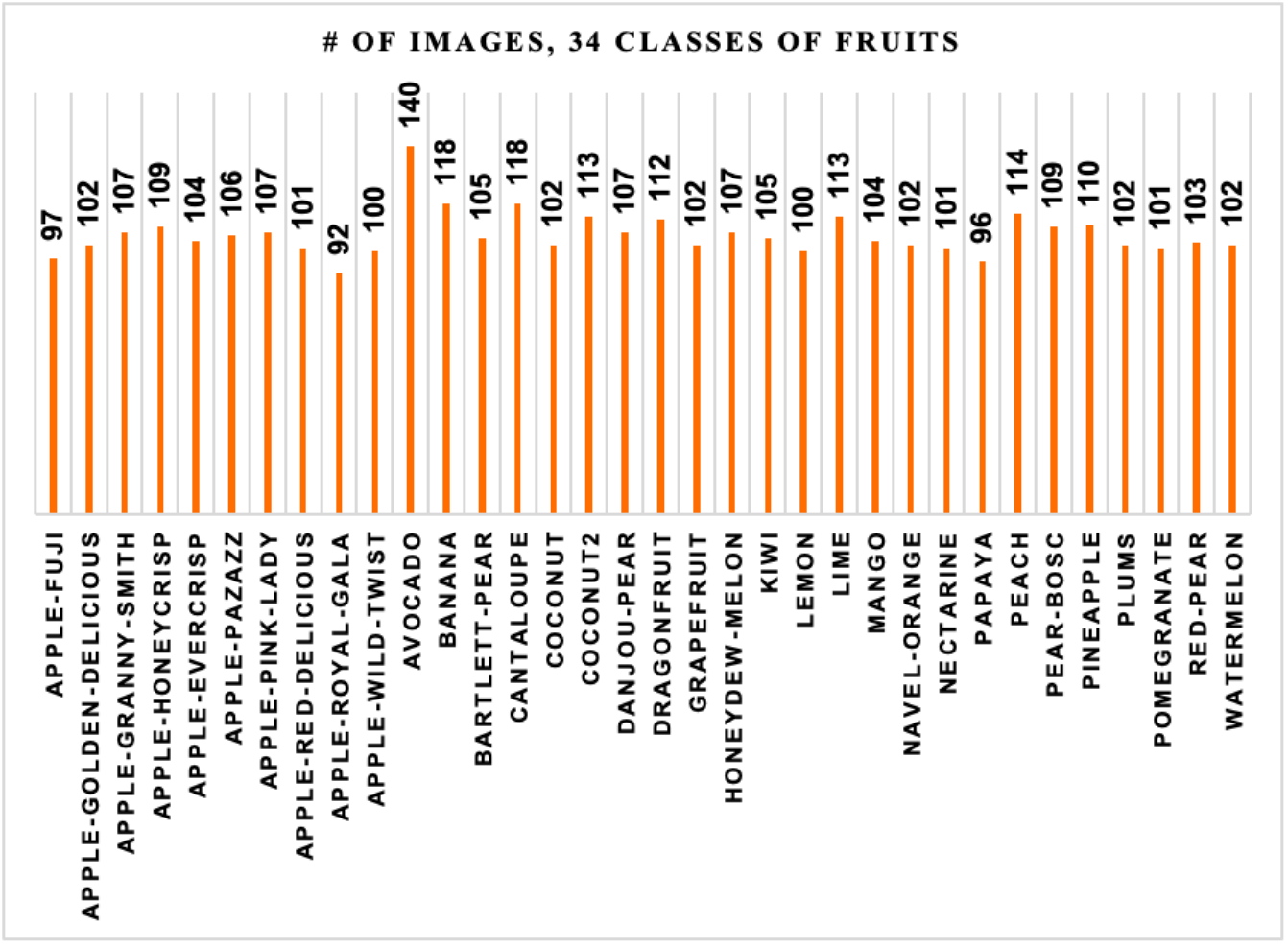}
\caption{34 Fruit Classes: each with the count of images.}
\label{Fig5}
\end{figure}

\begin{figure}[H]
\centering
\includegraphics[width=0.85\linewidth,trim={5mm 5mm 5mm 5mm},clip]{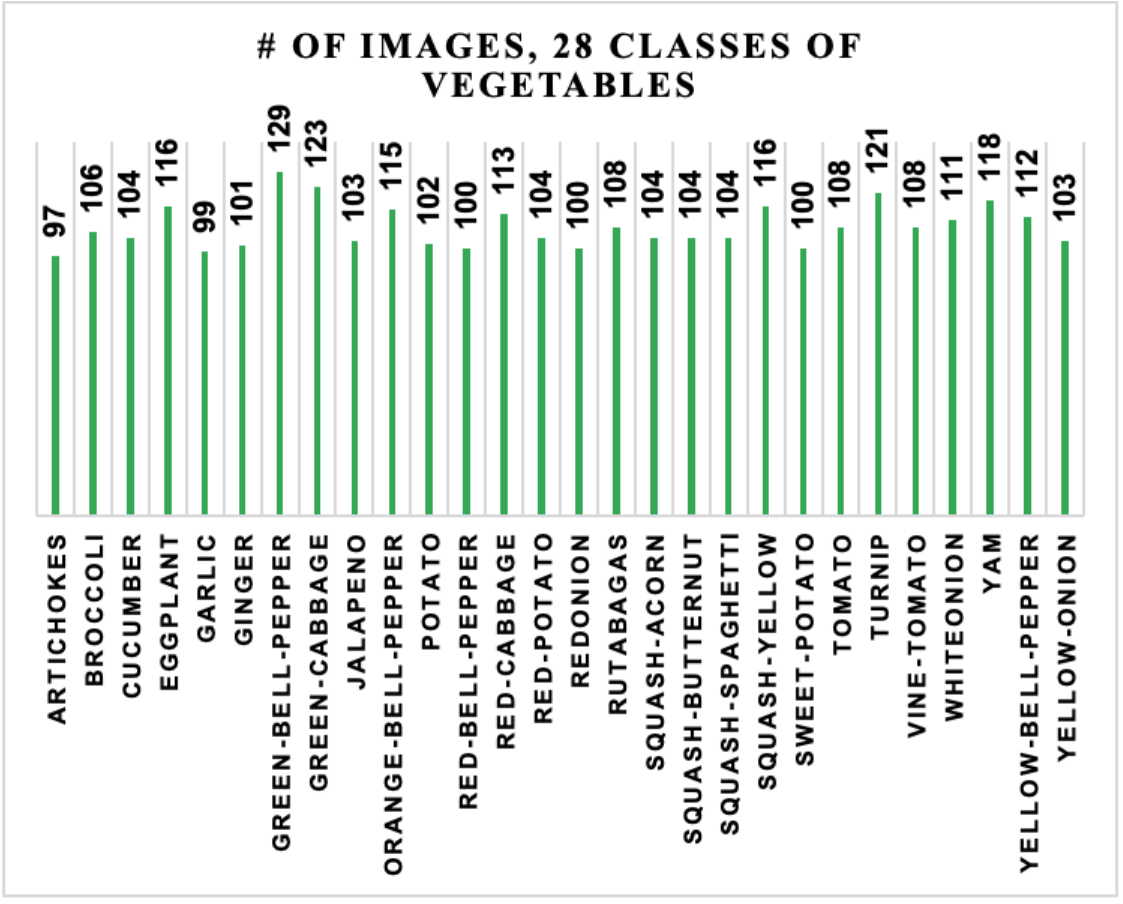}
\caption{28 Vegetable Classes: each with the count of images.}
\label{Fig6}
\end{figure}

\begin{figure}[H]
\centering
\includegraphics[width=0.85\linewidth,trim={5mm 5mm 5mm 5mm},clip]{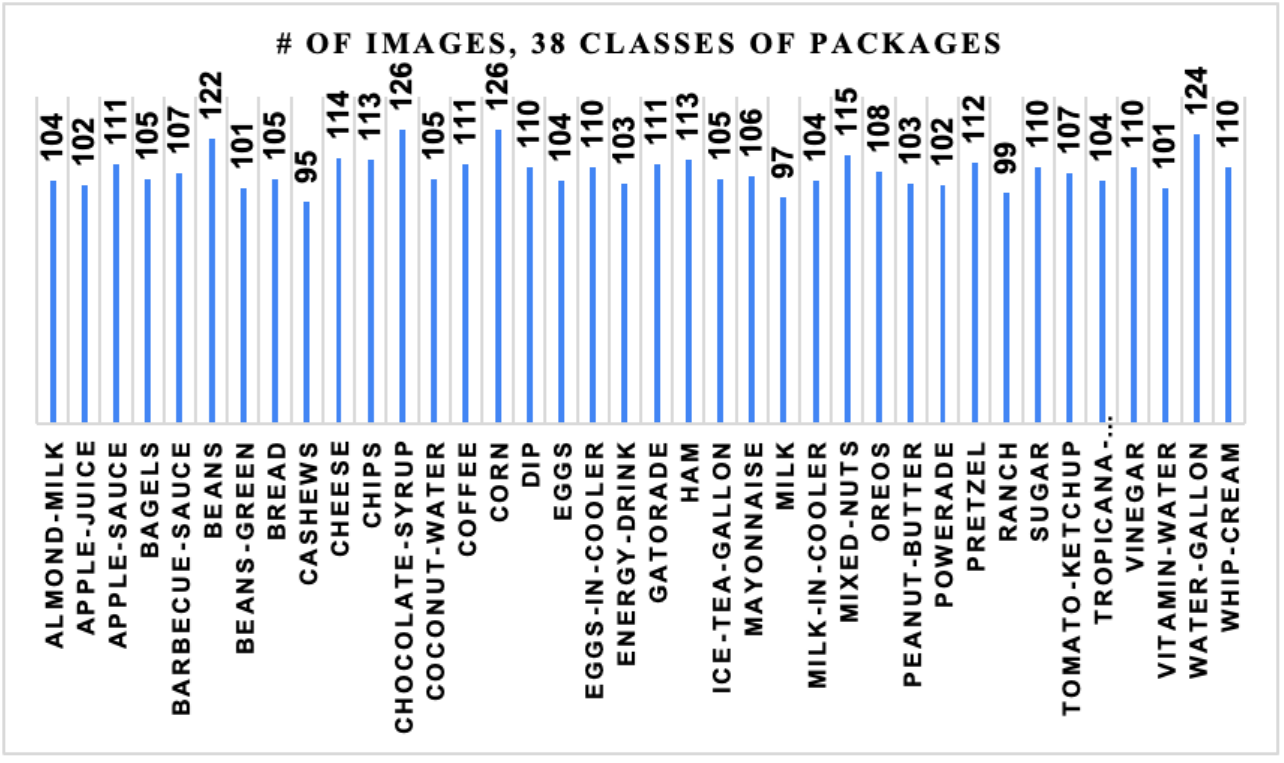}
\caption{38 Package Classes: each with the count of images.}
\label{Fig7}
\end{figure}

Our dataset consists of 10,755 RGB-D images spread across 100 classes of groceries, as shown in Figure~\ref{Fig5} for fruits, Figure~\ref{Fig6} for vegetables, and Figure~\ref{Fig7} for packages. Each of these images is annotated.

\begin{figure}[H]
\centering
\includegraphics[width=0.8\linewidth,trim={10mm 0mm 0mm 0mm},clip]{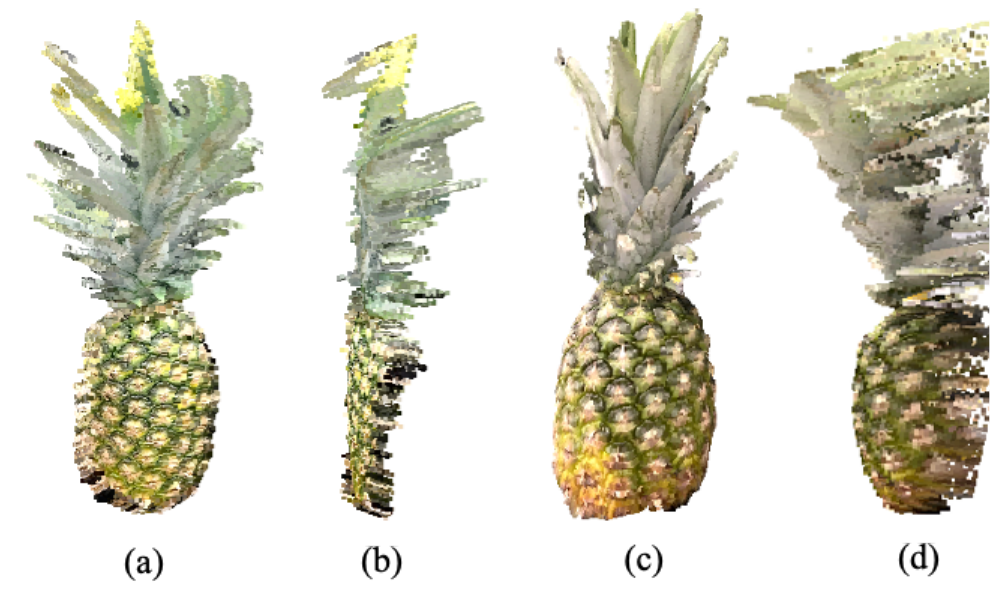}
\caption{Comparison between LiDAR (a) front view (b) side view and Stereo Vision (c) front view (d) side view of pineapple instance after conversion to a point cloud.}
\label{Fig8}
\end{figure}

\begin{figure}[H]
\centering
\includegraphics[width=0.99\linewidth]{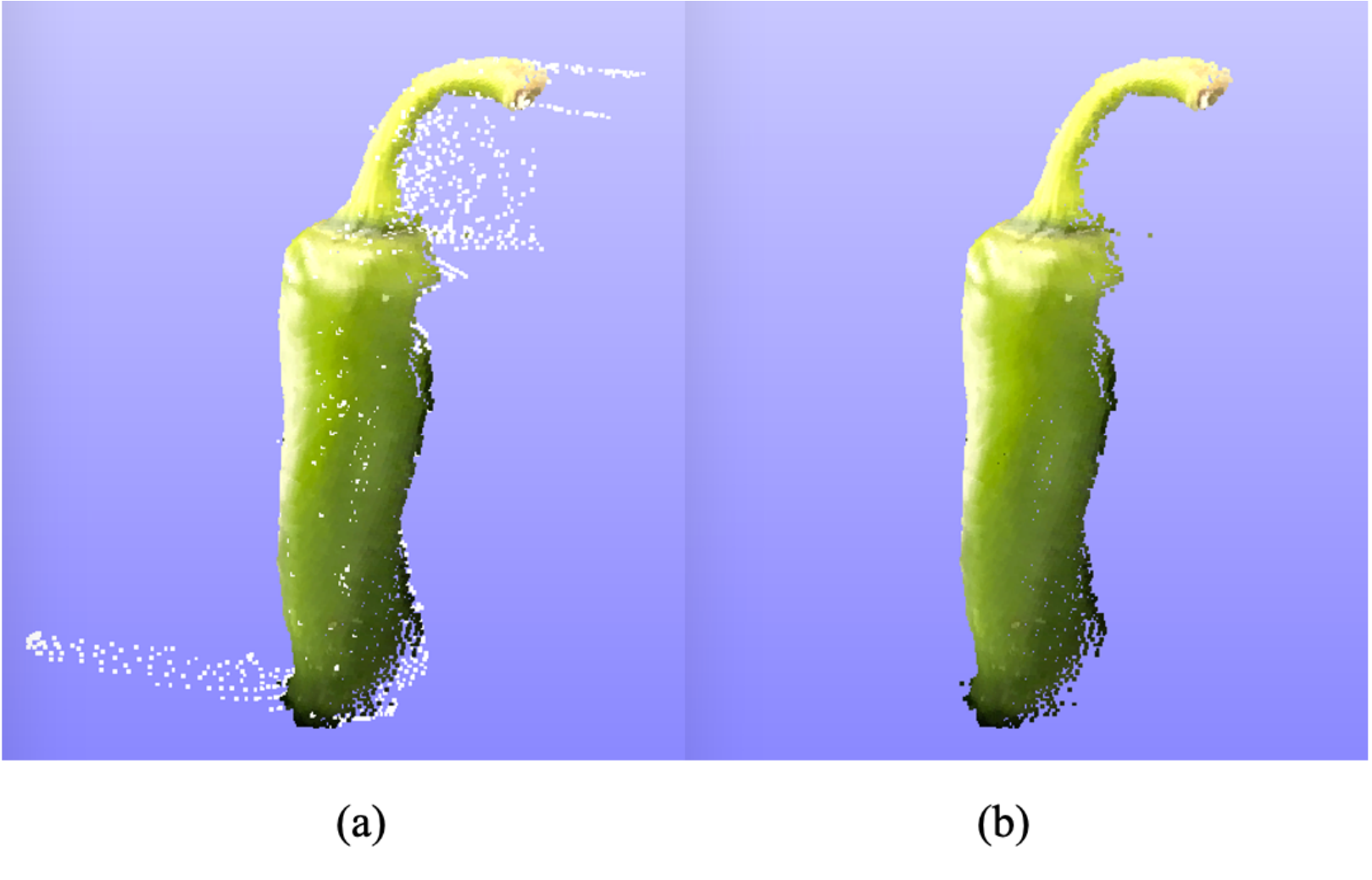}
\caption{Jalapeno object example: a) point cloud with outliers (colored with white for visibility)  b) point cloud after removing outliers using PointCleanNet~\cite{rakotosaona2020pointcleannet} with a threshold of 0.4.}
\label{Fig9}
\end{figure}

\section{Visual samples of 100 classes}

\begin{figure}[H]
\centering
(a) apple-evercrisp
\includegraphics[width=0.95\linewidth,trim={2mm 0mm 6mm 7mm},clip]{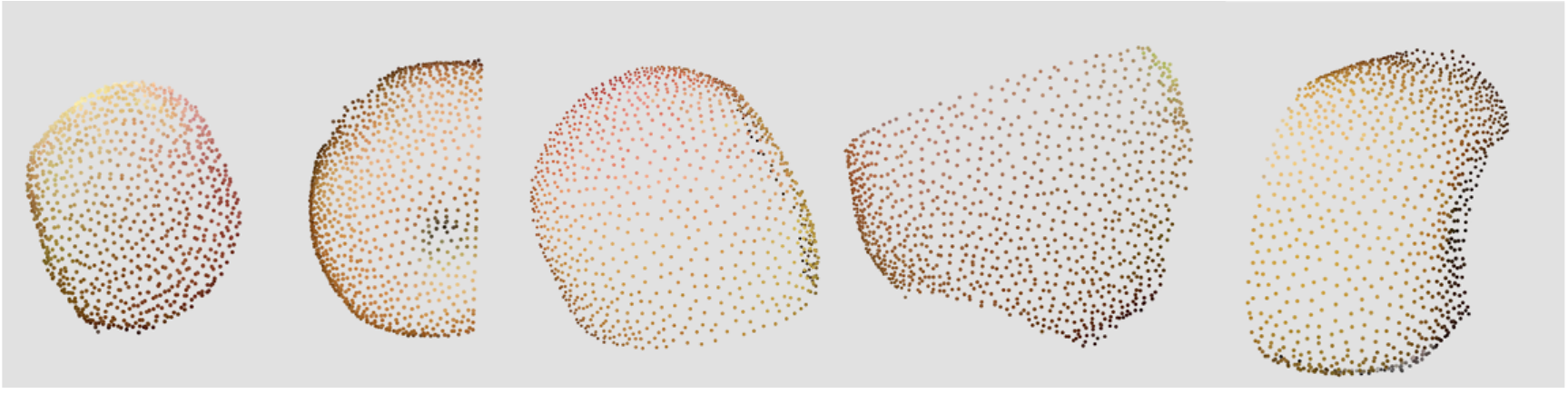}
(b) apple-fuji
\includegraphics[width=0.95\linewidth,trim={2mm 4mm 2mm 7mm},clip]{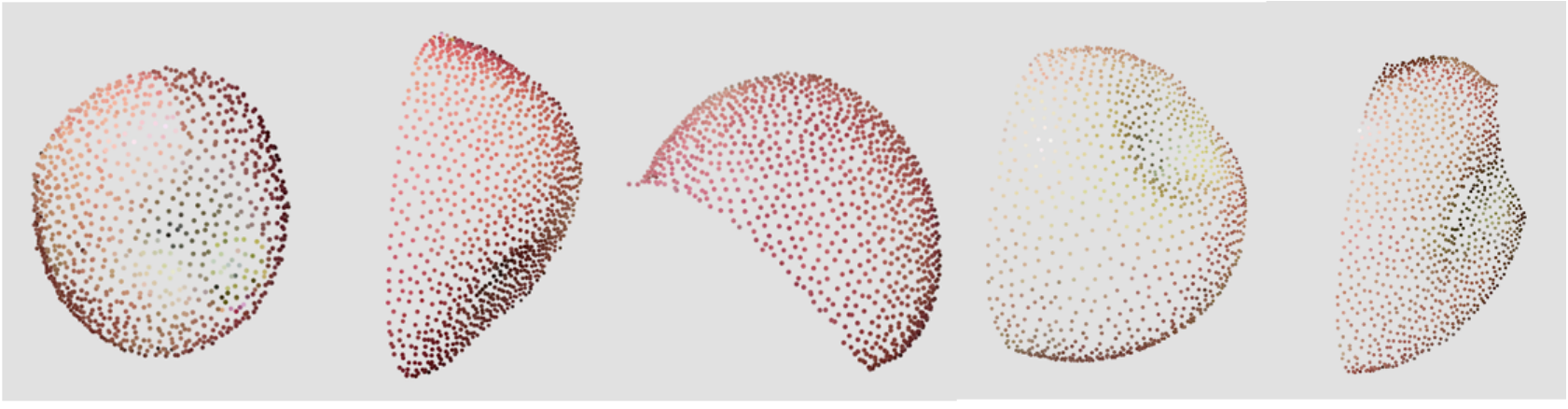}
(c) apple-golden-delicious
\includegraphics[width=0.95\linewidth,trim={1mm 7mm 1mm 9mm},clip]{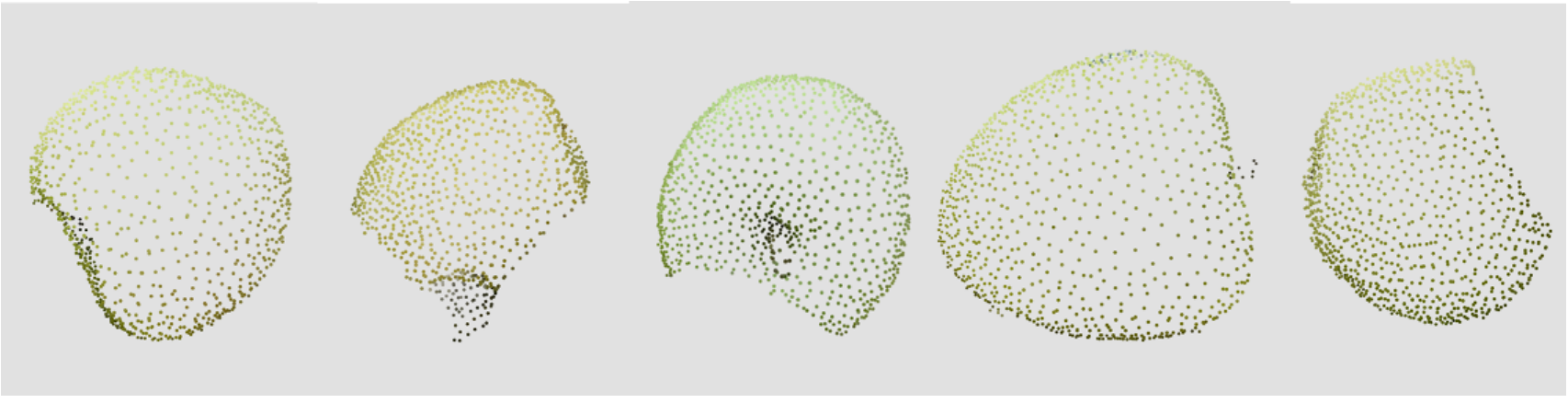}
(d) apple-granny-smith
\includegraphics[width=0.95\linewidth,trim={0mm 6mm 2mm 8mm},clip]{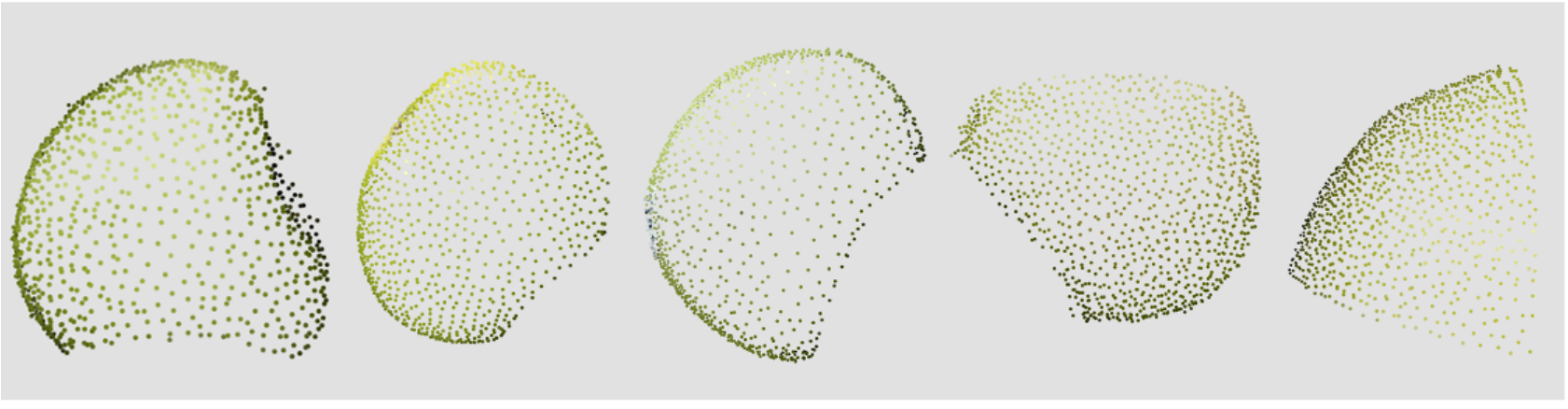}
(e) apple-honeycrisp
\includegraphics[width=0.95\linewidth,trim={0mm 6mm 0mm 10mm},clip]{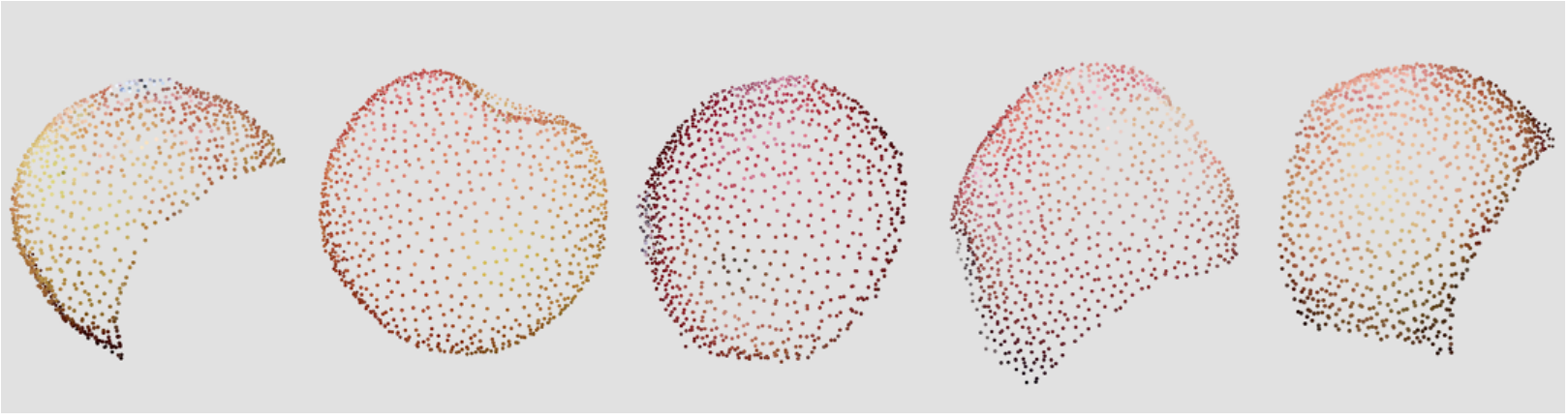}
(f) apple-pazazz
\includegraphics[width=0.95\linewidth,trim={2mm 7mm 1mm 10mm},clip]{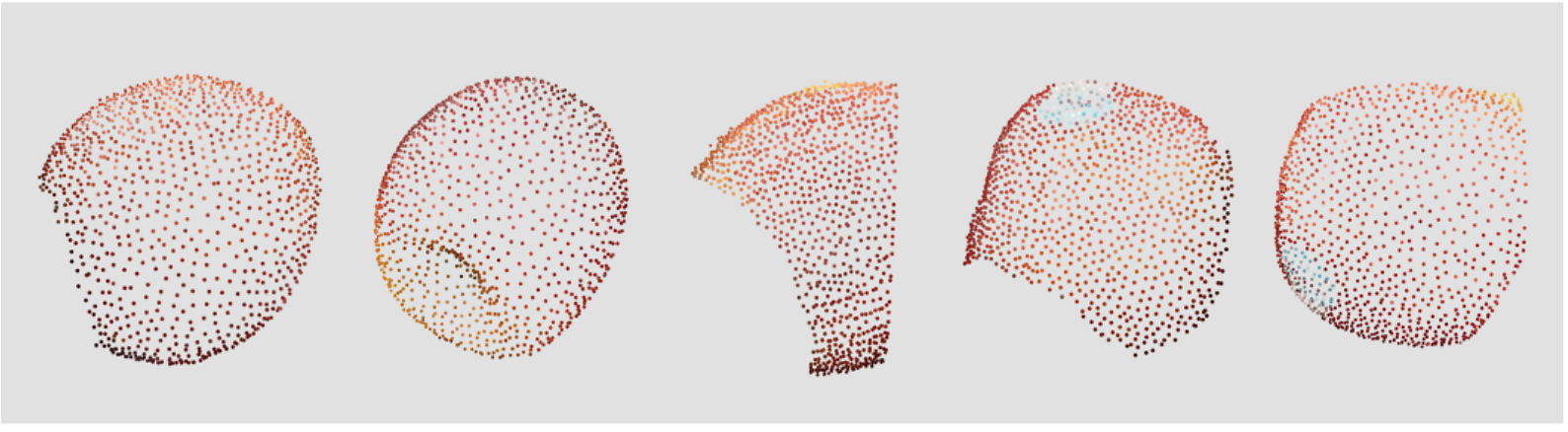}
(g) apple-pink-lady
\includegraphics[width=0.95\linewidth,trim={2mm 9mm 1mm 8mm},clip]{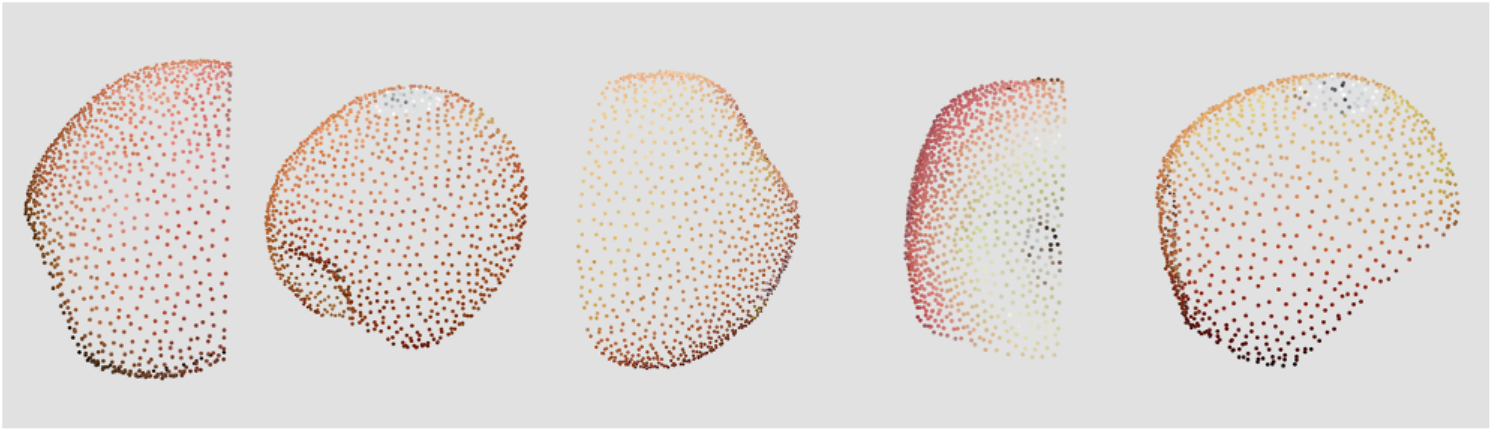}
(h) apple-red-delicious
\includegraphics[width=0.95\linewidth,trim={2mm 6mm 1mm 9mm},clip]{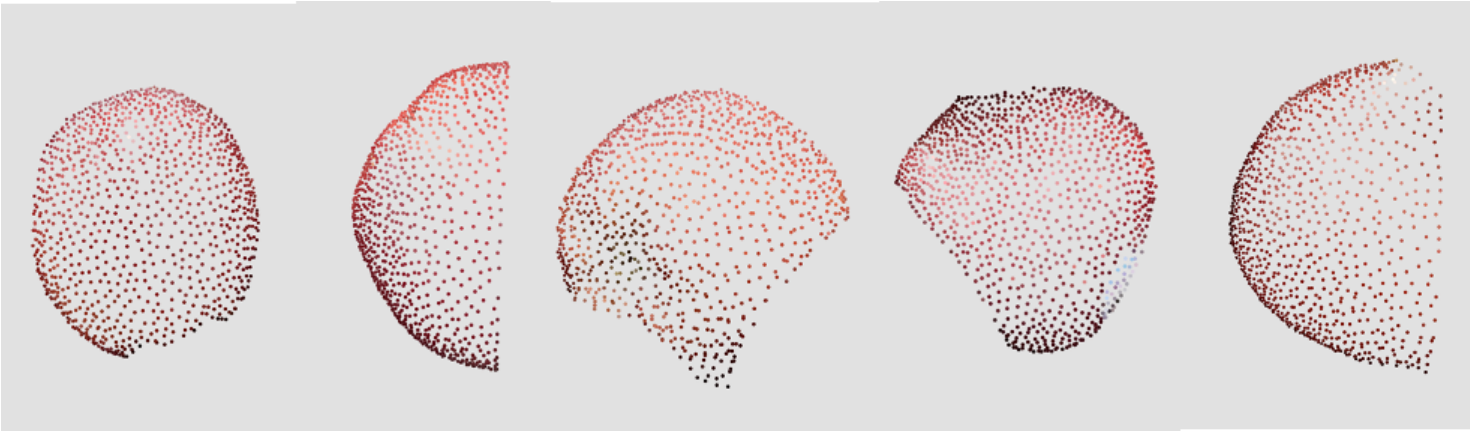}

(i) apple-royal-gala
\includegraphics[width=0.95\linewidth,trim={2mm 9mm 2mm 12mm},clip]{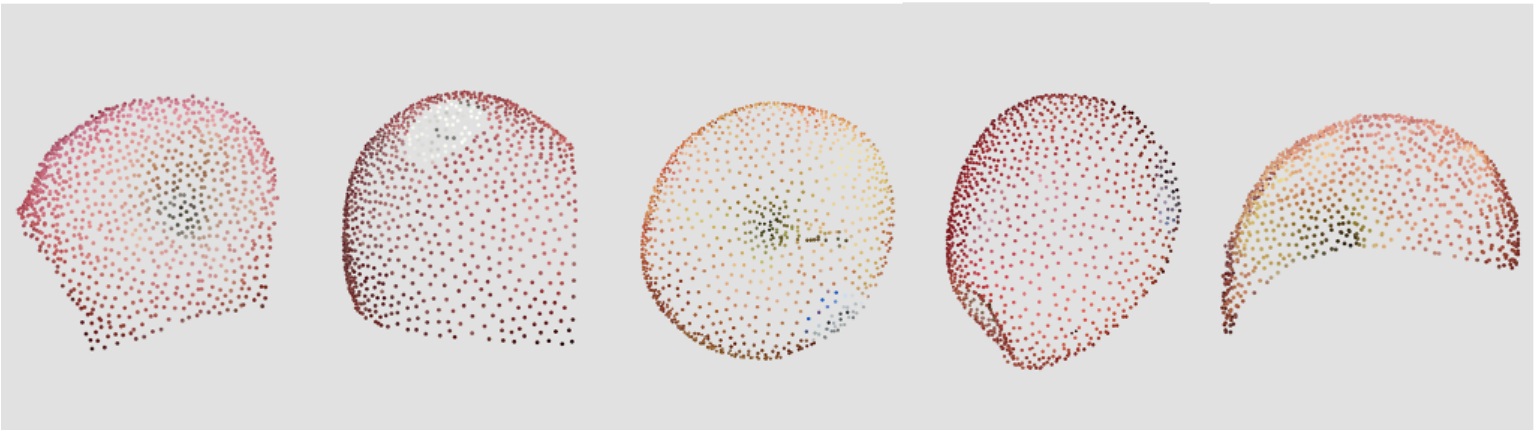}

(j) apple-wild-twist
\includegraphics[width=0.95\linewidth,trim={2mm 10mm 2mm 10mm},clip]{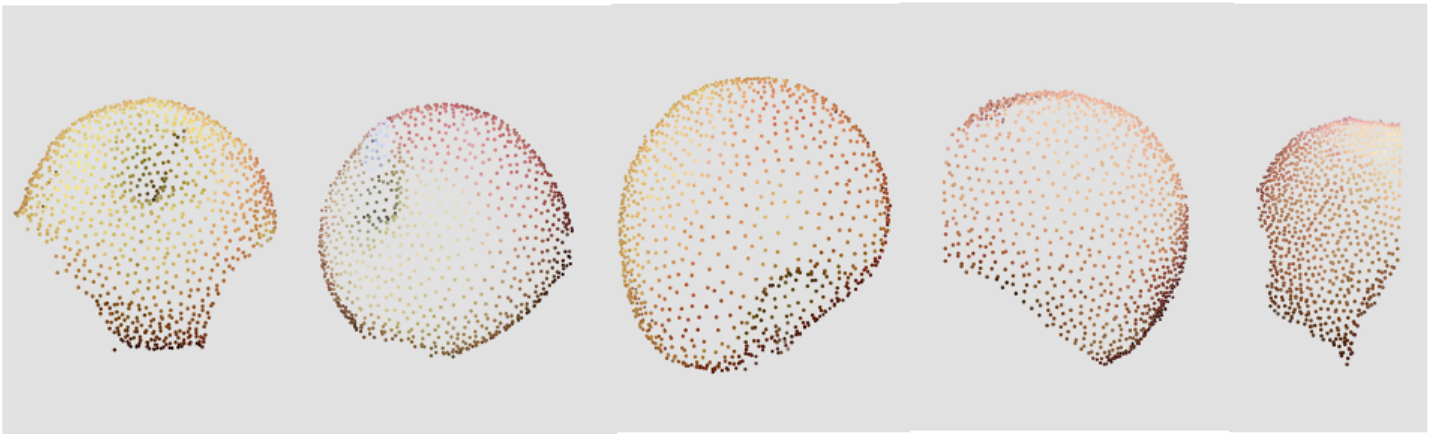}
\caption{Visual samples (1024 points) from 10 Apple classes. Labels on top of the objects. Zoom in for better visibility.}
\label{Fig10}
\end{figure}

\begin{figure}[H]
\centering
(a) avocado
\includegraphics[width=0.95\linewidth,trim={2mm 5mm 1mm 15mm},clip]{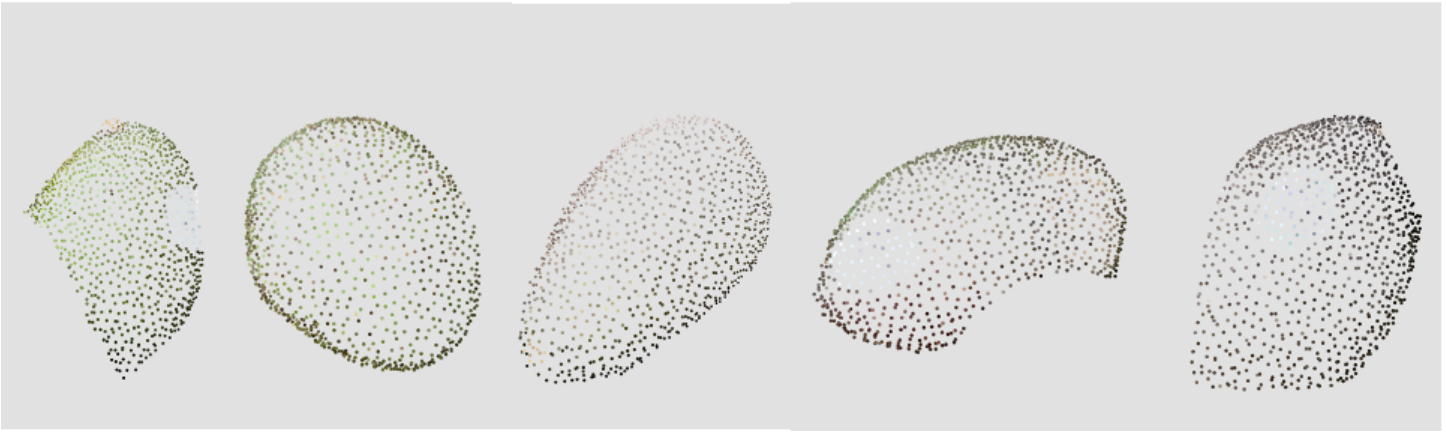}
(b) banana
\includegraphics[width=0.95\linewidth,trim={2mm 8mm 1mm 12mm},clip]{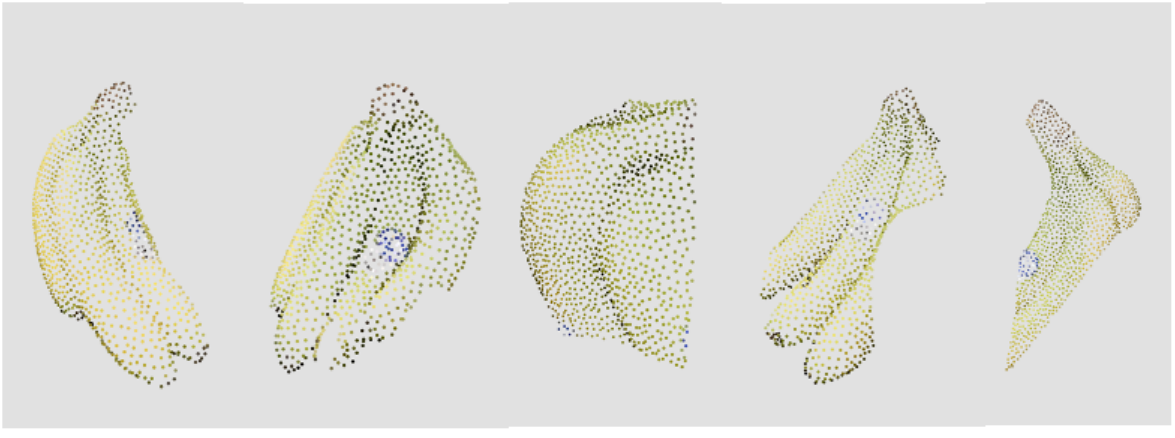}
(c) bartlett-pear
\includegraphics[width=0.95\linewidth,trim={1mm 11mm 5mm 2mm},clip]{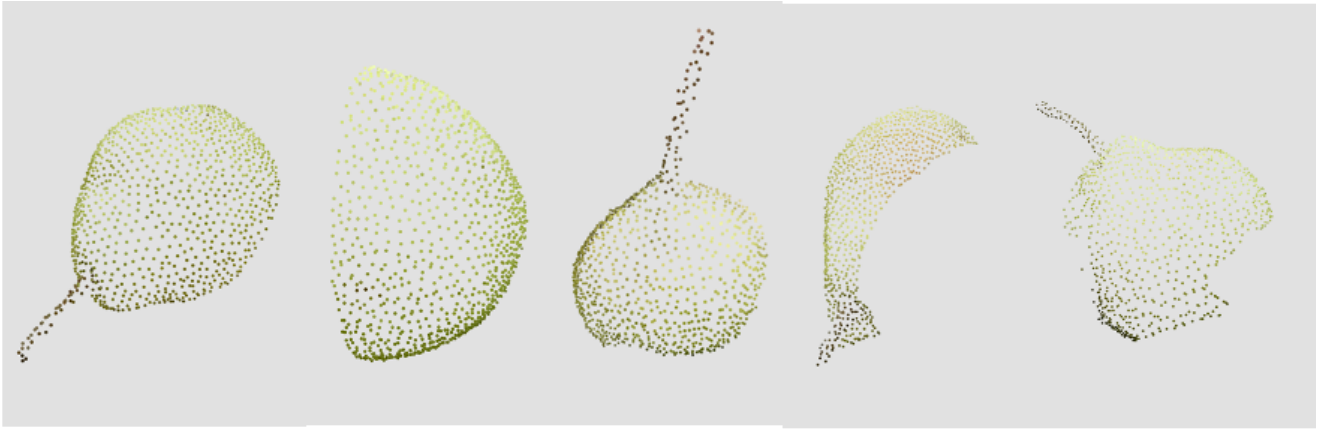}
(d) cantaloupe
\includegraphics[width=0.95\linewidth,trim={0mm 10mm 2mm 9mm},clip]{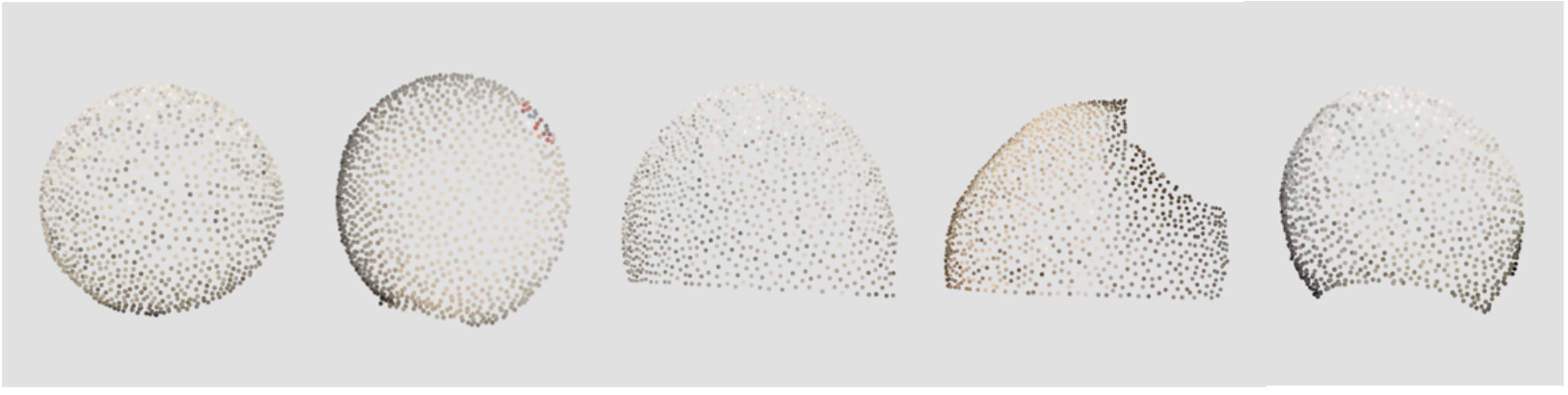}
(e) coconut
\includegraphics[width=0.95\linewidth,trim={1mm 6mm 5mm 10mm},clip]{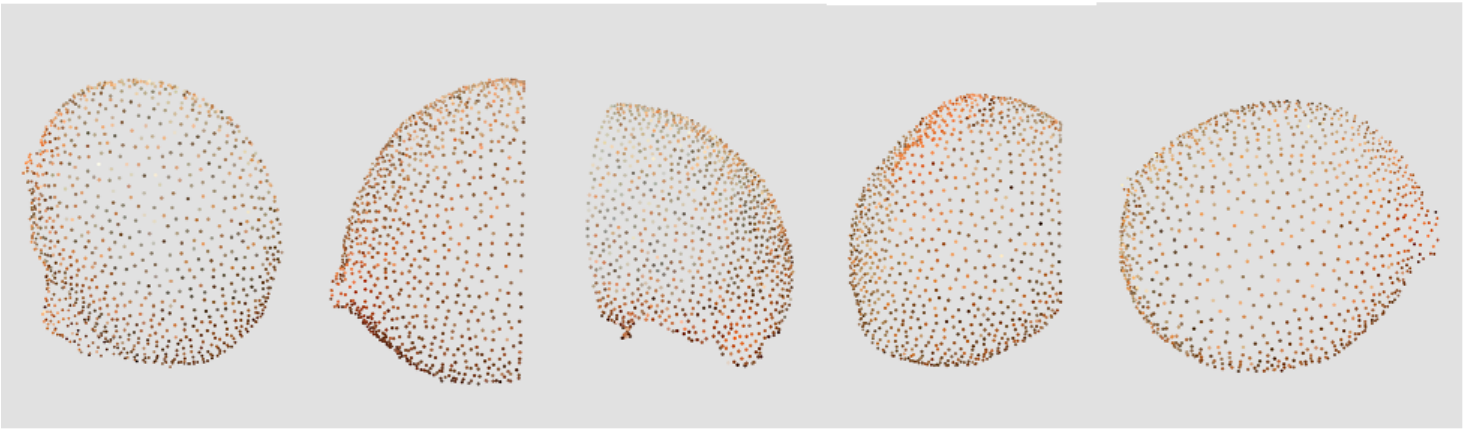}
(f) coconut2
\includegraphics[width=0.95\linewidth,trim={2mm 8mm 4mm 11mm},clip]{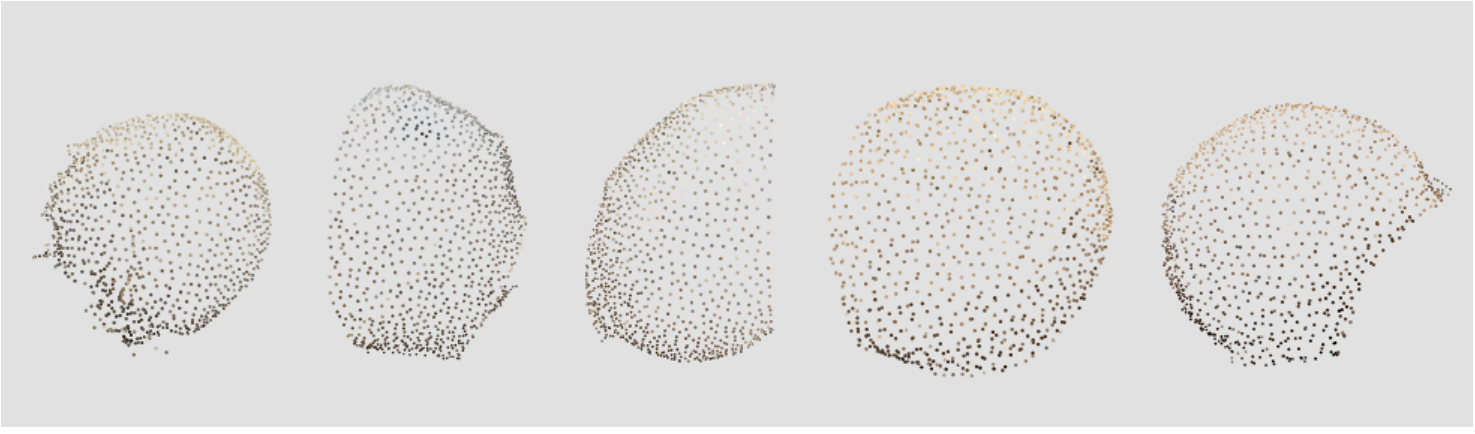}
(g) danjou-pear
\includegraphics[width=0.95\linewidth,trim={1mm 13mm 2mm 10mm},clip]{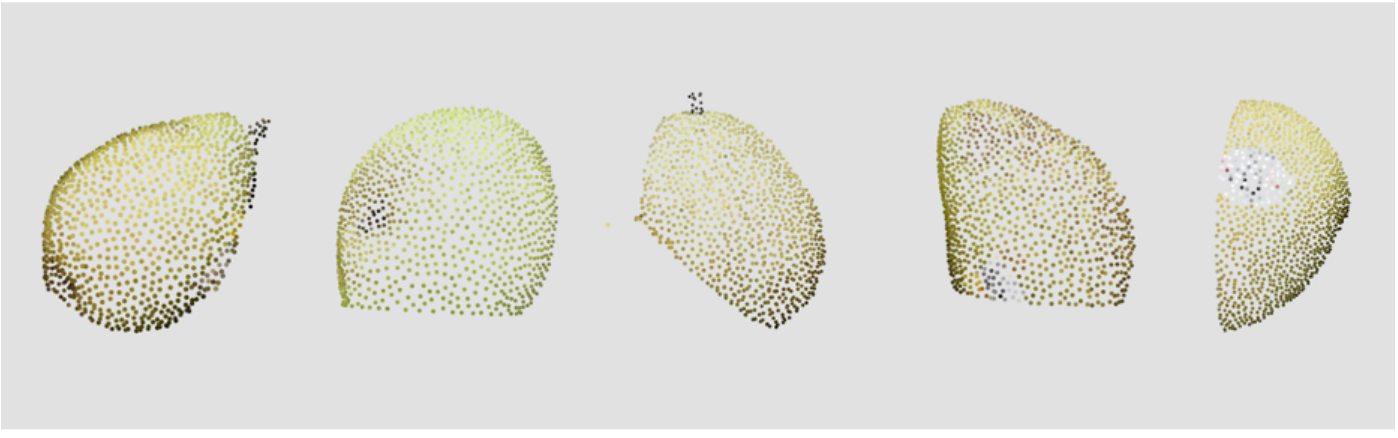}
(h) dragonfruit
\includegraphics[width=0.95\linewidth,trim={1mm 6mm 0mm 10mm},clip]{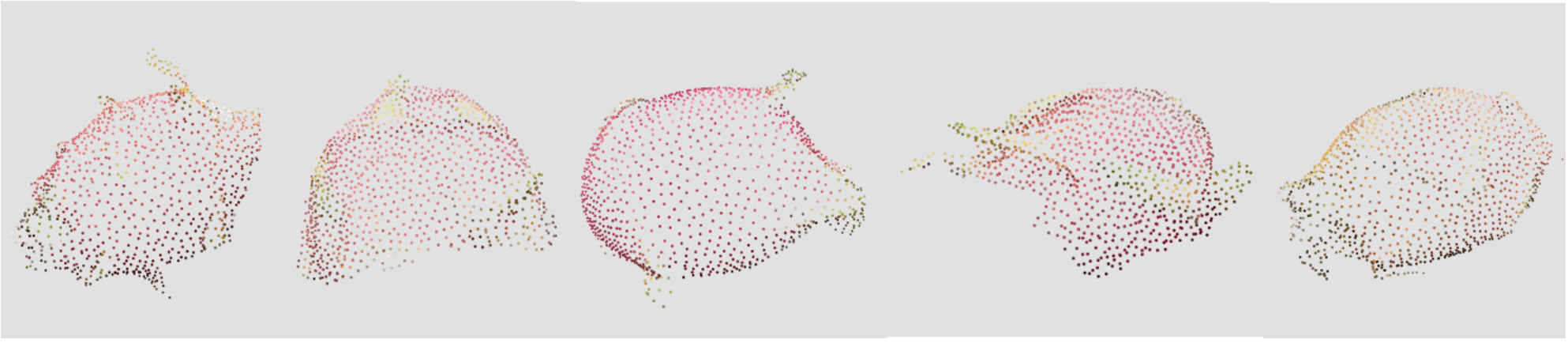}

(i) grapefruit
\includegraphics[width=0.95\linewidth,trim={1mm 5mm 0mm 13mm},clip]{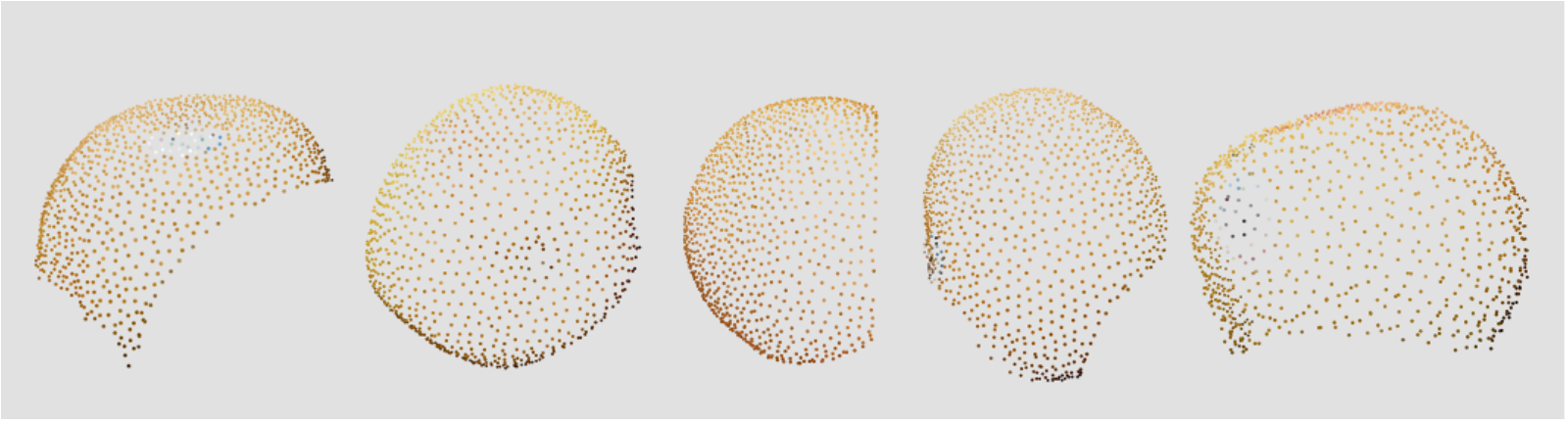}

(j) honeydew-melon
\includegraphics[width=0.95\linewidth,trim={1mm 5mm 0mm 10mm},clip]{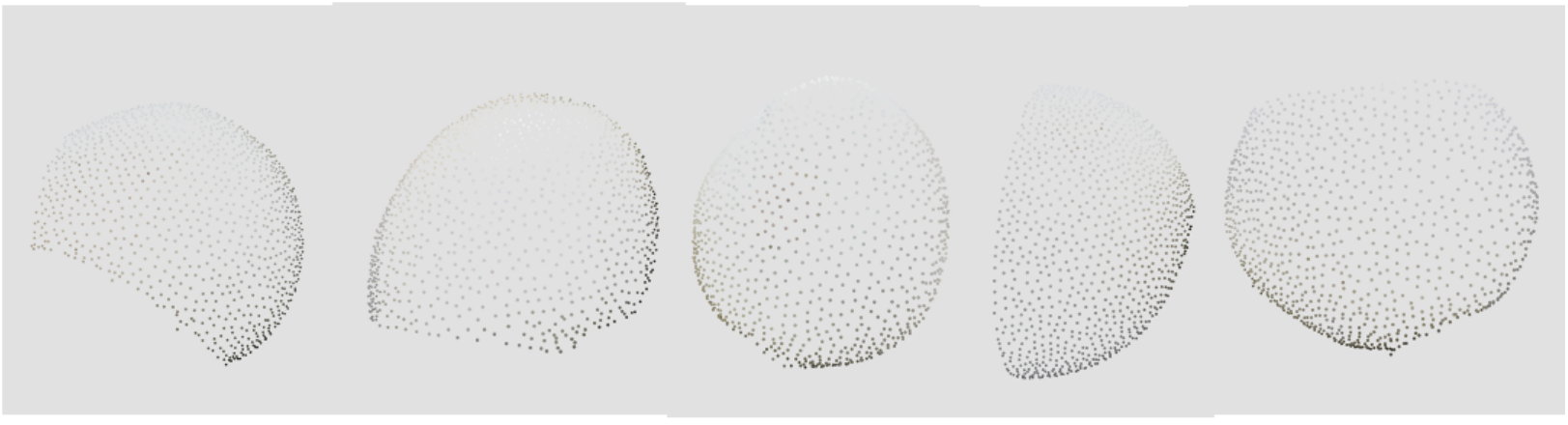}
\caption{Visual samples (1024 points) from 10 of 24 non-apple fruit classes. Labels on top of the objects.}
\label{Fig11}
\end{figure}

\begin{figure}[H]
\centering
(a) kiwi
\includegraphics[width=0.95\linewidth,trim={2mm 7mm 5mm 13mm},clip]{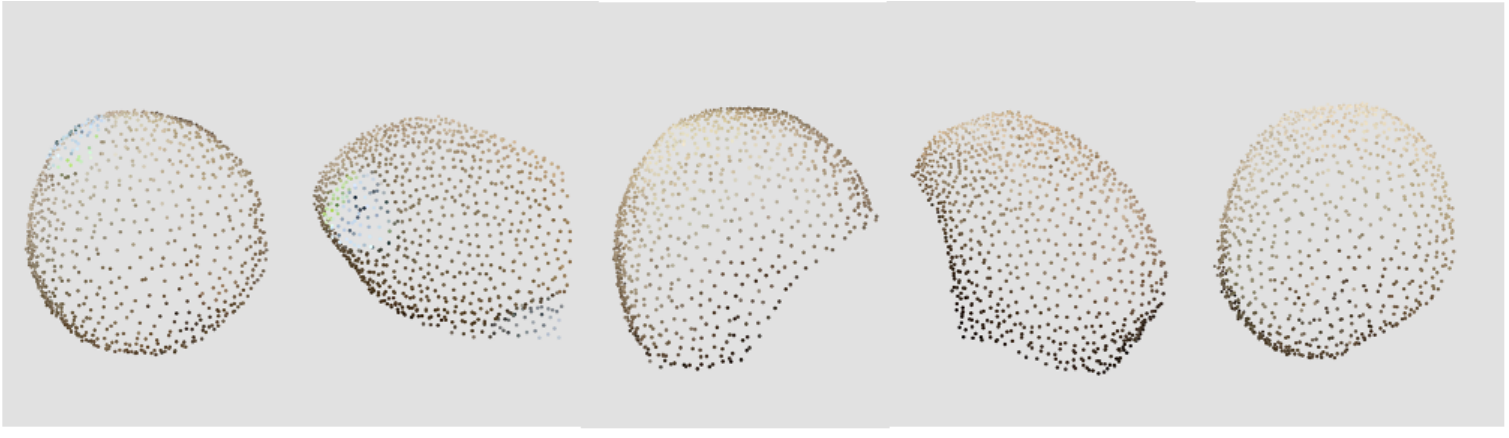}
(b) lemon
\includegraphics[width=0.95\linewidth,trim={2mm 6mm 5mm 13mm},clip]{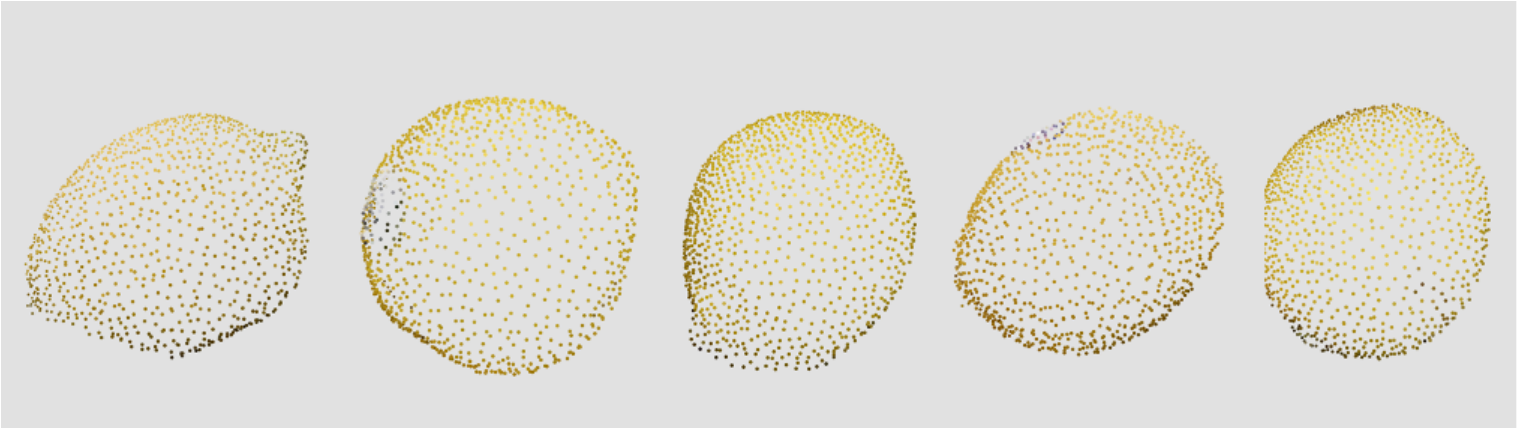}
(c) lime
\includegraphics[width=0.95\linewidth,trim={2mm 8mm 5mm 11mm},clip]{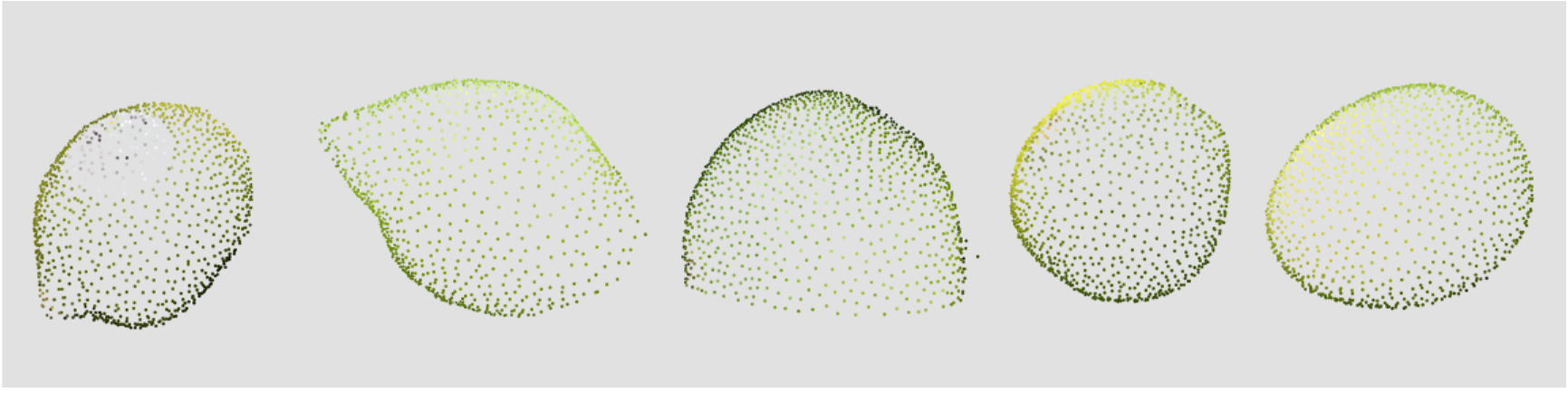}
(d) mango
\includegraphics[width=0.95\linewidth,trim={0mm 6mm 3mm 11mm},clip]{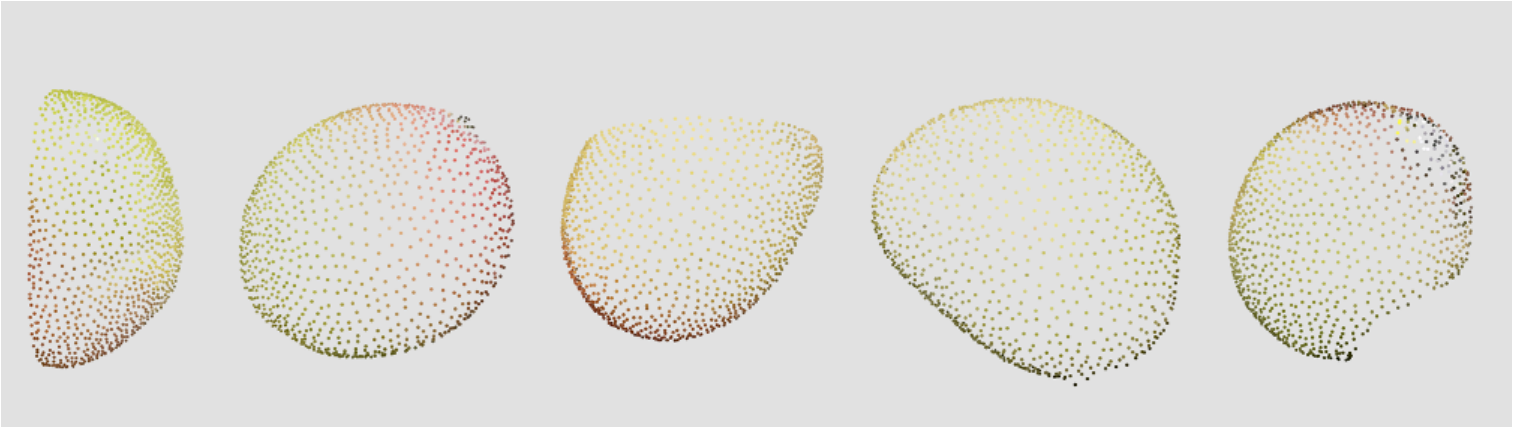}
(e) navel-orange
\includegraphics[width=0.95\linewidth,trim={0mm 5mm 3mm 6mm},clip]{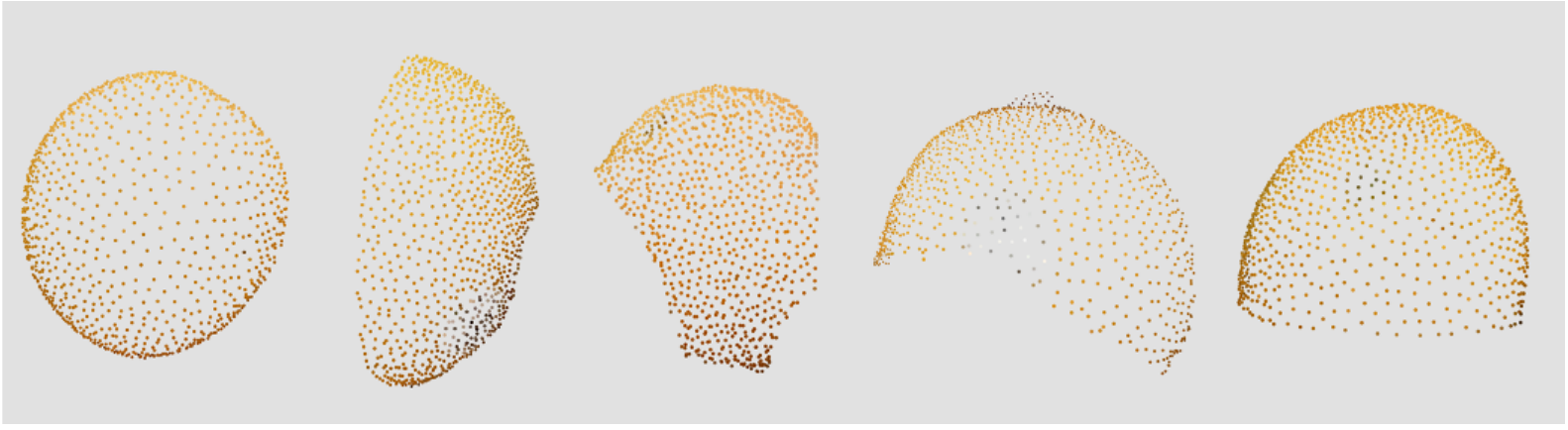}
(f) nectarine
\includegraphics[width=0.95\linewidth,trim={0mm 4mm 1mm 8mm},clip]{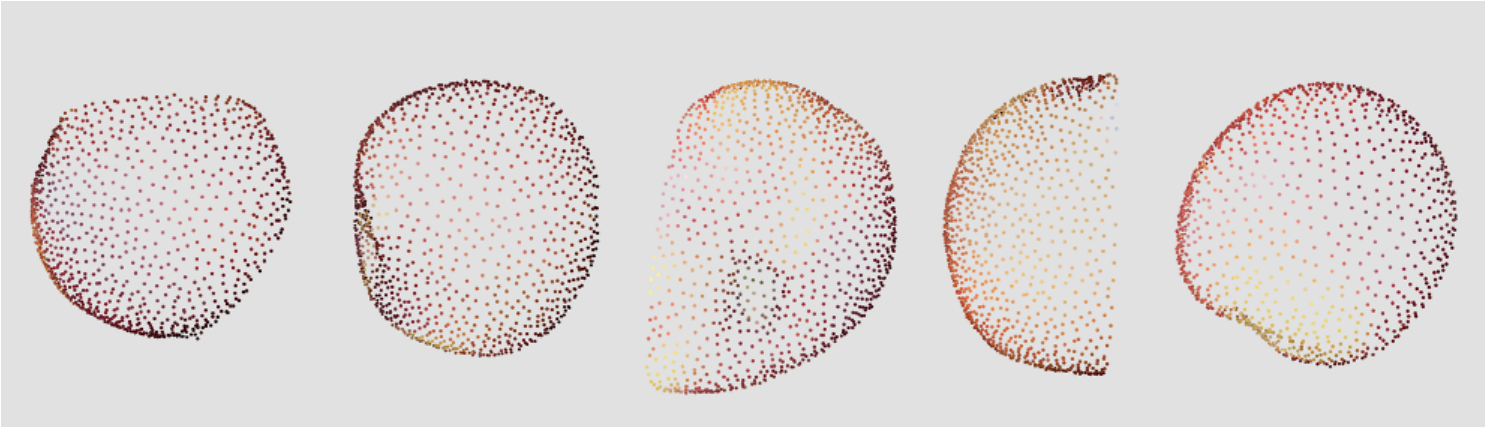}
(g) papaya
\includegraphics[width=0.95\linewidth,trim={0mm 5mm 3mm 12mm},clip]{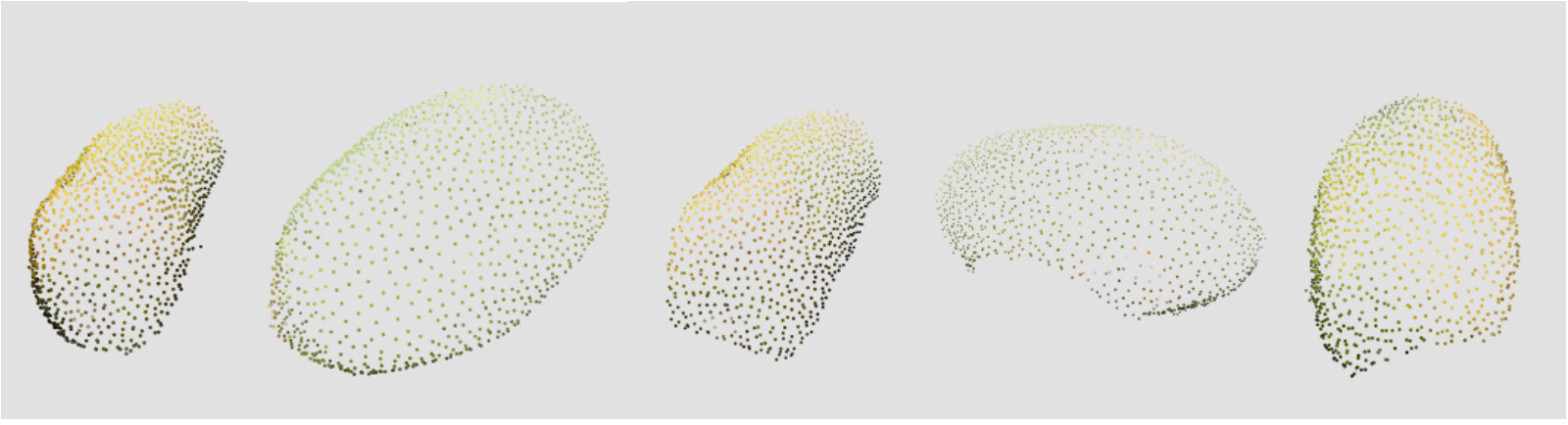}
(h) peach
\includegraphics[width=0.95\linewidth,trim={0mm 5mm 5mm 6mm},clip]{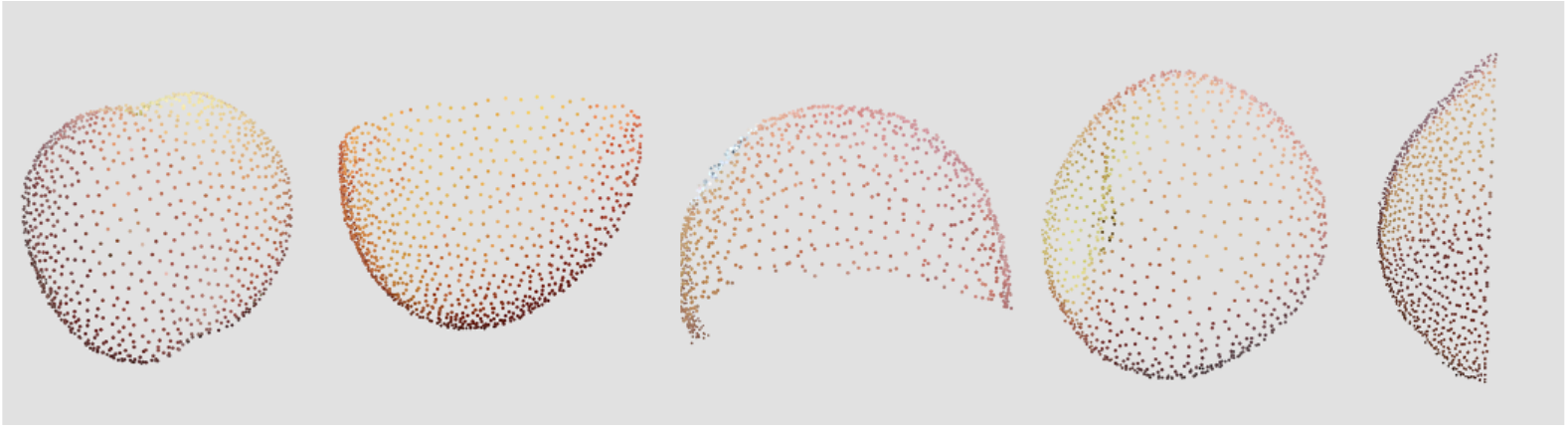}

(i) pear-bosc
\includegraphics[width=0.95\linewidth,trim={0mm 6mm 3mm 12mm},clip]{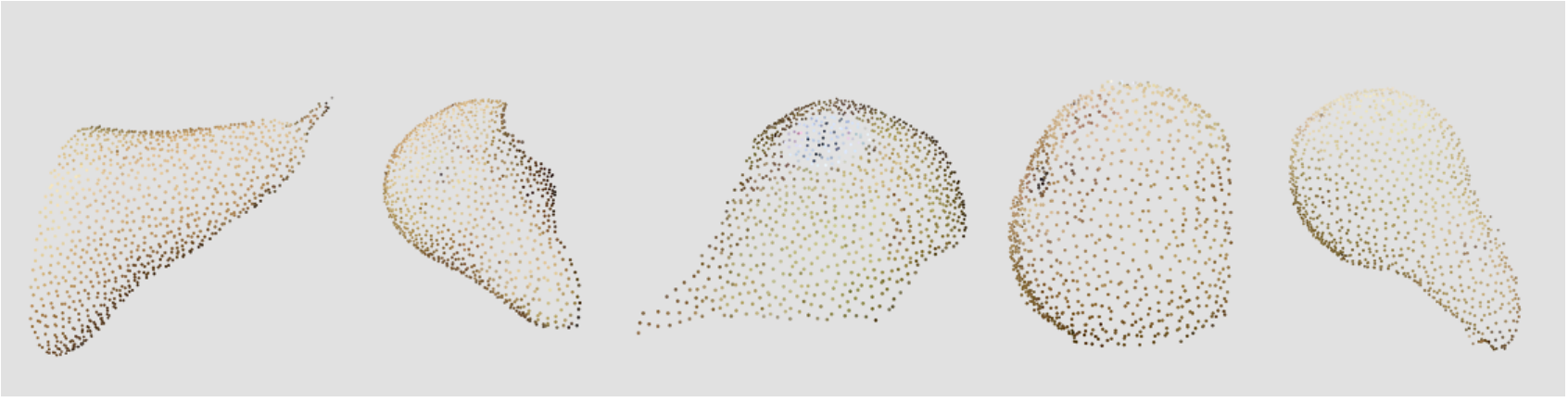}

(j) pineapple
\includegraphics[width=0.95\linewidth,trim={0mm 7mm 3mm 5mm},clip]{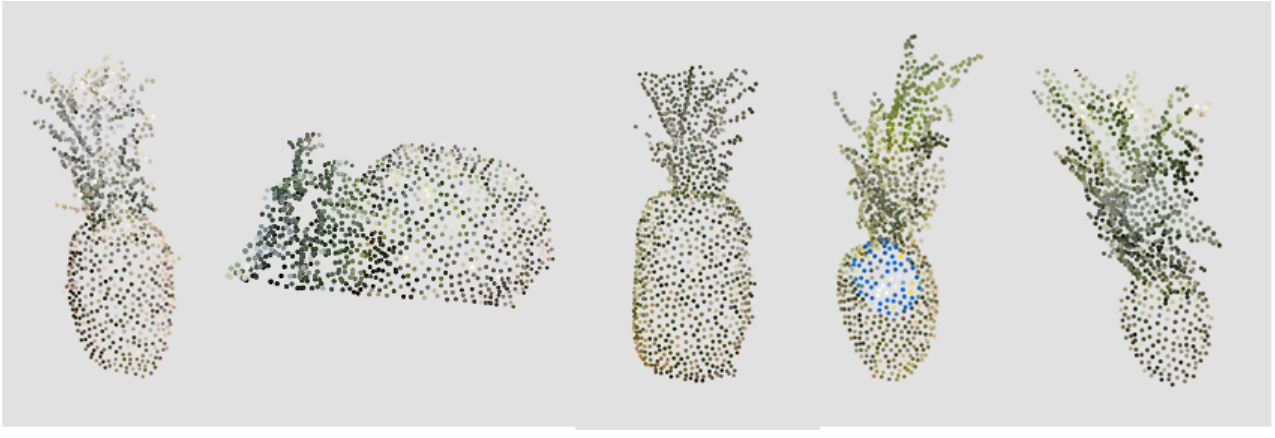}
\caption{Visual samples (1024 points) from 10 of 24 non-apple fruit classes. Labels on top of the objects. Zoom in for better visibility.}
\label{Fig12}
\end{figure}

\begin{figure}[H]
\centering
(a) plums
\includegraphics[width=0.95\linewidth,trim={0mm 6mm 2mm 8mm},clip]{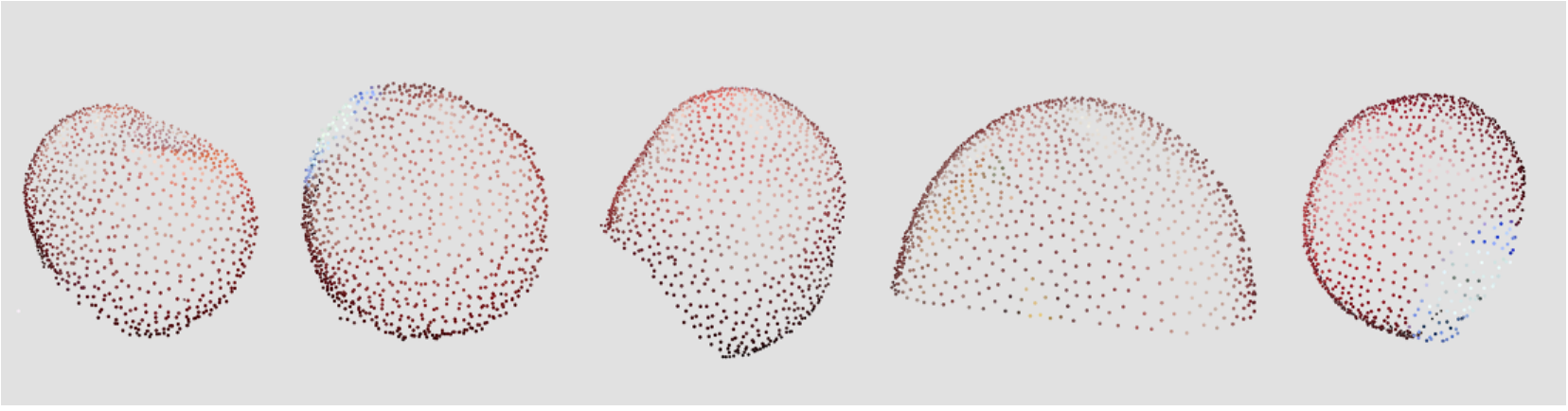}
(b) pomegranate
\includegraphics[width=0.95\linewidth,trim={2mm 5mm 3mm 10mm},clip]{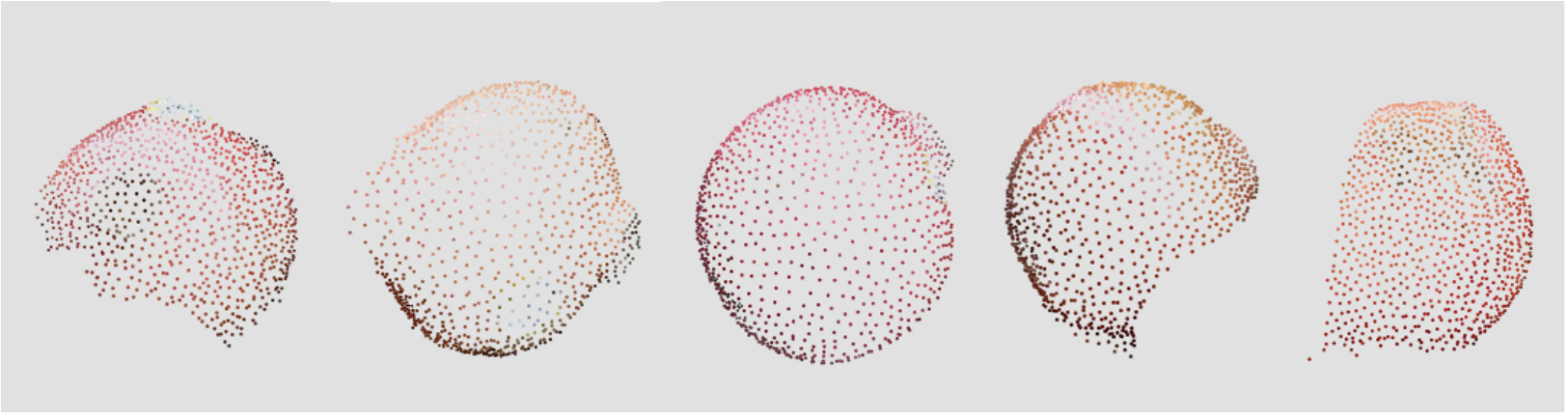}
(c) red-pear
\includegraphics[width=0.95\linewidth,trim={0mm 6mm 2mm 10mm},clip]{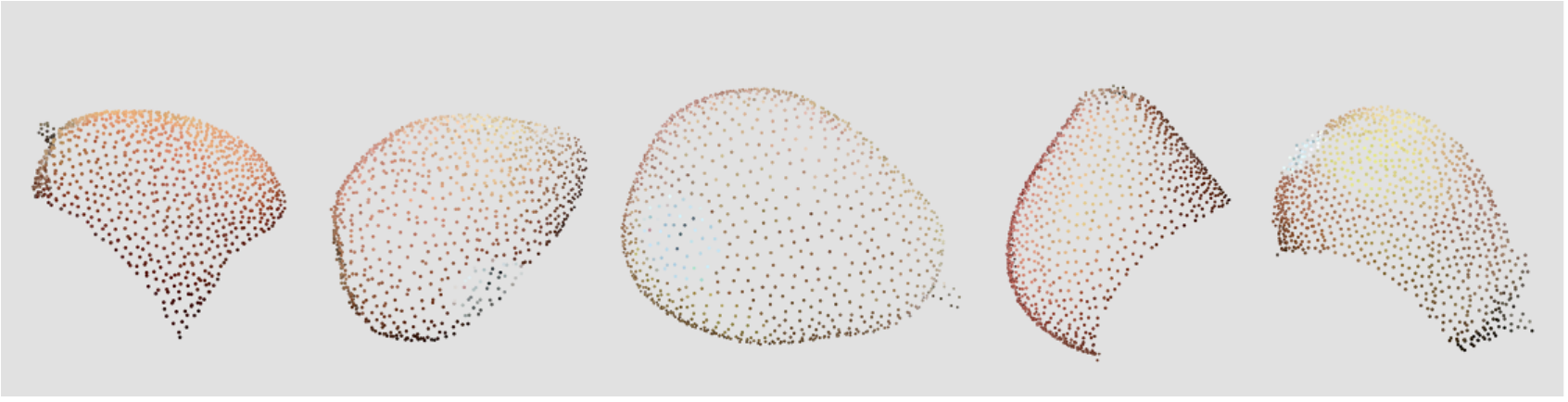}
(d) watermelon
\includegraphics[width=0.95\linewidth,trim={3mm 12mm 4mm 10mm},clip]{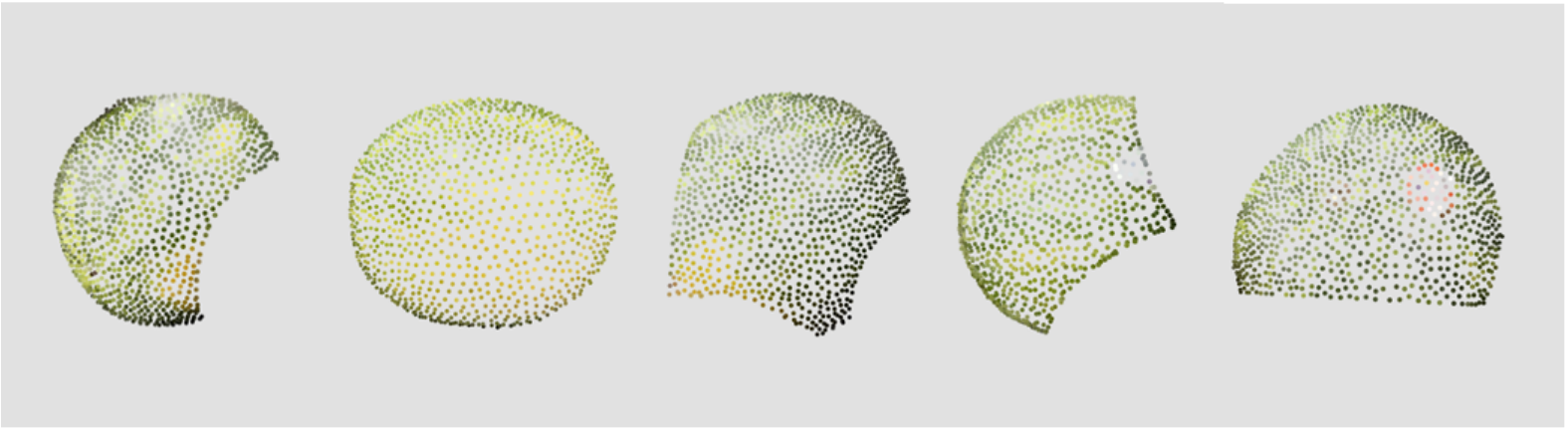}
\caption{Visual samples (1024 points) from 4 or 24 non-apple fruit classes. Labels on top of the objects. Zoom in for better visibility.}
\label{Fig13}
\end{figure}

\begin{figure}[H]
\centering
(a) artichokes
\includegraphics[width=0.95\linewidth,trim={2mm 10mm 0mm 10mm},clip]{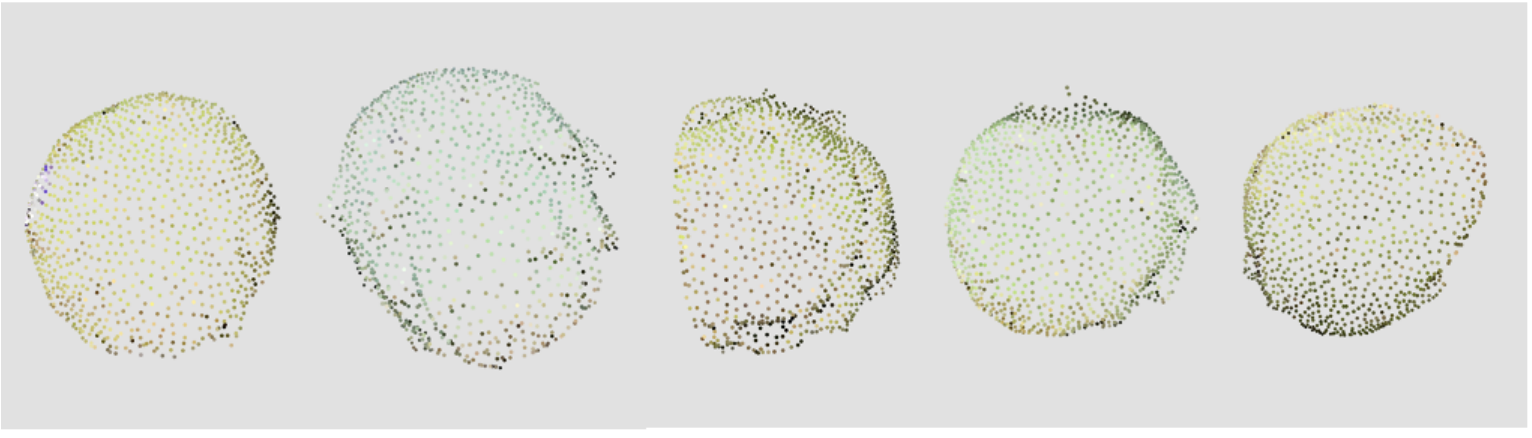}
(b) broccoli
\includegraphics[width=0.95\linewidth,trim={1mm 8mm 0mm 15mm},clip]{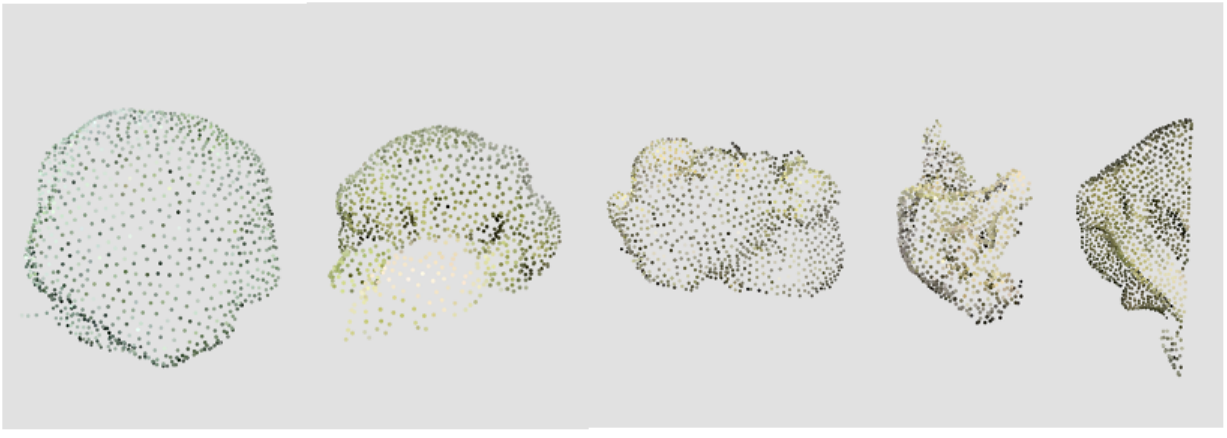}
(c) cucumbers
\includegraphics[width=0.95\linewidth,trim={3mm 10mm 0mm 12mm},clip]{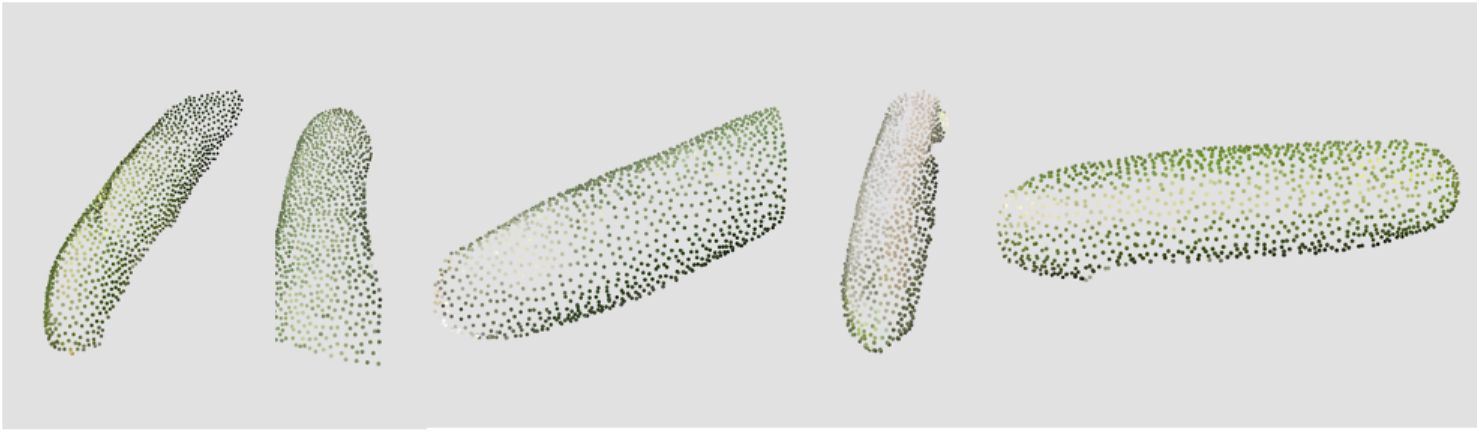}
(d) eggplant
\includegraphics[width=0.95\linewidth,trim={1mm 2mm 0mm 10mm},clip]{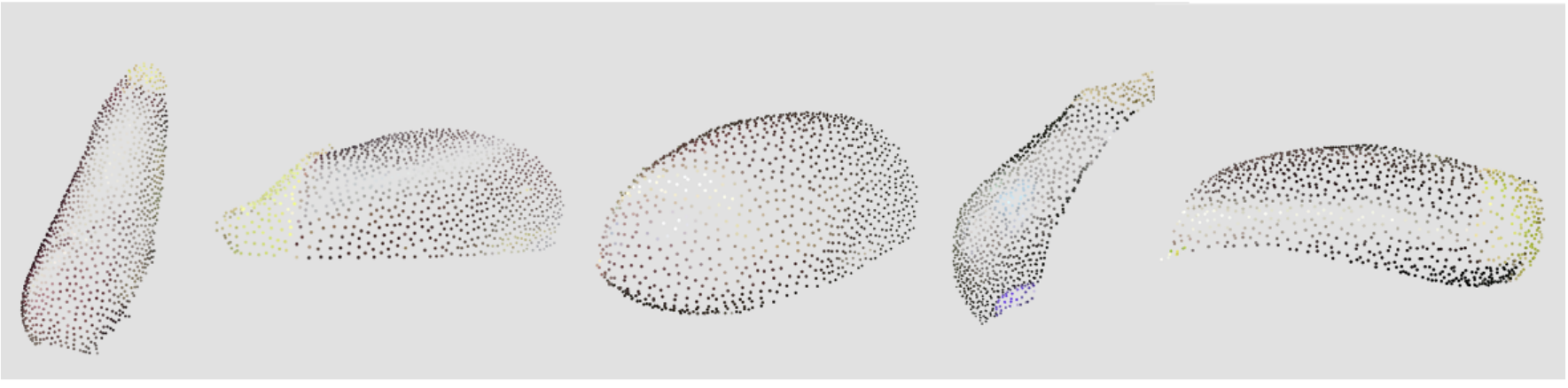}
(e) garlic
\includegraphics[width=0.95\linewidth,trim={2mm 10mm 0mm 12mm},clip]{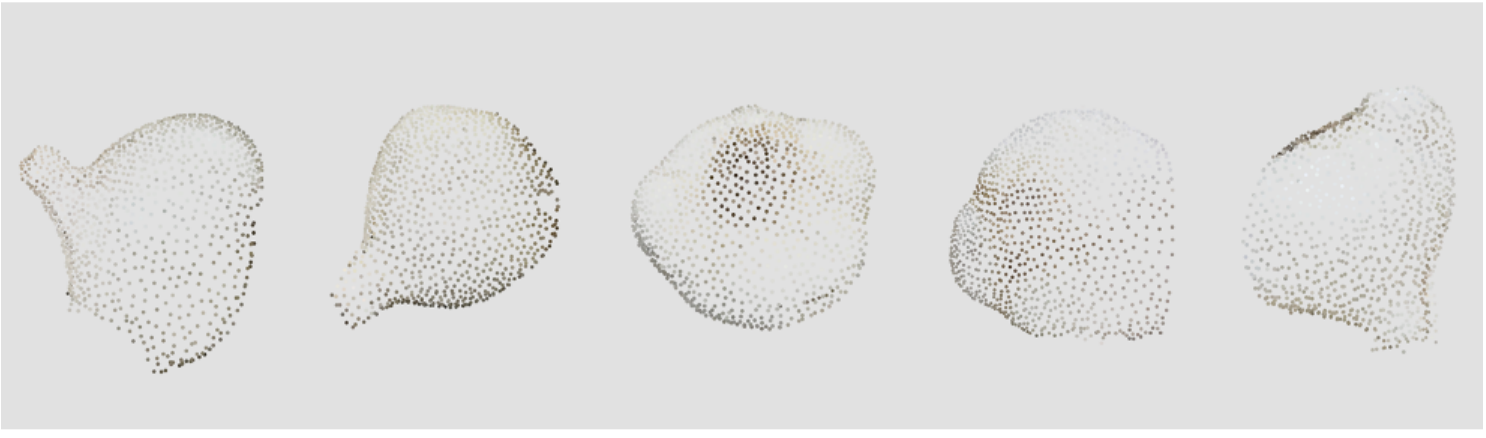}
(f) ginger
\includegraphics[width=0.95\linewidth,trim={2mm 12mm 0mm 14mm},clip]{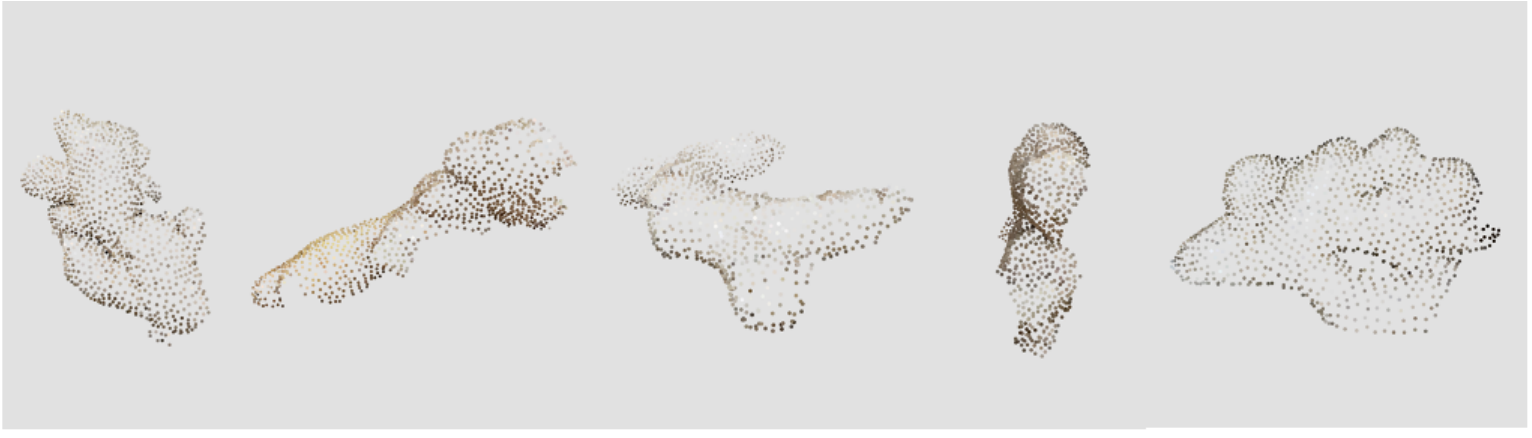}
(g) green-bell-pepper
\includegraphics[width=0.95\linewidth,trim={1mm 11mm 0mm 11mm},clip]{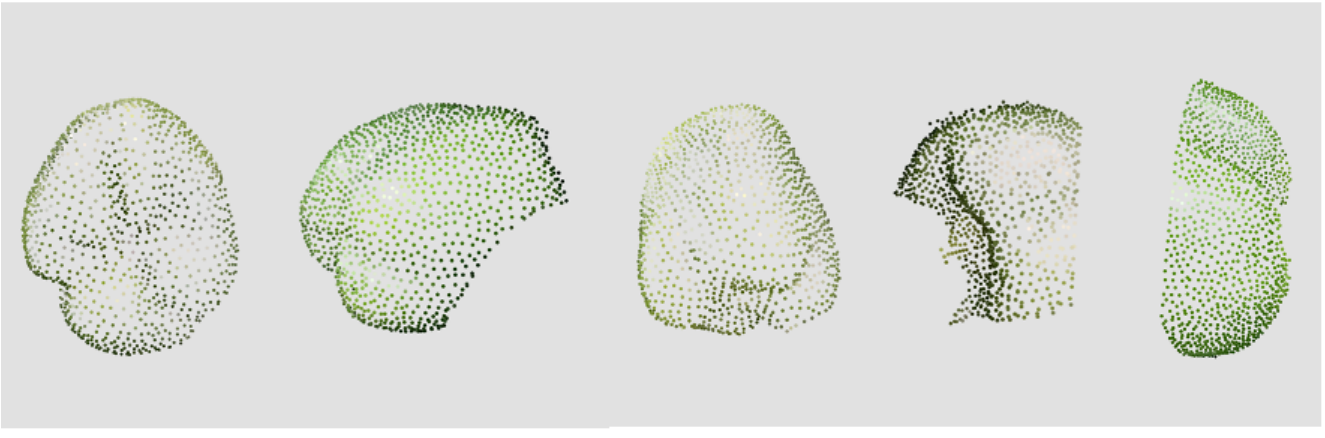}
(h) green-cabbage
\includegraphics[width=0.95\linewidth,trim={1mm 12mm 2mm 15mm},clip]{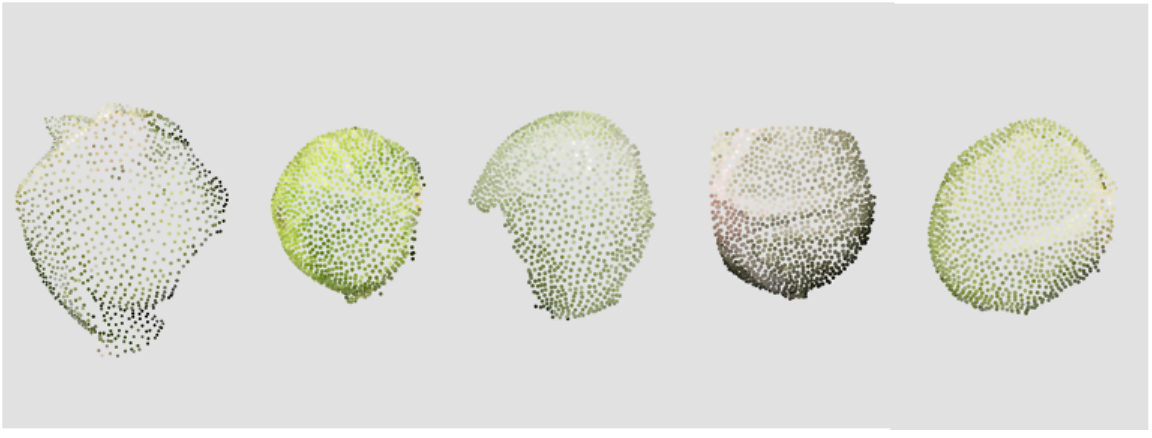}

(i) jalapeno
\includegraphics[width=0.95\linewidth,trim={1mm 10mm 0mm 11mm},clip]{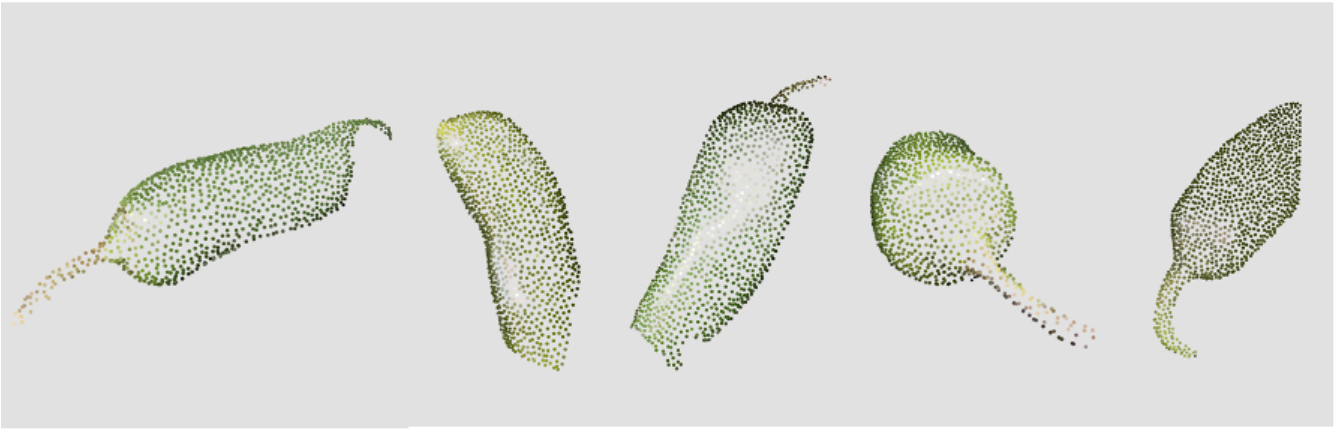}

(j) orange-bell-pepper
\includegraphics[width=0.95\linewidth,trim={1mm 12mm 2mm 12mm},clip]{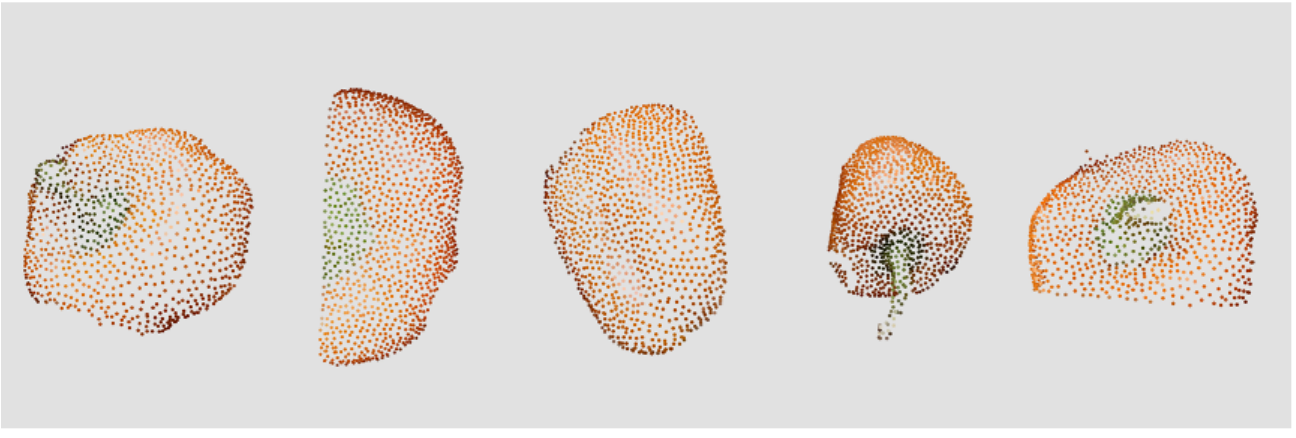}
\caption{Visual samples (1024 points) from 10 vegetable classes. Labels on top of the objects. Zoom in for better visibility.}
\label{Fig14}
\end{figure}

\begin{figure}[H]
\centering
(a) potato
\includegraphics[width=0.95\linewidth,trim={1mm 12mm 0mm 14mm},clip]{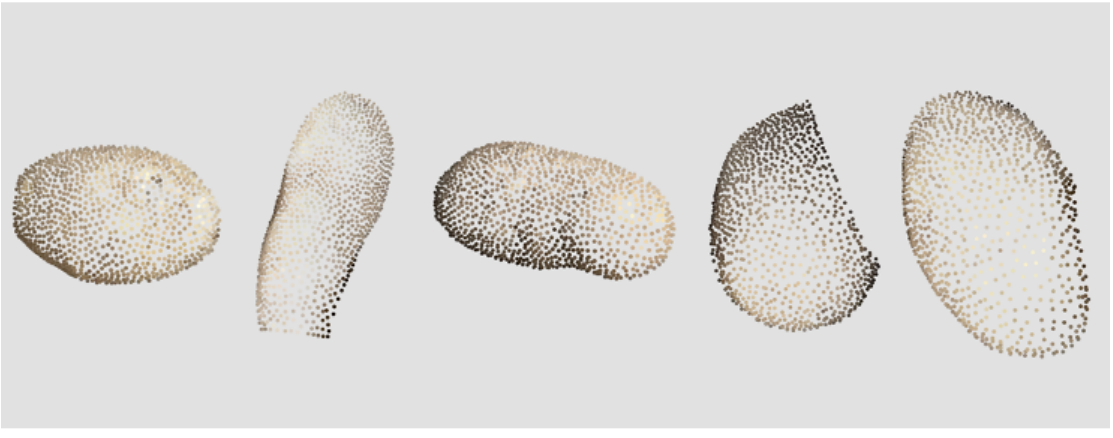}
(b) red-bell-pepper
\includegraphics[width=0.95\linewidth,trim={1mm 12mm 2mm 12mm},clip]{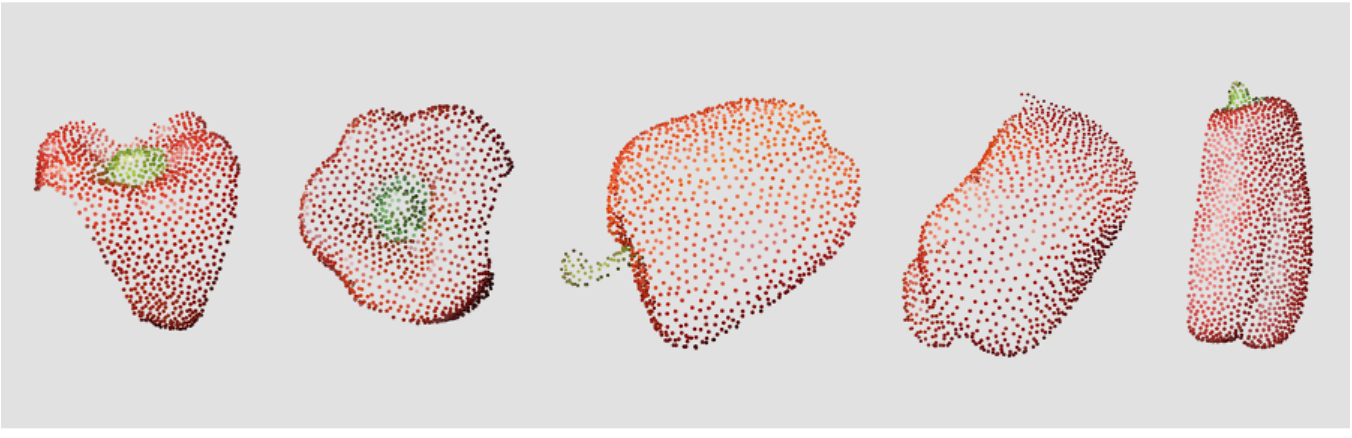}
(c) red-cabbage
\includegraphics[width=0.95\linewidth,trim={1mm 10mm 2mm 15mm},clip]{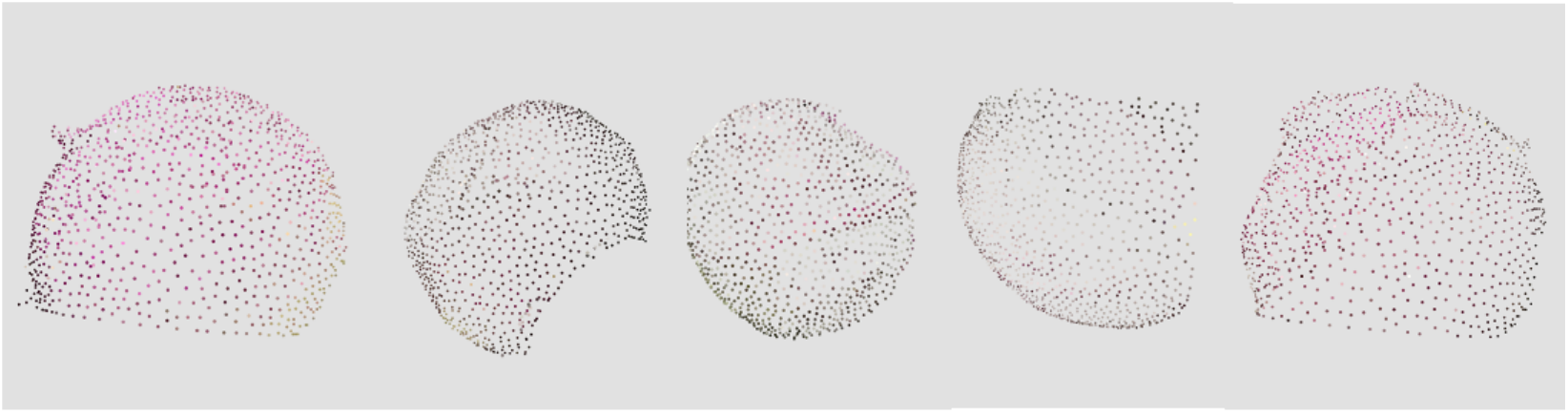}
(d) red-potato
\includegraphics[width=0.95\linewidth,trim={1mm 16mm 1mm 15mm},clip]{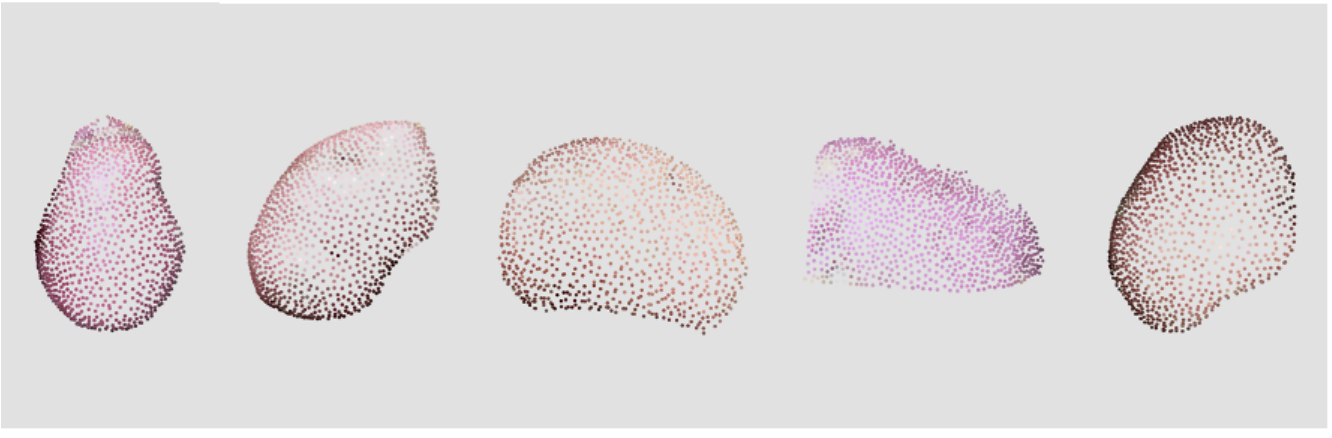}
(e) red-onion
\includegraphics[width=0.95\linewidth,trim={1mm 14mm 0mm 16mm},clip]{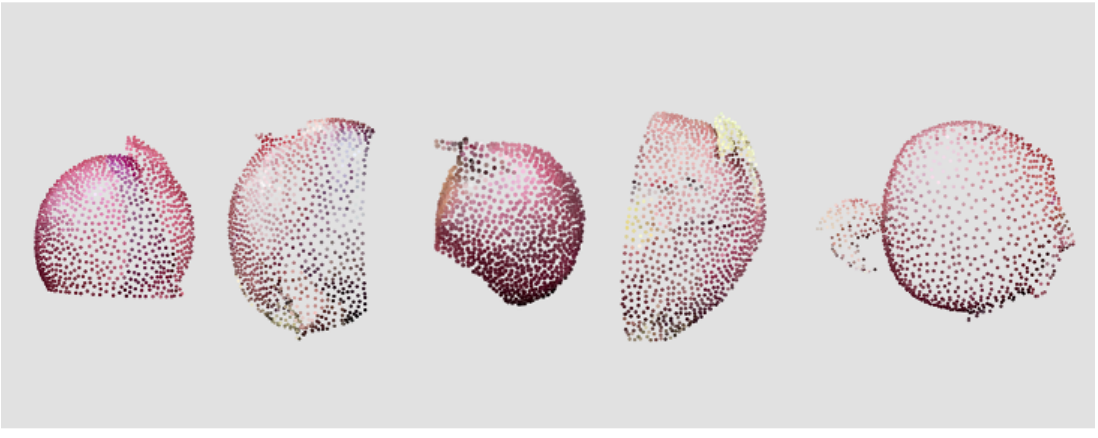}
(f) rutabagas
\includegraphics[width=0.95\linewidth,trim={1mm 12mm 0mm 12mm},clip]{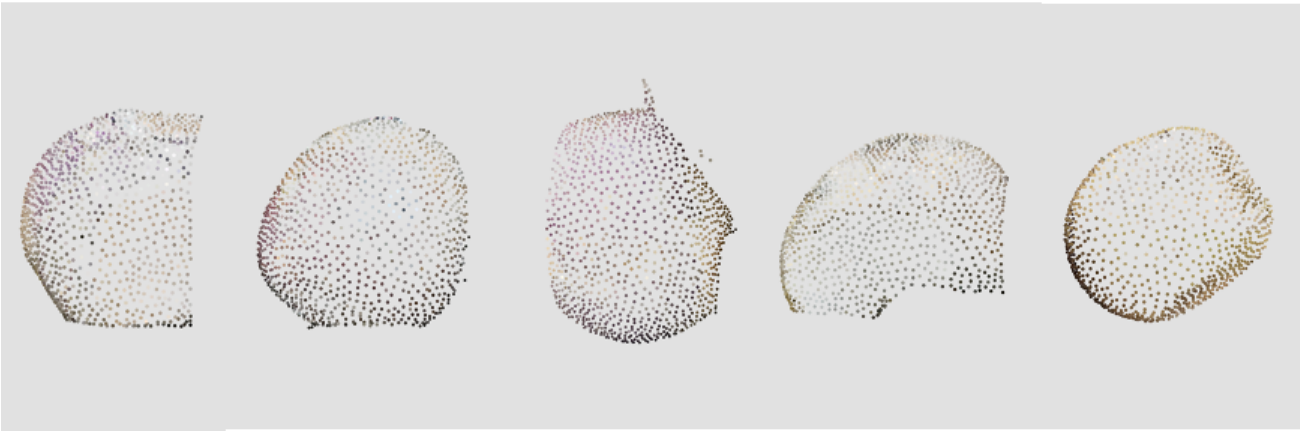}
(g) squash-acorn
\includegraphics[width=0.95\linewidth,trim={1mm 12mm 0mm 15mm},clip]{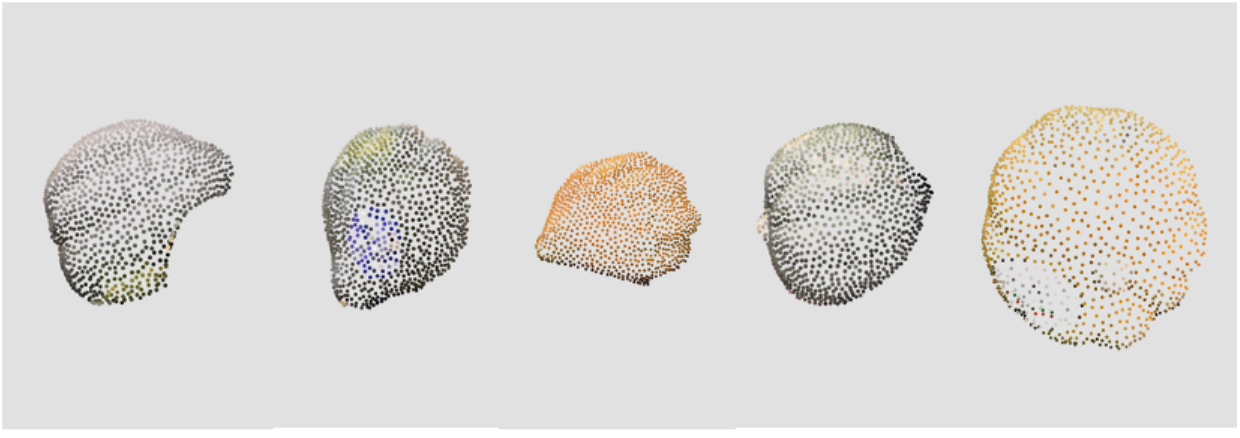}
(h) squash-butternut
\includegraphics[width=0.95\linewidth,trim={1mm 8mm 0mm 14mm},clip]{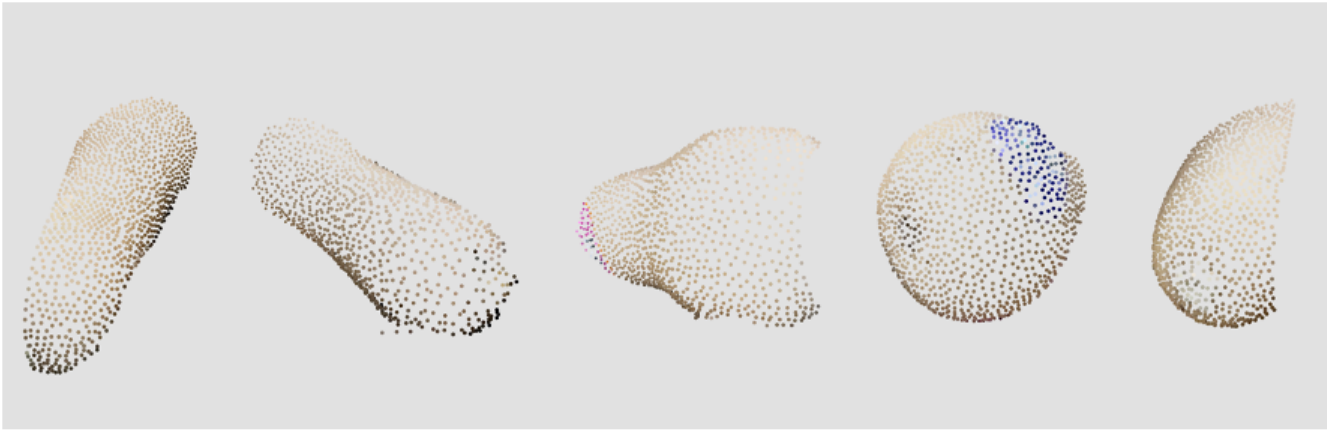}

(i) squash-spaghetti
\includegraphics[width=0.95\linewidth,trim={1mm 12mm 0mm 15mm},clip]{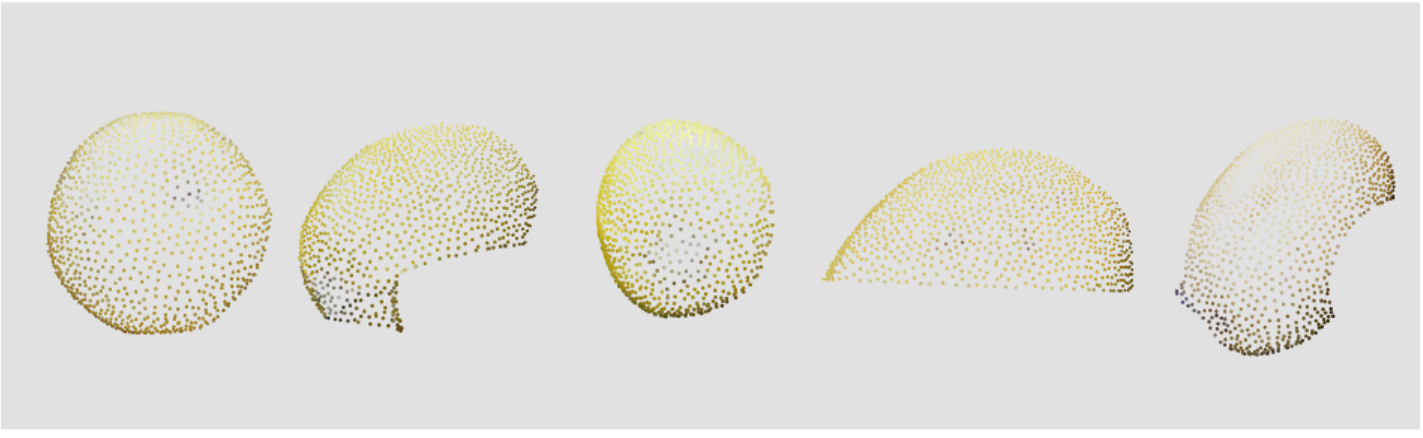}

(j) squash-yellow
\includegraphics[width=0.95\linewidth,trim={0mm 6mm 0mm 12mm},clip]{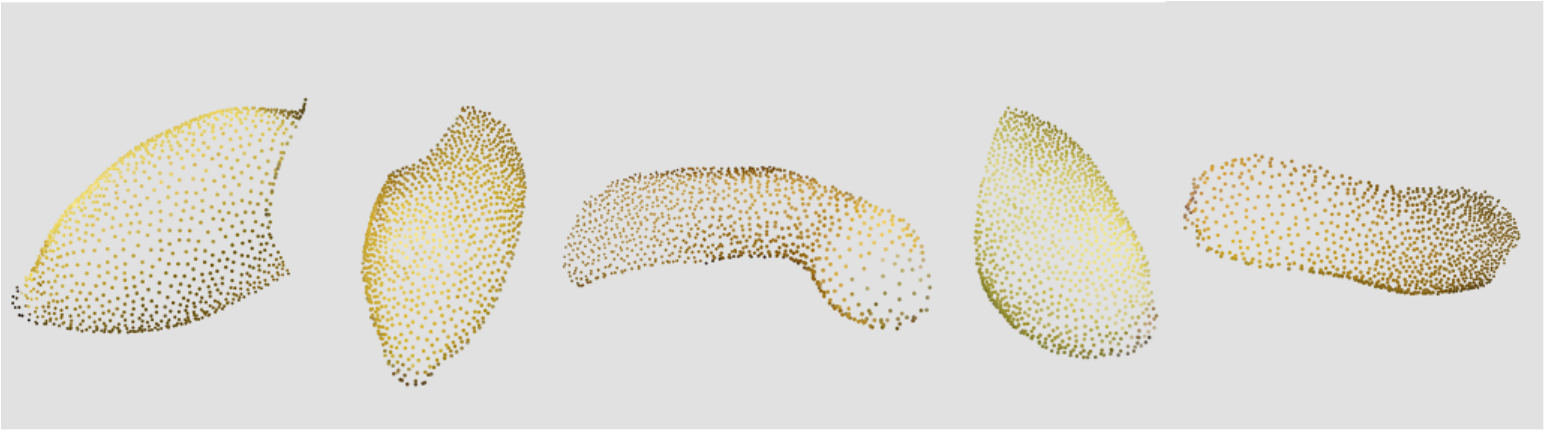}
\caption{Visual samples (1024 points) from 10 vegetable classes. Labels on top of the objects. Zoom in for better visibility.}
\label{Fig15}
\end{figure}

\begin{figure}[H]
\centering
(a) sweet-potato
\includegraphics[width=0.95\linewidth,trim={1mm 12mm 0mm 13mm},clip]{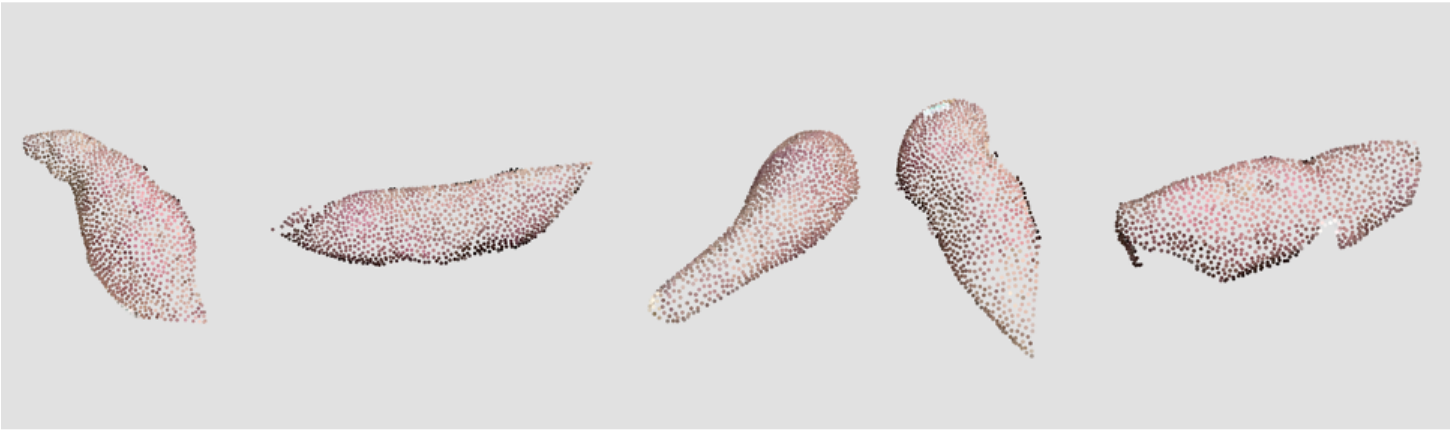}
(b) tomato
\includegraphics[width=0.95\linewidth,trim={2mm 14mm 0mm 17mm},clip]{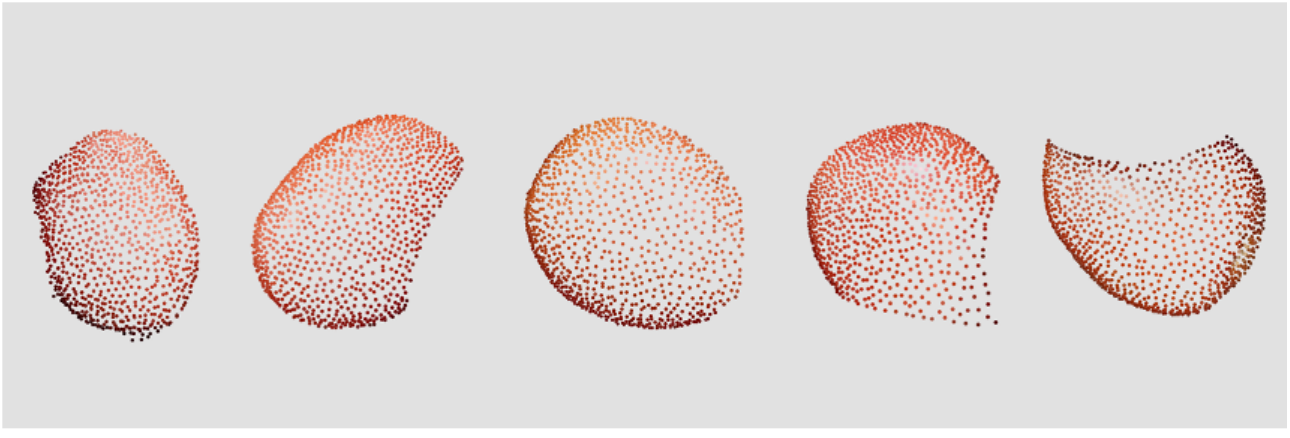}
(c) turnip
\includegraphics[width=0.95\linewidth,trim={0mm 13mm 0mm 14mm},clip]{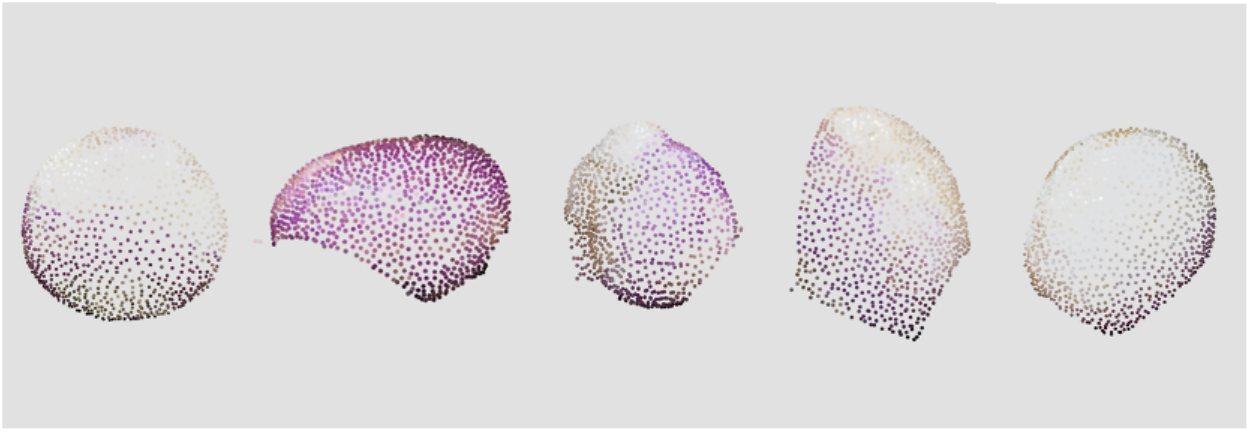}
(d) vine-tomato
\includegraphics[width=0.95\linewidth,trim={0mm 13mm 0mm 13mm},clip]{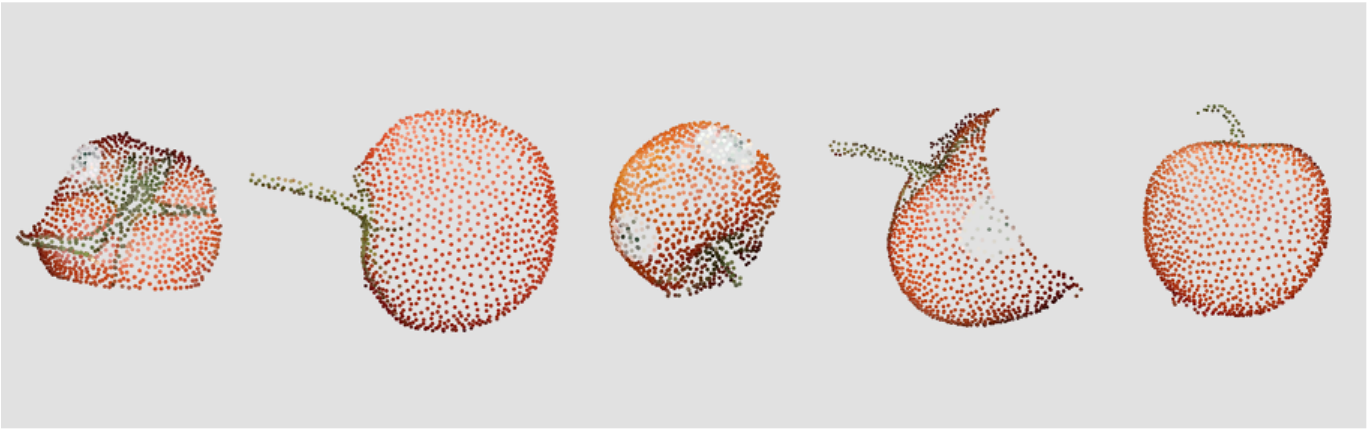}
(e) white-onion
\includegraphics[width=0.95\linewidth,trim={0mm 18mm 0mm 18mm},clip]{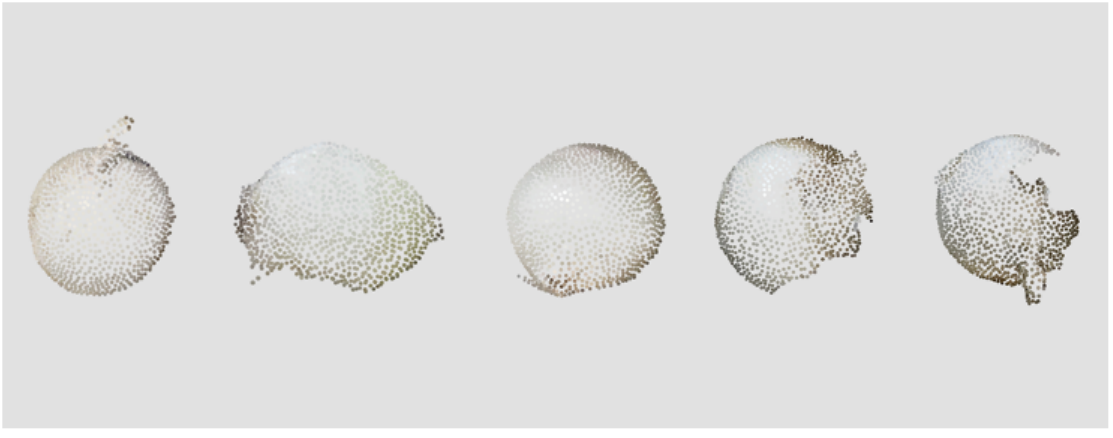}
(f) yam
\includegraphics[width=0.95\linewidth,trim={0mm 10mm 0mm 14mm},clip]{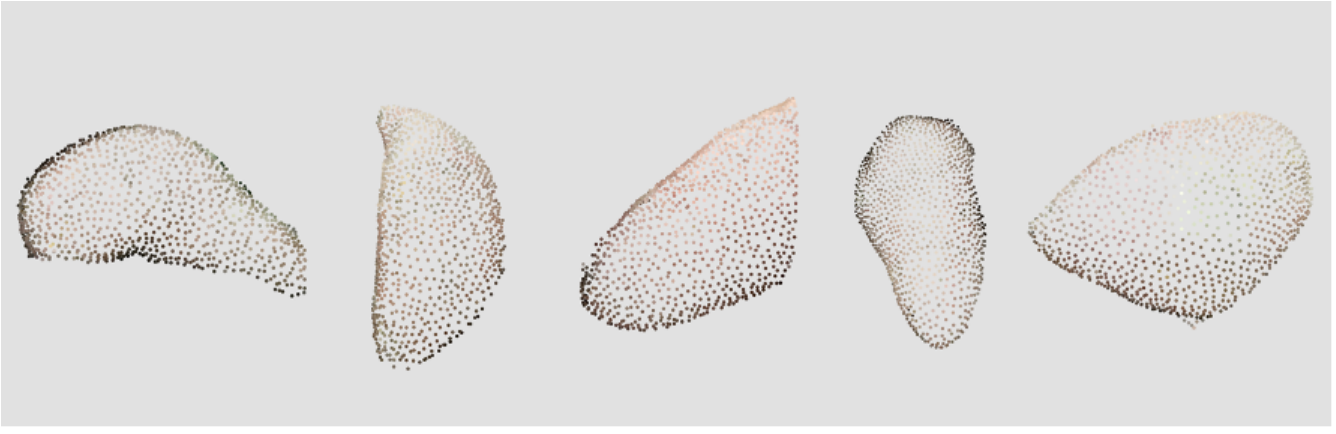}
(g) yellow-bell-pepper
\includegraphics[width=0.95\linewidth,trim={0mm 16mm 0mm 12mm},clip]{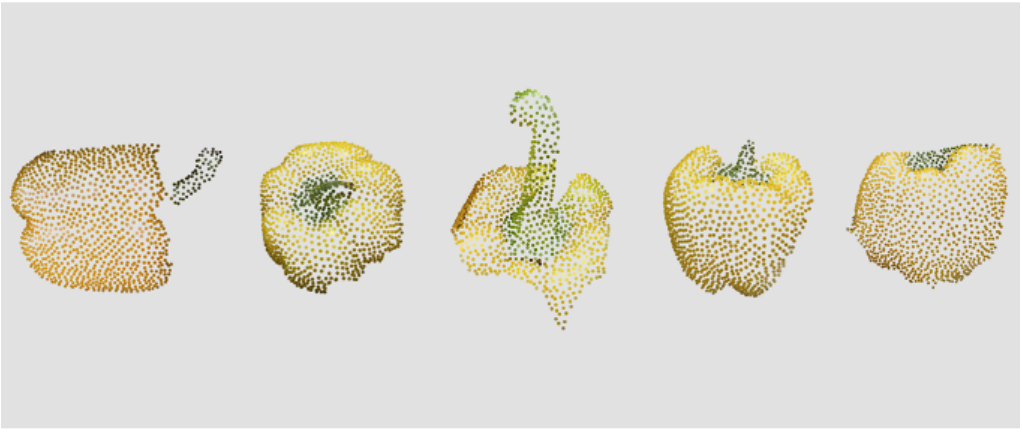}
(h) yellow-onion
\includegraphics[width=0.95\linewidth,trim={0mm 20mm 0mm 20mm},clip]{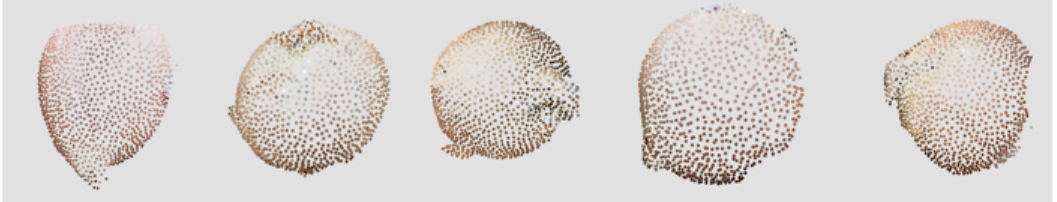}
\caption{Visual samples (1024 points) from 8 vegetable classes. Labels on top of the objects. Zoom in for better visibility.}
\label{Fig16}
\end{figure}

\begin{figure}[H]
\centering
(a) almond-milk
\includegraphics[width=0.95\linewidth,trim={0mm 8mm 0mm 9mm},clip]{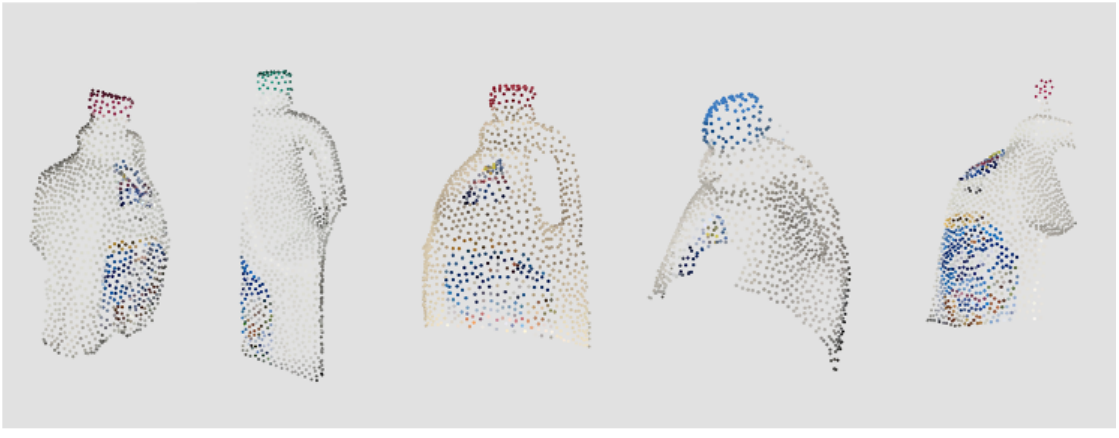}
(b) apple-juice
\includegraphics[width=0.95\linewidth,trim={0mm 8mm 0mm 13mm},clip]{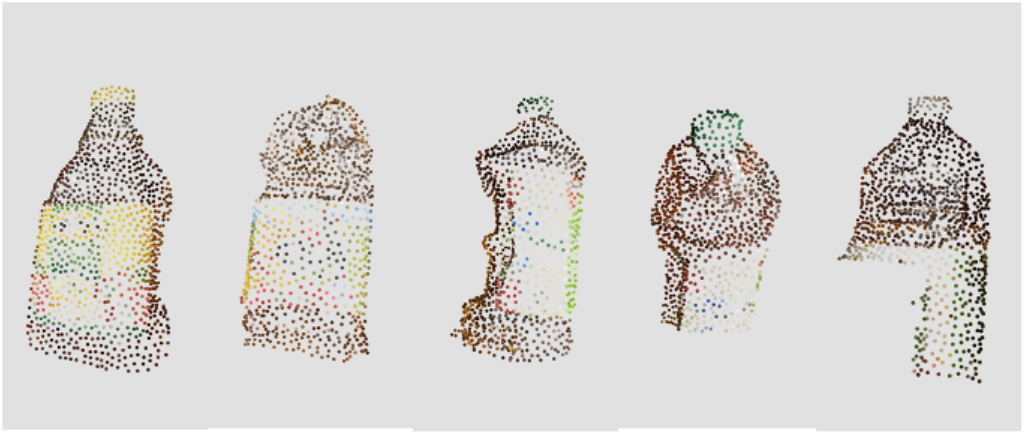}
(c) apple-sauce
\includegraphics[width=0.95\linewidth,trim={0mm 10mm 0mm 10mm},clip]{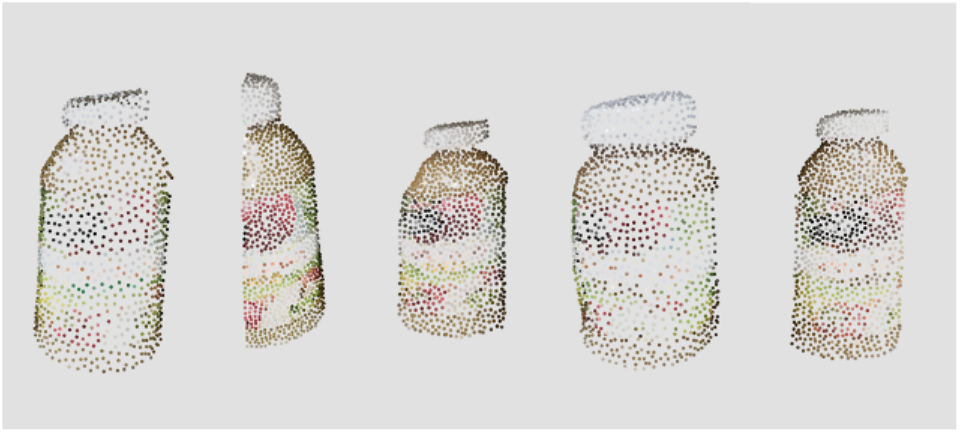}
(d) bagels
\includegraphics[width=0.95\linewidth,trim={0mm 10mm 0mm 12mm},clip]{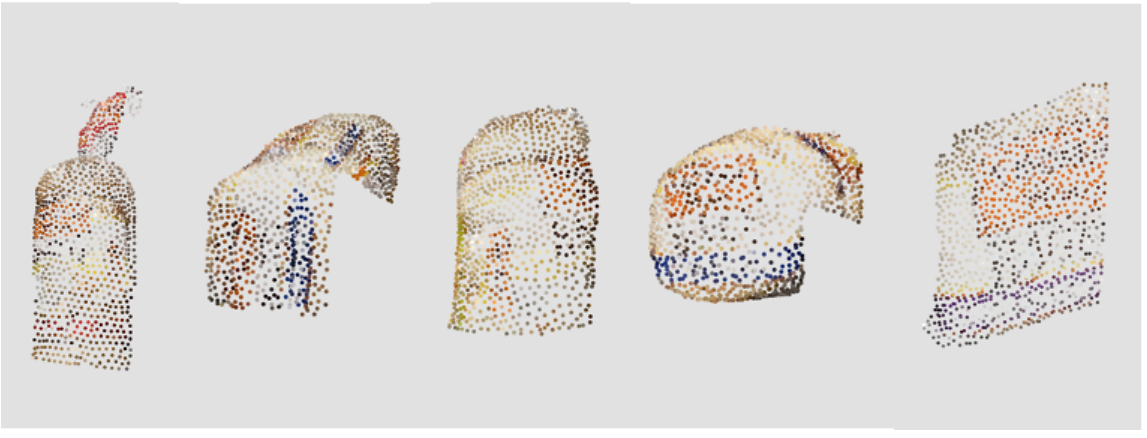}
(e) barbecue-sauce
\includegraphics[width=0.95\linewidth,trim={0mm 8mm 0mm 12mm},clip]{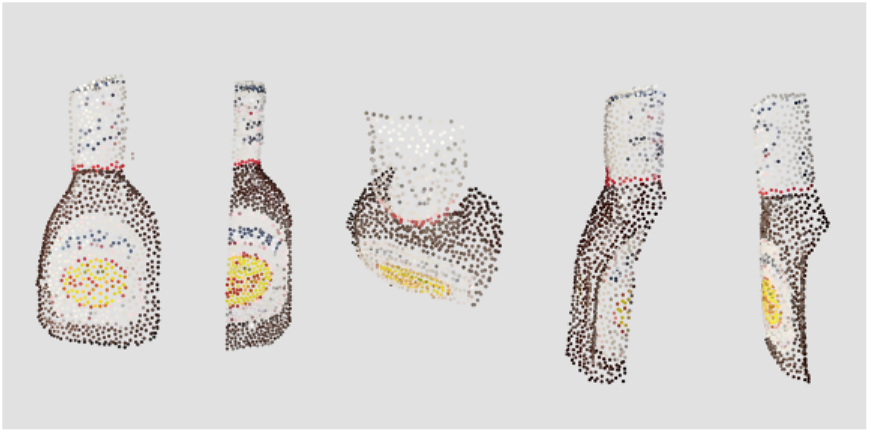}
(f) beans
\includegraphics[width=0.95\linewidth,trim={0mm 13mm 0mm 17mm},clip]{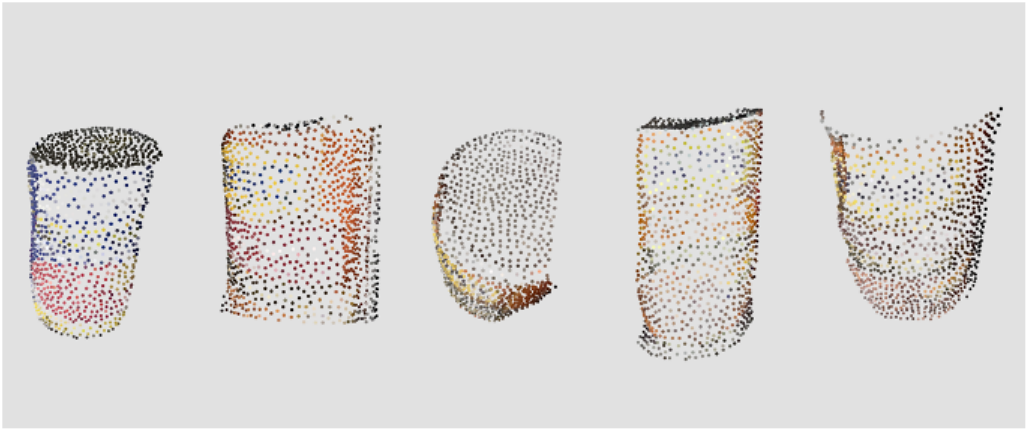}
(g) beans-green
\includegraphics[width=0.95\linewidth,trim={0mm 10mm 0mm 13mm},clip]{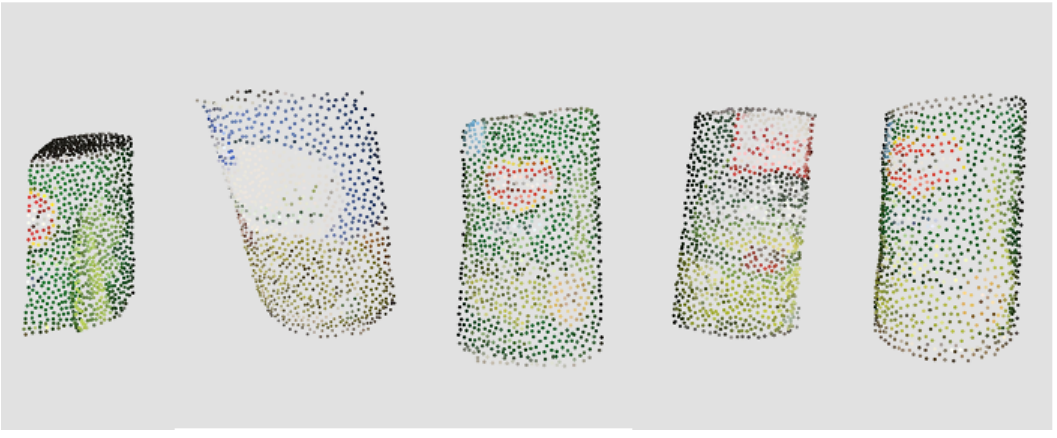}
(h) bread
\includegraphics[width=0.95\linewidth,trim={0mm 12mm 4mm 15mm},clip]{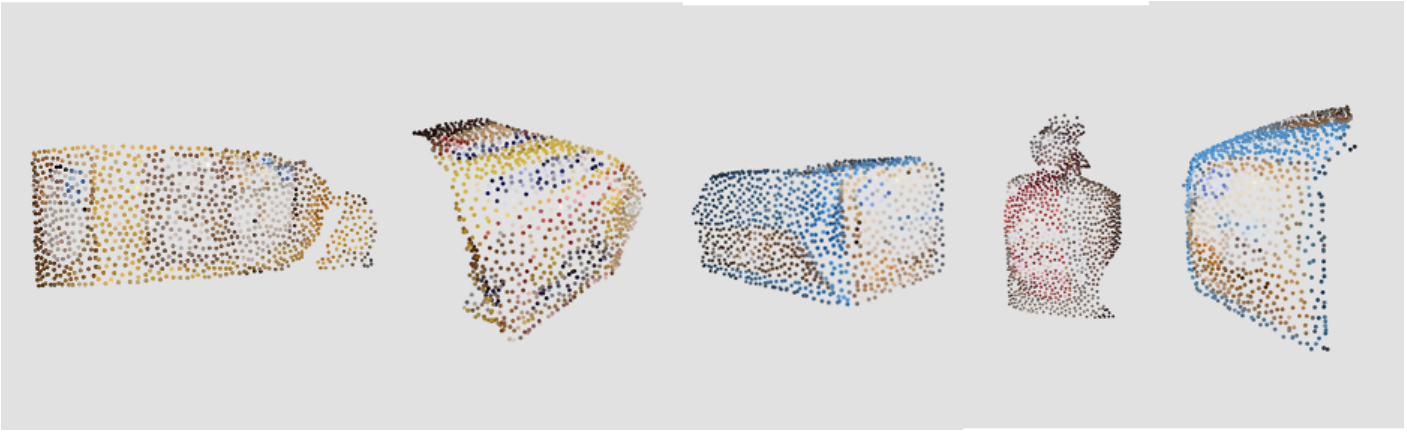}
\caption{Visual samples (1024 points) from 8 package classes. Labels on top of the objects. Zoom in for better visibility.}
\label{Fig17}
\end{figure}

\begin{figure}[H]
\centering
(a) cashews
\includegraphics[width=0.95\linewidth,trim={0mm 9mm 0mm 13mm},clip]{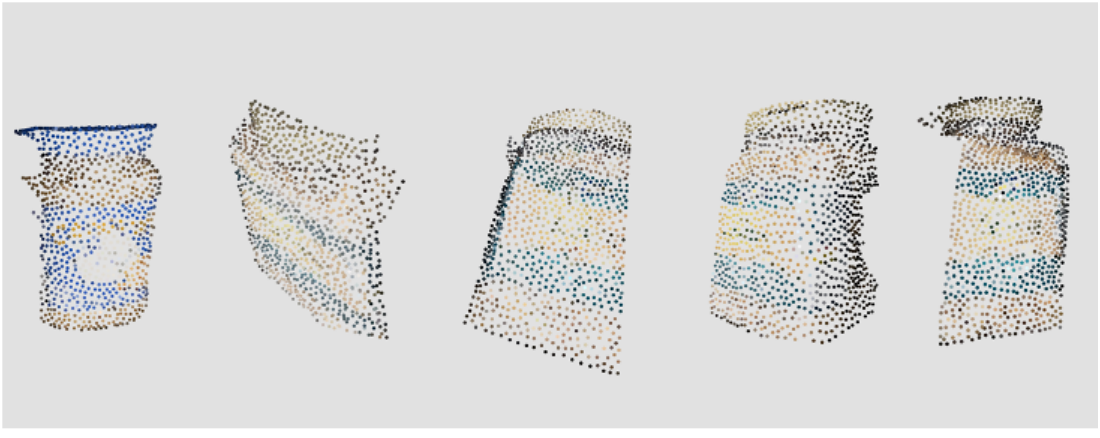}
(b) cheese
\includegraphics[width=0.95\linewidth,trim={0mm 15mm 1mm 13mm},clip]{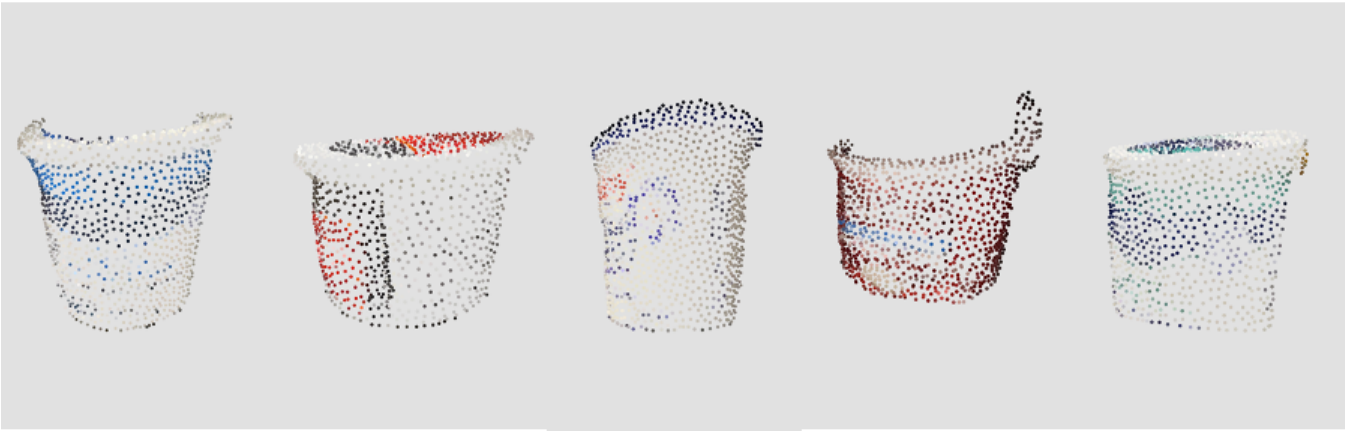}
(c) chips
\includegraphics[width=0.95\linewidth,trim={0mm 5mm 0mm 10mm},clip]{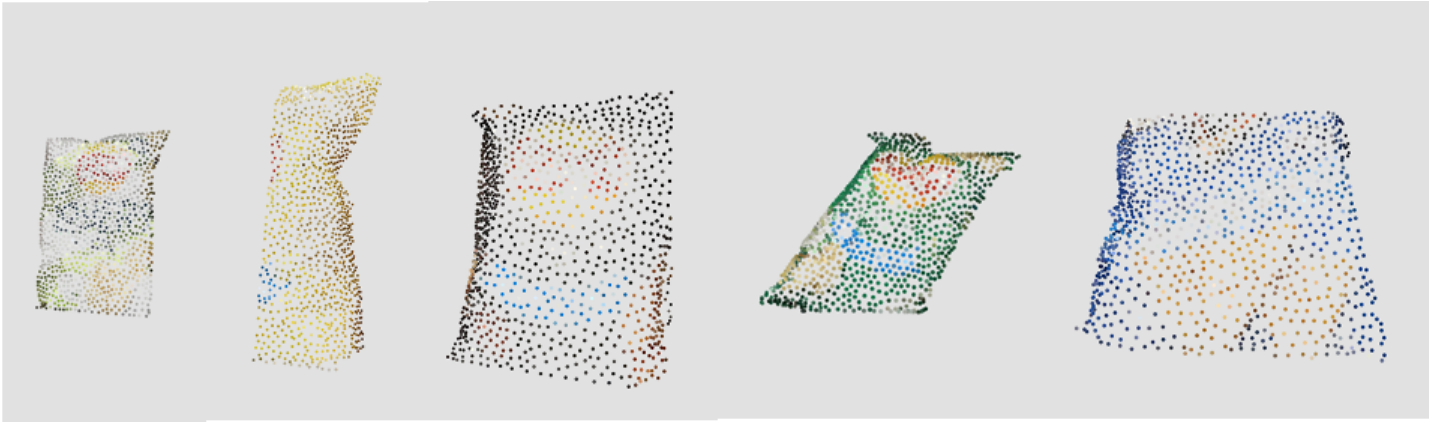}
(d) chocolate-syrup
\includegraphics[width=0.95\linewidth,trim={0mm 6mm 0mm 11mm},clip]{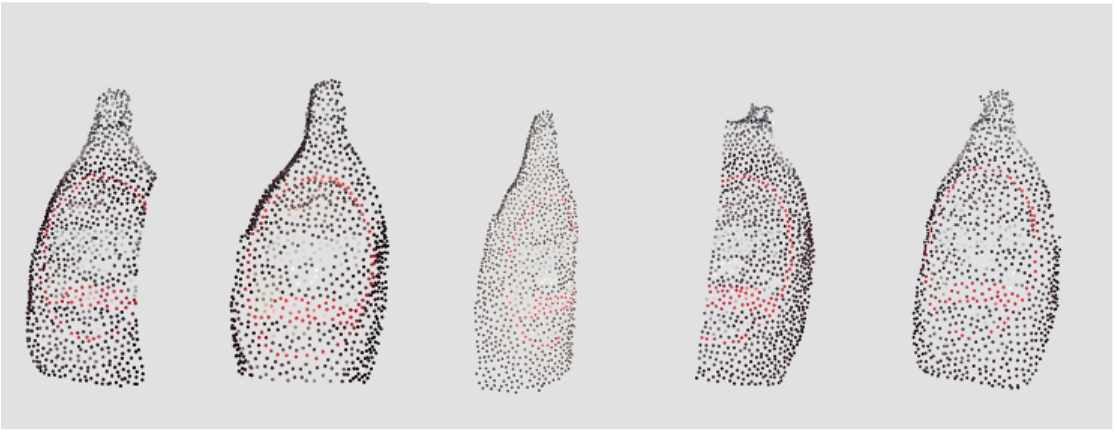}
(e) coconut-water
\includegraphics[width=0.95\linewidth,trim={2mm 11mm 2mm 11mm},clip]{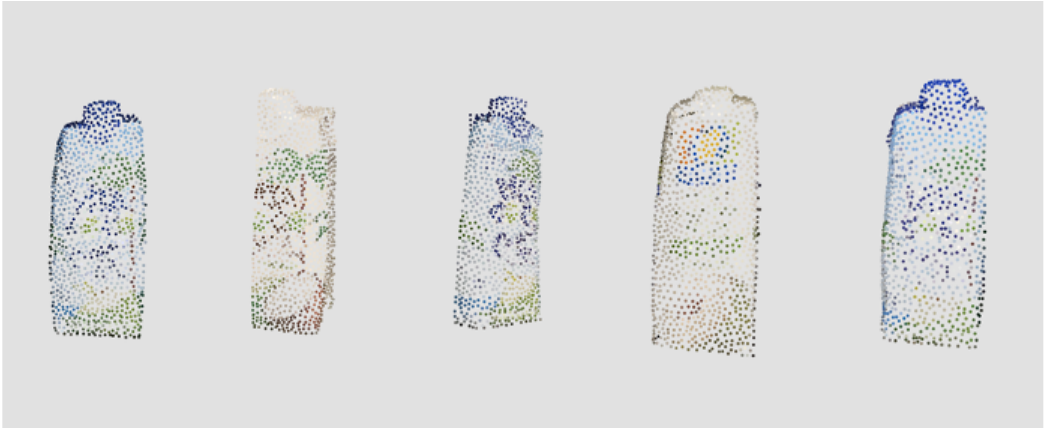}
(f) coffee
\includegraphics[width=0.95\linewidth,trim={0mm 6mm 1mm 8mm},clip]{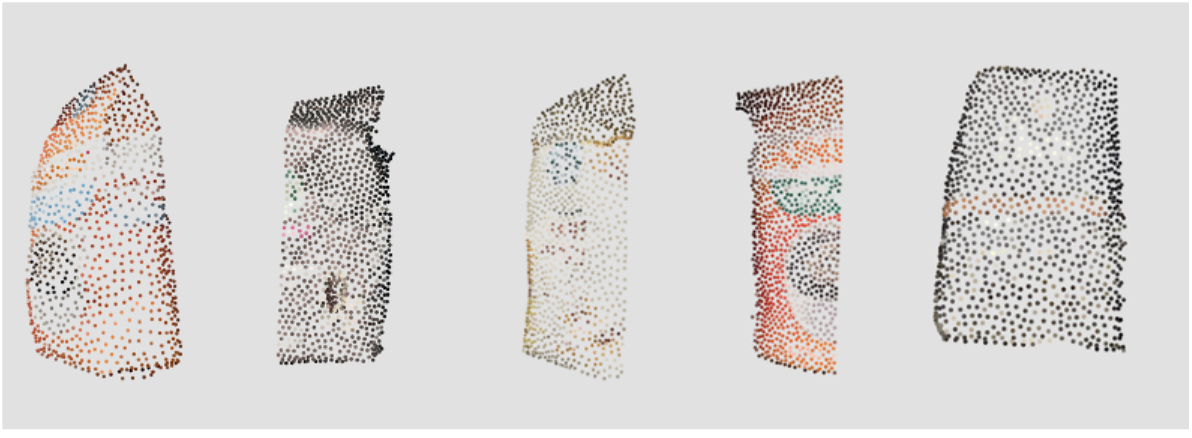}
(g) corn
\includegraphics[width=0.95\linewidth,trim={0mm 6mm 5mm 5mm},clip]{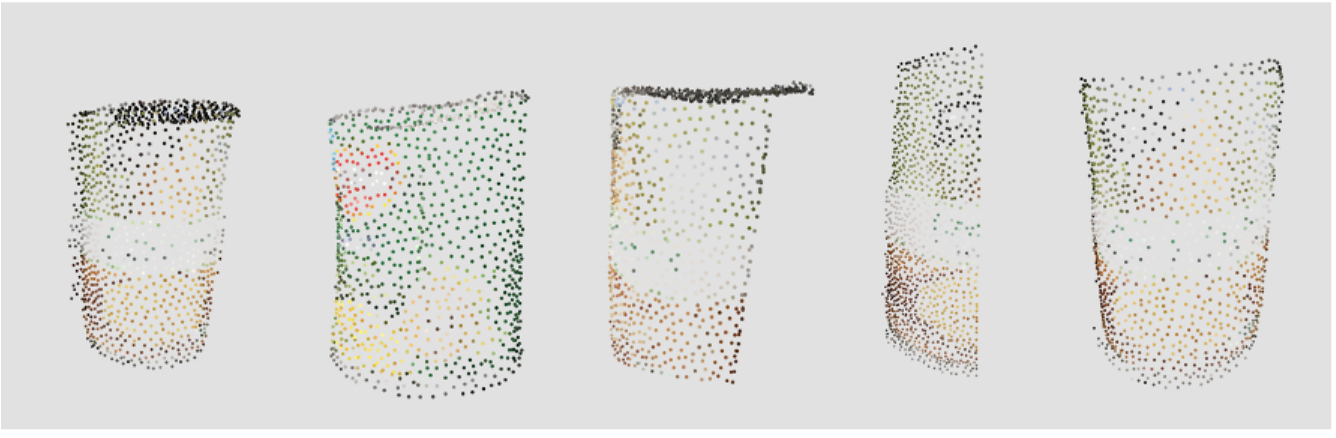}
(h) dip
\includegraphics[width=0.95\linewidth,trim={0mm 9mm 5mm 9mm},clip]{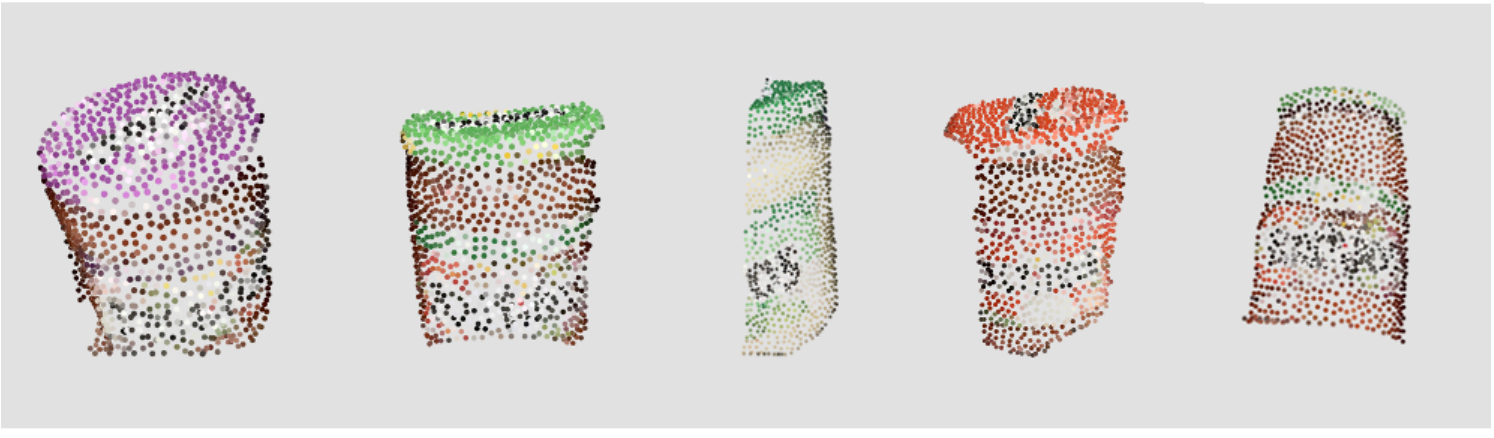}

(i) eggs
\includegraphics[width=0.95\linewidth,trim={0mm 6mm 0mm 8mm},clip]{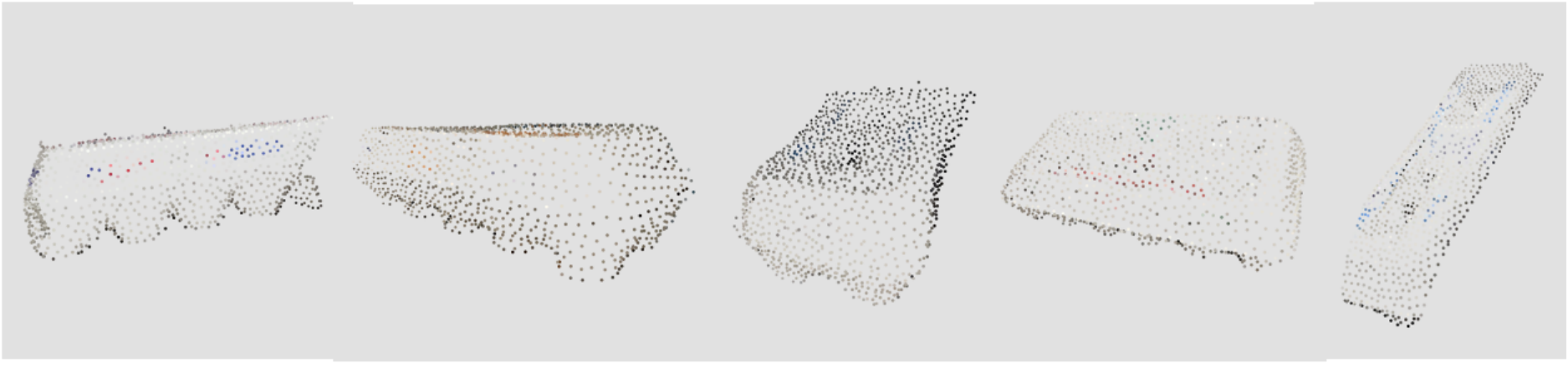}
\caption{Visual samples (1024 points) from 9 package classes. Labels on top of the objects. Zoom in for better visibility.}
\label{Fig18}
\end{figure}

\begin{figure}[H]
\centering
(a) eggs-in-cooler
\includegraphics[width=0.95\linewidth,trim={0mm 12mm 5mm 12mm},clip]{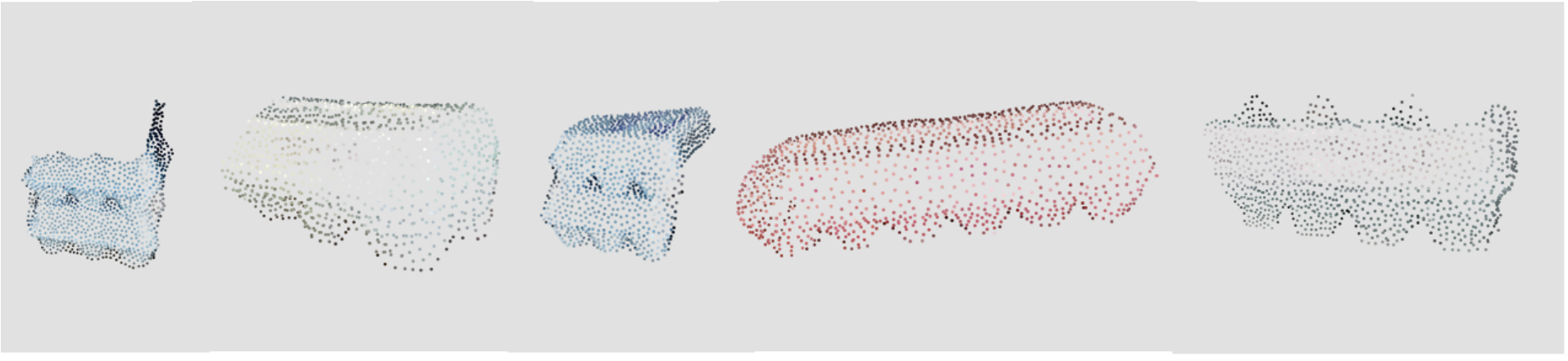}
(b) energy-drink
\includegraphics[width=0.95\linewidth,trim={0mm 7mm 2mm 9mm},clip]{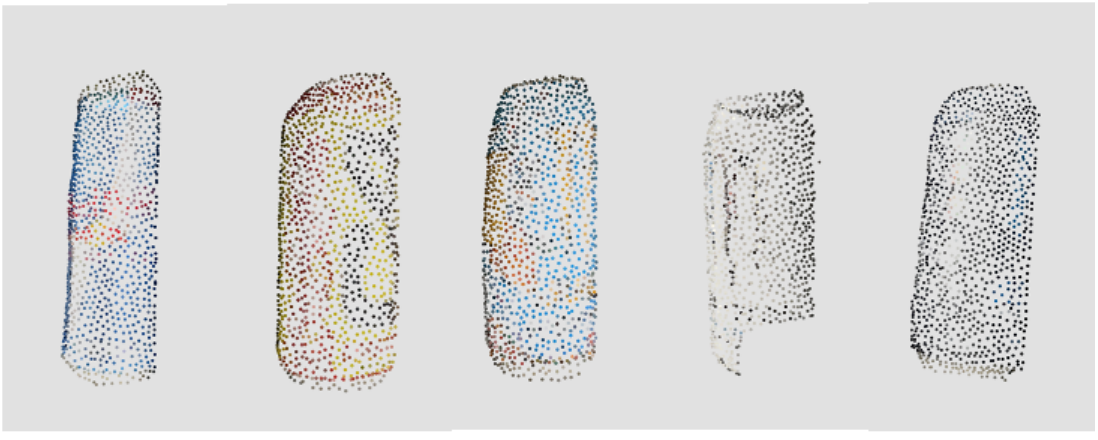}
(c) gatorade
\includegraphics[width=0.95\linewidth,trim={0mm 7mm 2mm 9mm},clip]{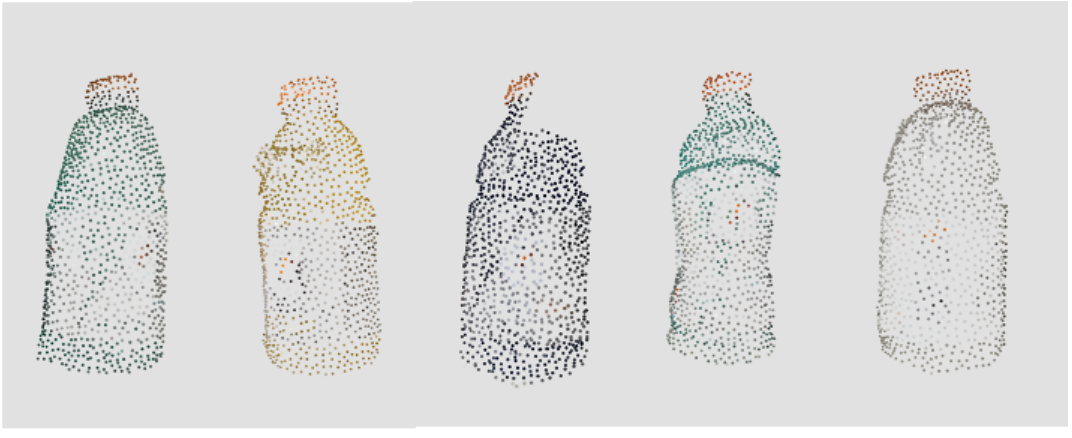}
(d) ham
\includegraphics[width=0.95\linewidth,trim={0mm 16mm 0mm 16mm},clip]{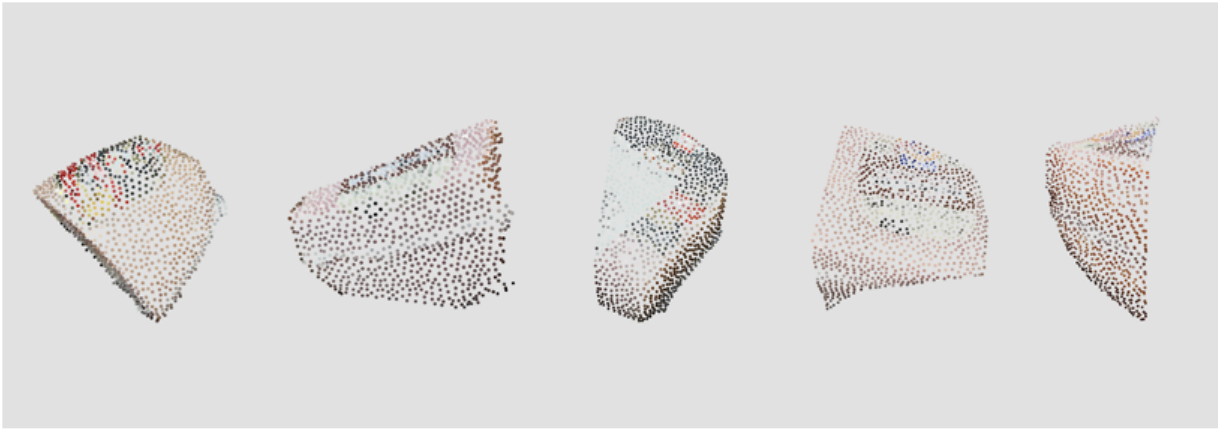}
(e) ice-tea-gallon
\includegraphics[width=0.95\linewidth,trim={0mm 5mm 0mm 8mm},clip]{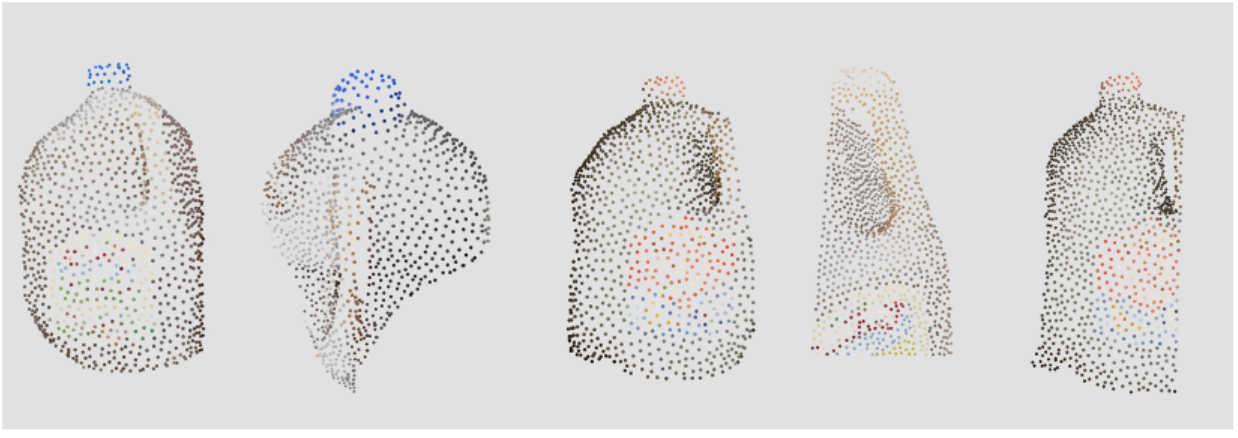}
(f) mayonaise
\includegraphics[width=0.95\linewidth,trim={0mm 6mm 2mm 8mm},clip]{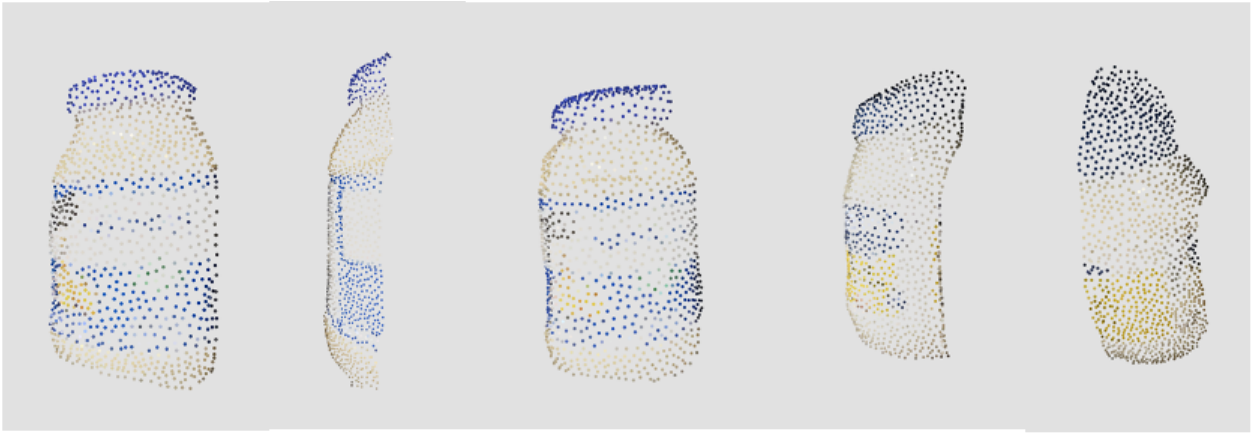}
(g) milk
\includegraphics[width=0.95\linewidth,trim={0mm 6mm 2mm 9mm},clip]{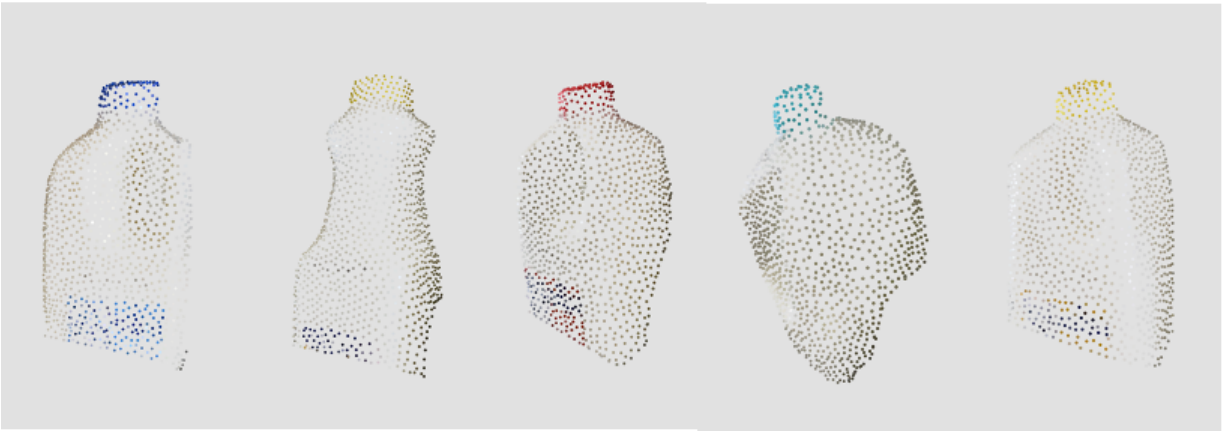}
(h) milk-in-cooler
\includegraphics[width=0.95\linewidth,trim={0mm 4mm 2mm 7mm},clip]{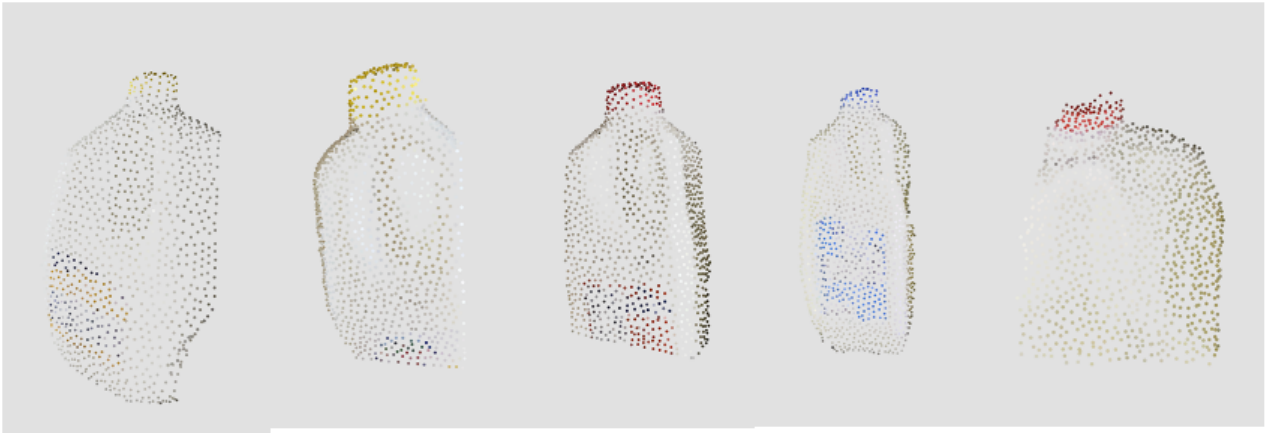}
\caption{Visual samples (1024 points) from 8 package classes. Labels on top of the objects. Zoom in for better visibility.}
\label{Fig19}
\end{figure}

\begin{figure}[H]
\centering
(a) mixed-nuts
\includegraphics[width=0.95\linewidth,trim={0mm 5mm 0mm 4mm},clip]{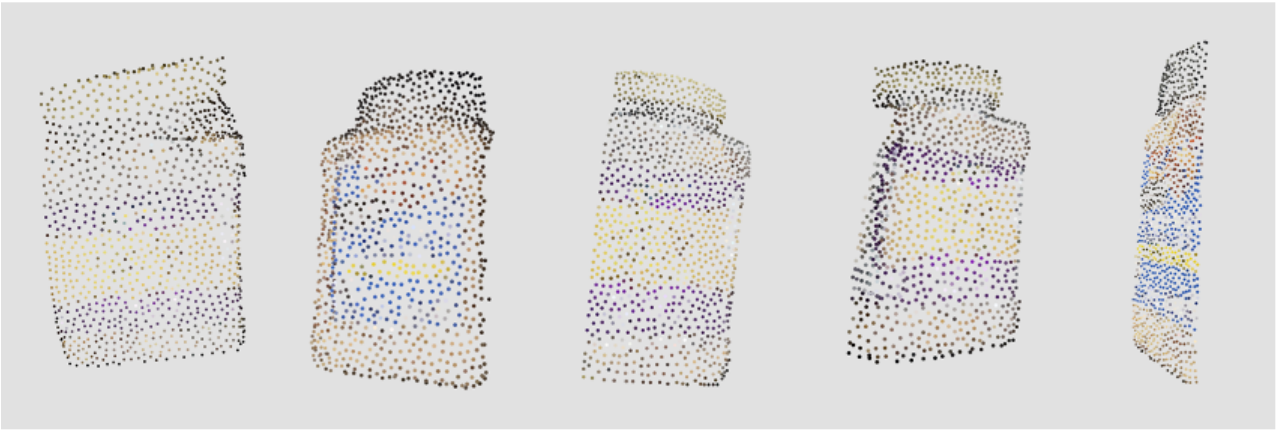}
(b) oreos
\includegraphics[width=0.95\linewidth,trim={0mm 11mm 0mm 9mm},clip]{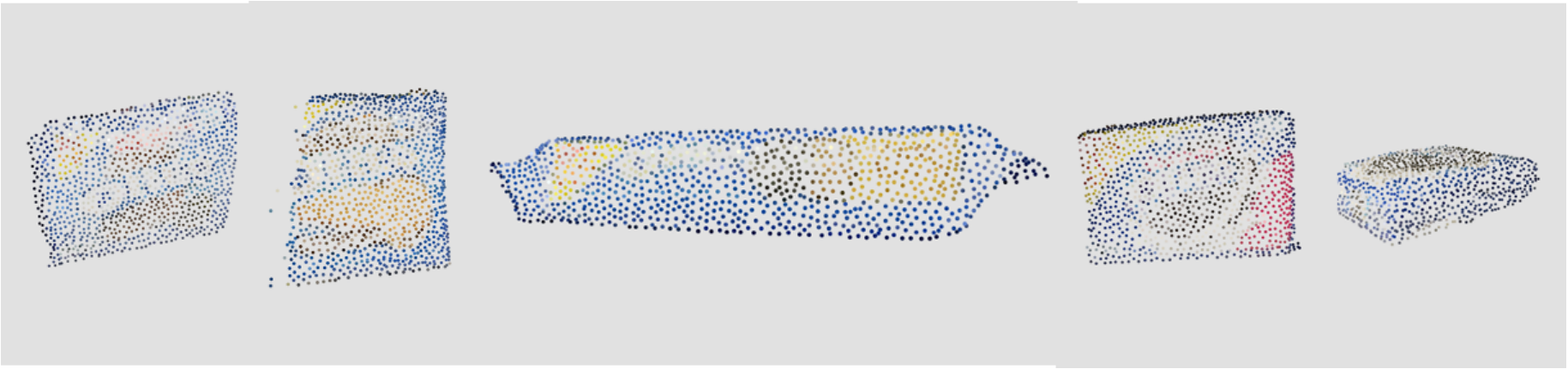}
(c) peanut-butter
\includegraphics[width=0.95\linewidth,trim={0mm 4mm 0mm 4mm},clip]{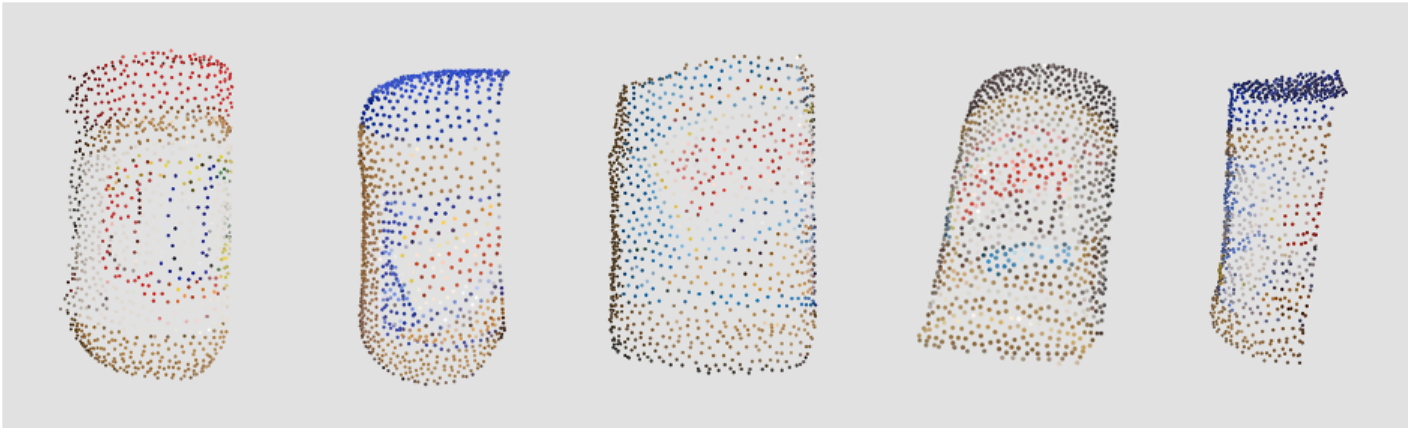}
(d) powerade
\includegraphics[width=0.95\linewidth,trim={0mm 0mm 1mm 2mm},clip]{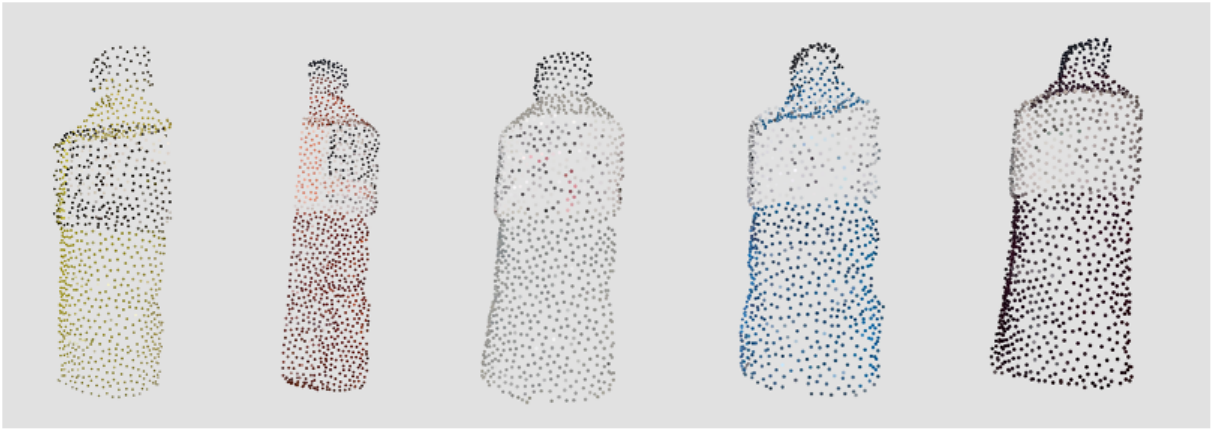}
(e) pretzel
\includegraphics[width=0.95\linewidth,trim={0mm 2mm 1mm 2mm},clip]{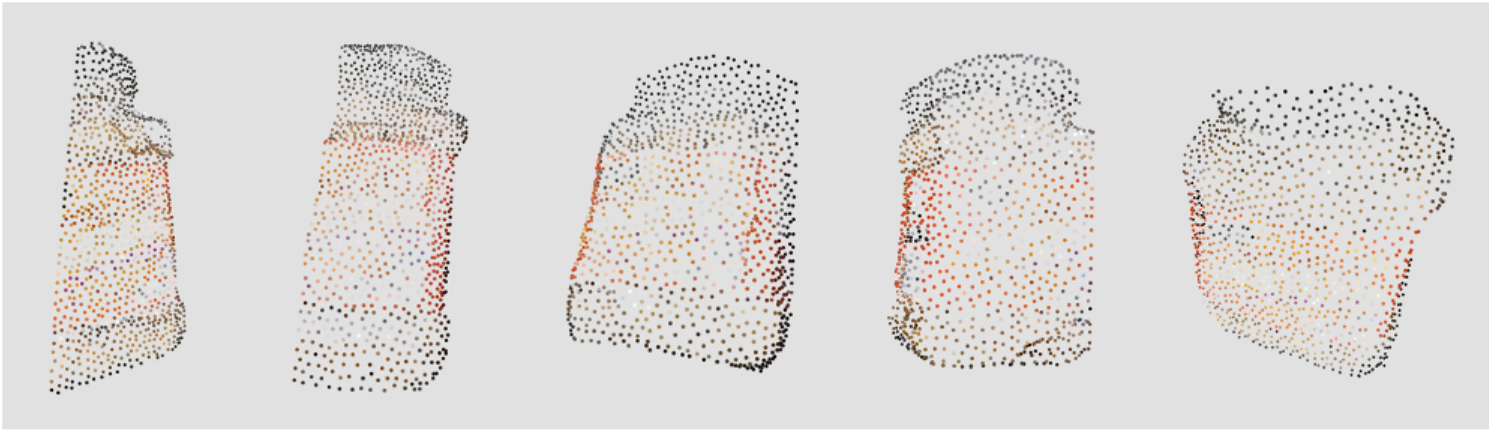}
(f) ranch
\includegraphics[width=0.95\linewidth,trim={0mm 10mm 1mm 10mm},clip]{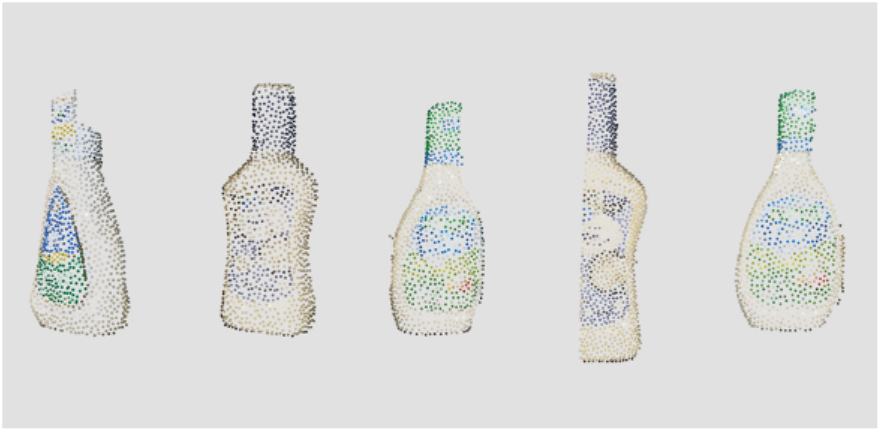}
(g) sugar
\includegraphics[width=0.95\linewidth,trim={0mm 5mm 1mm 5mm},clip]{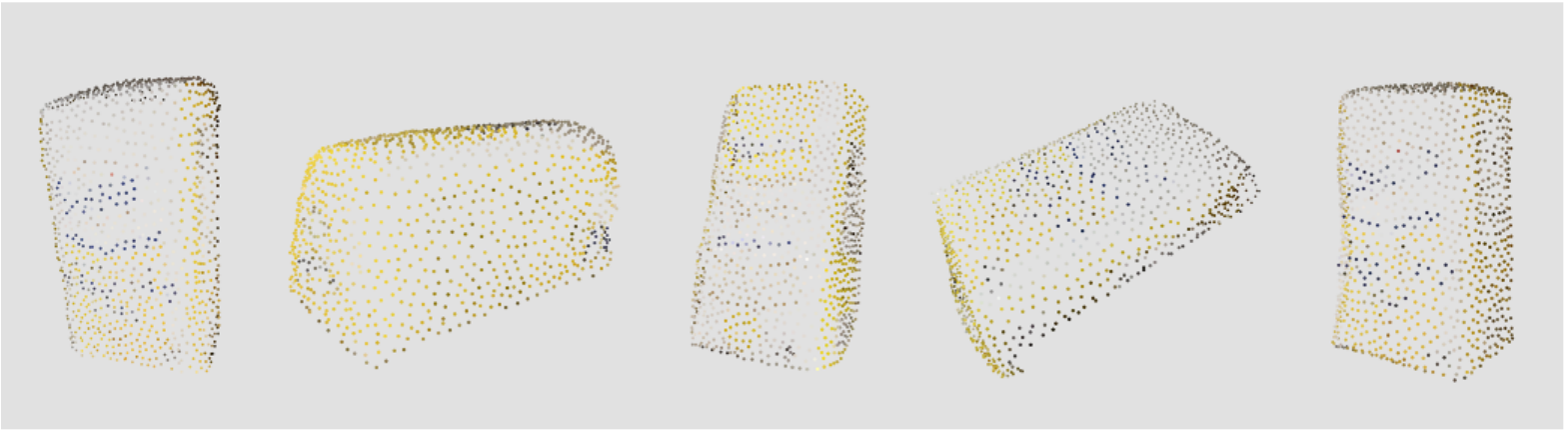}
(h) tomato-ketchup
\includegraphics[width=0.95\linewidth,trim={2mm 0mm 3mm 5mm},clip]{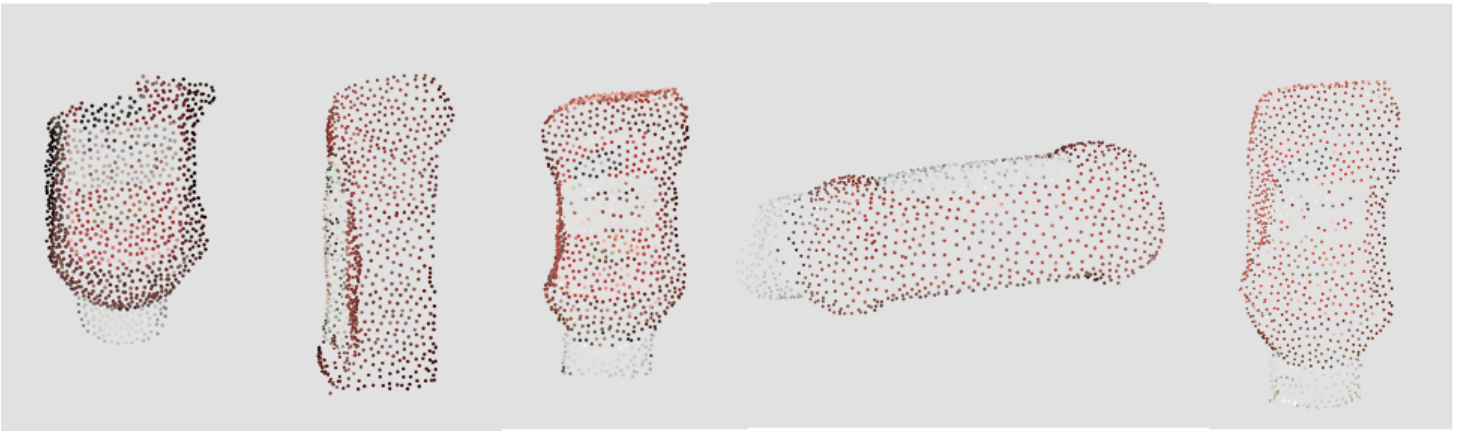}
\caption{Visual samples (1024 points) from 8 package classes. Labels on top of the objects. Zoom in for better visibility.}
\label{Fig20}
\end{figure}

\begin{figure}[H]
\centering
(a) tropicana-orange-juice
\includegraphics[width=0.95\linewidth,trim={1mm 3mm 1mm 2mm},clip]{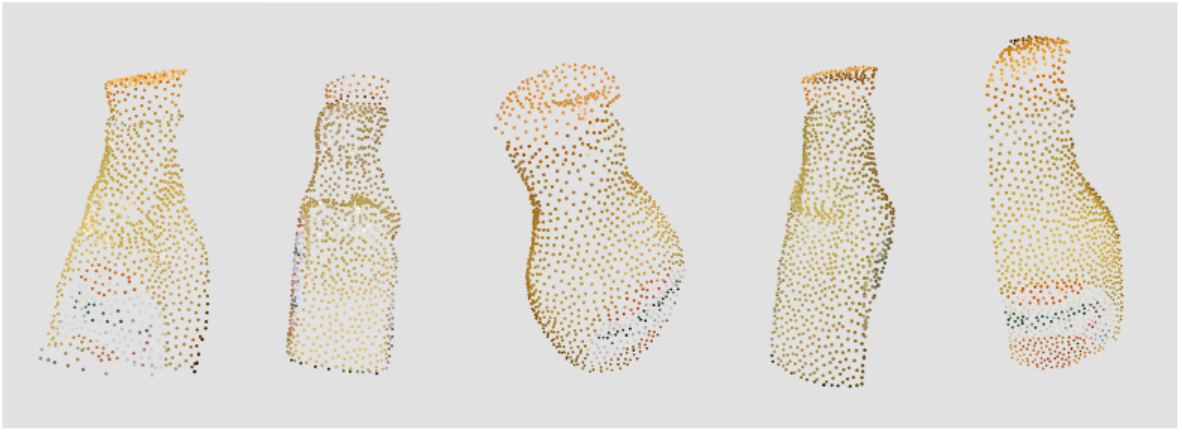}
(b) vinegar
\includegraphics[width=0.95\linewidth,trim={1mm 6mm 1mm 5mm},clip]{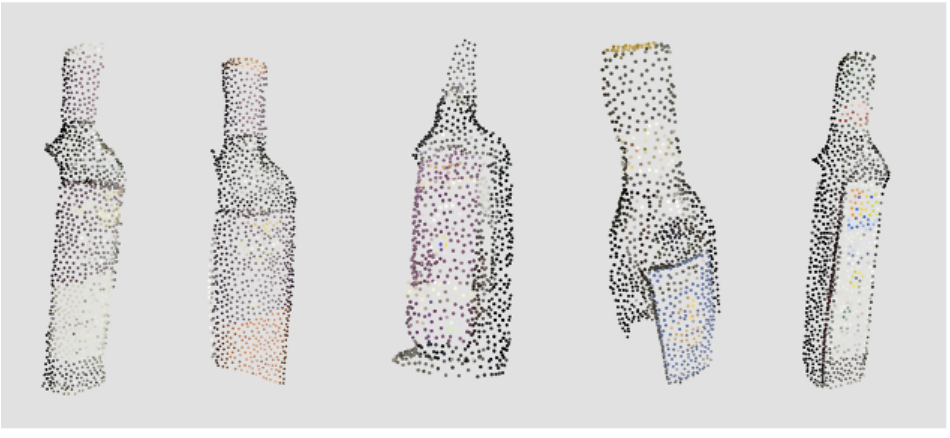}
(c) vitamin-water
\includegraphics[width=0.95\linewidth,trim={1mm 4mm 1mm 8mm},clip]{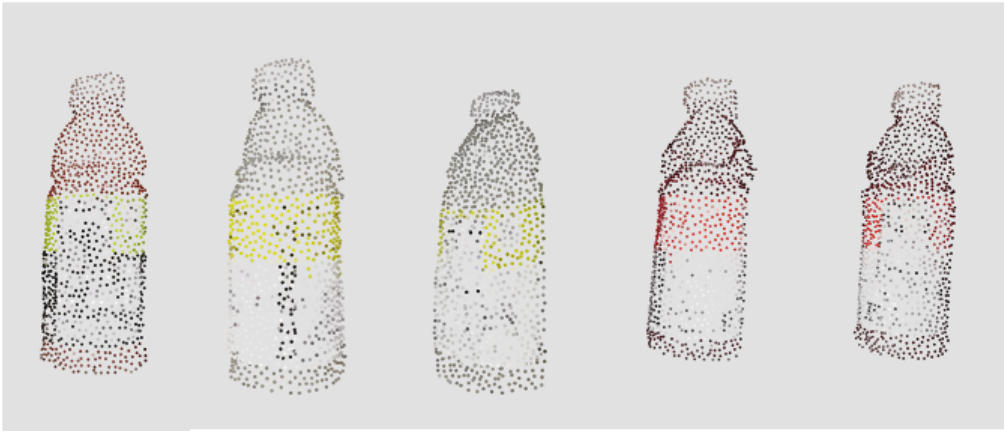}
(d) water-gallon
\includegraphics[width=0.95\linewidth,trim={0mm 8mm 1mm 8mm},clip]{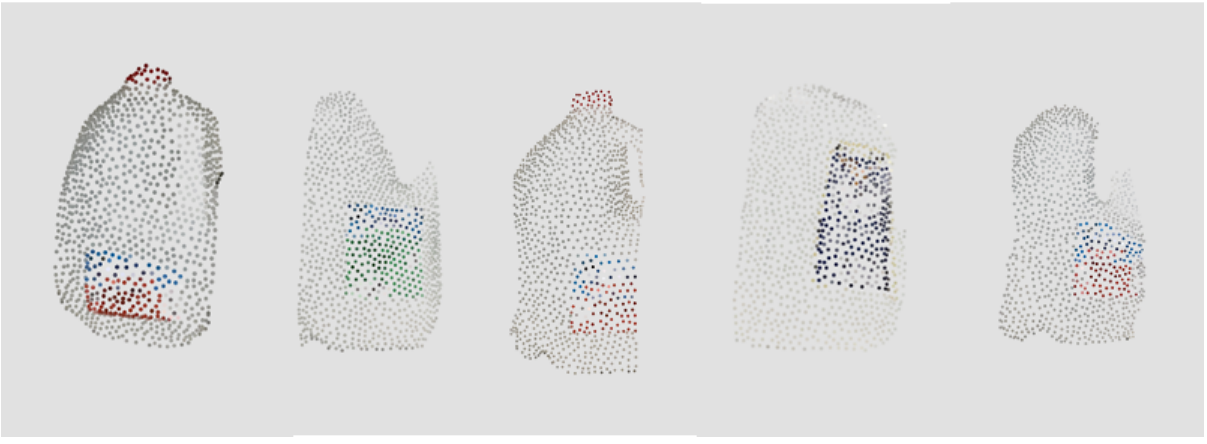}
(e) whip-cream
\includegraphics[width=0.95\linewidth,trim={0mm 10mm 1mm 8mm},clip]{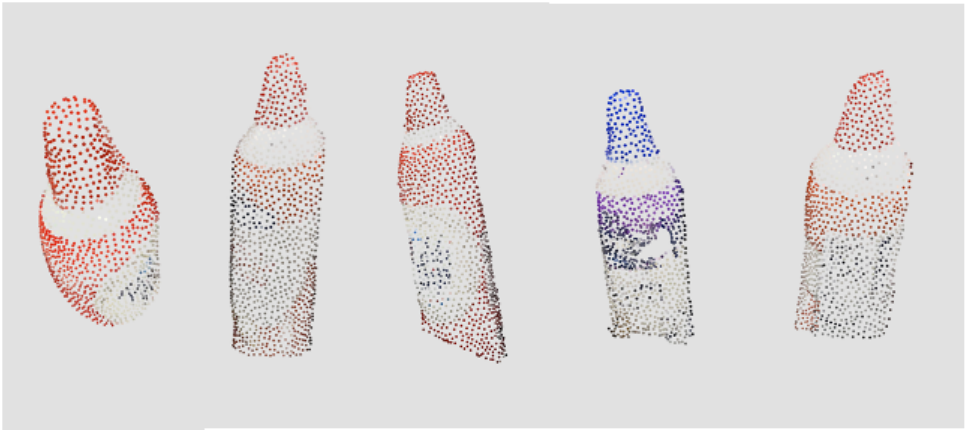}
\caption{Visual samples (1024 points) from 5 package classes. Labels on top of the objects. Zoom in for better visibility.}
\label{Fig21}
\end{figure}

Figure~\ref{Fig10} shows five samples from each of the ten apple classes. Each sample consists of 1024 points sampled using the farthest point sampling method. Figures~\ref{Fig11},~\ref{Fig12}, and ~\ref{Fig13} together show five samples from each of the 24 non-apple fruit classes. Figures~\ref{Fig14},~\ref{Fig15}, and ~\ref{Fig16} together show five samples from each of the 28 vegetable classes. Figures~\ref{Fig17},~\ref{Fig18},~\ref{Fig19},~\ref{Fig20} and ~\ref{Fig21} together show five samples from each of the 38 package classes. Zoom in to a sample for better visibility of 3D points and colors associated with the points. 
\newpage

\begin{table*}[!htbp]\centering
\caption{3D point cloud classification results on Apple10 subset dataset with 10 classes obtained from 1025 images resulting in 12905 point clouds. The train and test set constitute 9706 and 3199 point clouds, respectively.}
\begin{adjustbox}{width=\linewidth}
\begin{tabular}{c|c|c|c|c|c|c|c|c|c|c}
\toprule
\textbf{Models}&\ang{\textbf{apple-fuji}}&\ang{\textbf{apple-golden-delicious}}&\ang{\textbf{apple-granny-smith}}&\ang{\textbf{apple-honeycrisp}}&\ang{\textbf{apple-evercrisp}}&\ang{\textbf{apple-pazazz}}&
\ang{\textbf{apple-pink-lady}}&\ang{\textbf{apple-red-delicious}}&\ang{\textbf{apple-royal-gala}}&\ang{\textbf{apple-wild-twist}}\\
\midrule 
\#PCs&1223&1172&1350&1401&1337&1216&1339&1239&1249&1379\\
\midrule
Train&915&881&1016&1041&998&923&1014&936&947&1035\\
Test&308&291&334&360&339&293&325&303&302&344\\
\midrule
\multicolumn{11}{c}{1024 input points without colors}\\
\midrule
PointNet~\cite{qi2017pointnet}&05.88&30.95&22.97&16.94&29.44&26.28&12.48&11.11&09.24&31.44\\
PointNet++~\cite{qi2017pointnet++}&\textbf{20.77}&31.25&27.87&17.22&32.22&27.44&\textbf{19.58}&29.27&18.24&38.73\\
DGCNN~\cite{wang2019dynamic}&03.90&\textbf{41.24}&\textbf{29.34}&03.33&\textbf{35.10}&14.33&02.77&02.97&01.00&\textbf{53.78}\\
PCT~\cite{guo2021pct}&10.44&29.02&18.14&11.55&27.10&24.06&13.32&14.71&16.64&31.12\\
PointMLP~\cite{ma2022rethinking}&09.74&18.21&08.98&\textbf{18.06}&19.76&\textbf{32.08}&12.92&17.82&13.25&40.11\\
PointNeXt~\cite{PointNeXt}&10.06&29.21&17.66&14.72&27.43&23.55&16.62&\textbf{30.36}&\textbf{20.53}&27.33\\
\midrule
\multicolumn{11}{c}{1024 input points with colors}\\
\midrule
PointNet~\cite{qi2017pointnet}&65.39&\textbf{98.51}&97.89&72.22&76.39&89.10&64.01&83.14&67.36&76.00\\
PointNet++~\cite{qi2017pointnet++}&71.18&97.32&98.19&75.83&79.17&91.67&78.34&91.57&77.87&84.19\\
DGCNN~\cite{wang2019dynamic}&68.51&96.91&99.10&75.28&81.42&90.10&71.69&86.47&80.13&87.79\\
PCT~\cite{guo2021pct}&68.23&94.78&97.14&70.92&79.90&\textbf{92.45}&75.19&91.82&73.42&83.43\\
PointMLP~\cite{ma2022rethinking}&\textbf{81.49}&98.28&\textbf{99.70}&\textbf{81.67}&\textbf{90.27}&89.08&\textbf{89.23}&\textbf{93.40}&\textbf{86.42}&\textbf{93.02}\\
PointNeXt~\cite{PointNeXt}&68.51&96.56&97.01&75.83&80.24&90.44&81.54&89.77&79.47&90.12\\
\bottomrule
\end{tabular}
\end{adjustbox}
\label{table7}
\end{table*}

\begin{table*}[!h]\centering
  \caption{3D point cloud classification results on the dataset's fruits subset (non-apples) - 37,587 point clouds from 3,611 images spread across 24 classes. The train and test split consist of 18,406 and 6,276 point clouds, respectively.}
 \begin{adjustbox}{width=\linewidth}
  \begin{tabular}{c|c|c|c|c|c|c|c|c|c|c|c|c|c|c|c}
    \toprule
    \multicolumn{4}{c|}{Input Points}&\multicolumn{6}{c|}{1024 without colors}&\multicolumn{6}{c}{1024 with colors}\\
    \midrule
    \textbf{Classes }&\#PCs&Train&Test&\rot{PointNet~\cite{qi2017pointnet}}&\rot{PointNet++~\cite{qi2017pointnet++}}&\rot{DGCNN~\cite{wang2019dynamic}}&\rot{PCT~\cite{guo2021pct}}&\rot{PointMLP~\cite{ma2022rethinking}}&\rot{PointNeXt~\cite{PointNeXt}}&\rot{PointNet~\cite{qi2017pointnet}}&\rot{PointNet++~\cite{qi2017pointnet++}}&\rot{DGCNN~\cite{wang2019dynamic}}&\rot{PCT~\cite{guo2021pct}}&\rot{PointMLP~\cite{ma2022rethinking}}&\rot{PointNeXt~\cite{PointNeXt}}\\
    \midrule
avocado&1725&1285&440&39.69&48.90&36.36&40.36&\textbf{60.45}&52.50&99.34&99.78&99.77&99.79&99.77&99.32\\
banana&951&707&244&91.86&96.59&95.90&96.25&\textbf{98.77}&97.54&98.86&98.86&99.59&100.00&98.36&98.77\\
bartlett-pear&1115&839&276&17.71&\textbf{37.15}&24.28&13.29&30.07&32.61&87.15&90.28&91.30&88.00&91.67&84.42\\
cantaloupe&664&486&178&\textbf{35.63}&35.21&35.39&34.78&19.66&42.70&98.96&98.44&98.32&100.00&99.44&97.75\\
coconut&660&497&163&24.49&\textbf{47.23}&38.65&22.61&30.06&25.77&100.00&100.00&100.00&100.00&100.00&100.00\\
coconut2&1069&804&265&57.07&\textbf{84.45}&64.53&62.60&77.74&72.00&100.00&100.00&100.00&100.00&99.25&99.62\\
danjou-pear&1198&890&308&14.09&\textbf{35.32}&33.44&22.02&22.72&31.49&89.09&90.28&88.34&93.48&91.23&93.83\\
dragonfruit&611&442&169&53.55&\textbf{75.76}&75.15&60.12&50.30&75.74&100.00&98.96&99.41&100.00&95.86&100.00\\
grapefruit&944&730&214&43.87&\textbf{51.13}&26.17&28.97&08.41&28.50&98.21&99.11&99.53&100.00&99.07&98.60\\
honeydew-melon&514&382&132&17.34&\textbf{58.95}&36.36&23.49&21.21&28.79&98.61&100.00&100.00&100.00&95.46&100.00\\
kiwi&1347&989&358&35.80&45.79&\textbf{46.65}&25.99&41.06&40.50&97.40&99.48&99.16&99.58&96.93&99.16\\
lemon&1384&1013&371&26.24&\textbf{43.54}&24.53&16.56&31.54&33.69&98.96&99.22&99.46&99.15&98.38&99.19\\
lime&1661&1265&396&47.73&59.77&\textbf{68.69}&53.83&44.95&60.10&99.54&99.78&99.75&100.00&100.00&100.00\\
mango&1120&837&283&14.68&\textbf{31.36}&22.26&10.74&13.43&21.91&91.11&92.81&96.82&94.19&95.41&93.29\\
navel-orange&1256&954&302&34.35&\textbf{41.09}&32.12&32.56&10.60&25.83&98.63&99.68&99.67&99.48&99.67&99.34\\
nectarine&1241&886&355&13.88&\textbf{23.29}&11.27&06.28&06.20&07.32&90.31&93.56&96.62&94.40&95.21&92.96\\
papaya&420&315&105&40.20&\textbf{57.38}&46.67&34.12&57.14&50.48&93.26&93.26&96.19&96.55&97.14&94.29\\
peach&1332&1005&327&13.65&23.89&23.55&18.31&10.09&\textbf{27.83}&92.10&97.02&95.41&96.22&97.55&96.94\\
pear-bosc&1155&871&284&37.95&\textbf{65.19}&23.94&36.59&56.34&58.10&98.40&99.36&99.30&99.70&99.65&99.65\\
pineapple&347&257&90&89.17&92.74&86.67&82.71&86.67&\textbf{93.33}&100.00&100.00&100.00&100.00&100.00&100.00\\
plums&1474&1101&373&41.54&\textbf{59.67}&37.27&41.07&44.50&36.73&99.22&98.91&100.00&99.38&99.73&98.39\\
pomegranate&821&603&218&05.94&\textbf{34.40}&17.43&10.41&21.10&21.10&96.14&97.92&95.87&96.80&95.41&94.50\\
red-pear&1113&821&292&18.02&\textbf{22.72}&06.85&08.38&21.23&21.23&97.74&99.36&100.00&99.21&99.32&99.32\\
watermelon&560&427&133&19.05&29.76&24.06&17.36&\textbf{47.37}&32.33&97.62&100.00&99.25&98.15&100.00&99.25\\
    \bottomrule
  \end{tabular}
  \end{adjustbox}
  \label{table8}
\end{table*}

\begin{table*}[!h]\centering
  \caption{3D point cloud classification results on the vegetable subset of the dataset - 27,149 point clouds from 3055 images spread across 28 classes. The train and test split consist of 20720 and 6987 point clouds, respectively.}
 \begin{adjustbox}{width=0.97\linewidth}
  \begin{tabular}{c|c|c|c|c|c|c|c|c|c|c|c|c|c|c|c}
    \toprule
    \multicolumn{4}{c|}{Input Points}&\multicolumn{6}{c|}{1024 without colors}&\multicolumn{6}{c}{1024 with colors}\\
    \midrule
    $\downarrow$\textbf{Classes }\textbf{Models}$\rightarrow$&\#PCs&Train&Test&\rot{PointNet~\cite{qi2017pointnet}}&\rot{PointNet++~\cite{qi2017pointnet++}}&\rot{DGCNN~\cite{wang2019dynamic}}&\rot{PCT~\cite{guo2021pct}}&\rot{PointMLP~\cite{ma2022rethinking}}&\rot{PointNeXt~\cite{PointNeXt}}&\rot{PointNet~\cite{qi2017pointnet}}&\rot{PointNet++~\cite{qi2017pointnet++}}&\rot{DGCNN~\cite{wang2019dynamic}}&\rot{PCT~\cite{guo2021pct}}&\rot{PointMLP~\cite{ma2022rethinking}}&\rot{PointNeXt~\cite{PointNeXt}}\\
    \midrule
	artichokes&558&431&127&42.26&\textbf{64.88}&58.27&60.62&61.42&55.91&98.81&99.40&99.21&99.04&100.00&99.21\\
	broccoli&901&677&224&29.58&\textbf{70.00}&54.91&50.68&67.41&59.82&97.70&98.33&99.11&99.60&99.11&98.66\\
	cucumber&1024&748&276&63.19&87.50&\textbf{90.58}&79.51&81.88&82.61&100.00&99.31&100.00&99.22&99.28&99.28\\
	eggplant&524&375&149&29.52&\textbf{51.55}&42.28&31.31&48.99&47.65&93.57&97.62&97.99&96.76&96.64&97.99\\
	garlic&1160&854&306&38.27&\textbf{63.32}&62.75&50.17&59.15&47.06&92.31&98.72&97.71&96.27&98.69&97.71\\
	ginger&936&707&229&62.12&\textbf{88.26}&72.05&74.25&73.36&75.11&100.00&99.24&99.56&99.26&99.56&99.13\\
	green-bell-pepper&1476&1146&330&47.78&\textbf{63.33}&53.63&55.63&62.73&53.33&98.06&96.39&97.27&97.88&96.36&94.85\\
	green-cabbage&551&413&138&32.41&37.96&23.19&30.30&43.48&\textbf{44.20}&96.53&100.00&100.00&99.24&97.83&97.83\\
	jalapeno&1262&936&326&43.71&\textbf{61.44}&24.54&48.27&55.22&49.39&95.41&95.92&95.71&92.77&96.32&92.33\\
	orange-bell-pepper&876&669&207&15.42&26.50&9.66&21.42&\textbf{28.50}&22.22&99.58&98.75&98.55&99.19&99.03&98.07\\
	potato&1079&826&253&27.57&\textbf{50.42}&38.34&38.58&26.48&37.55&94.88&99.62&98.81&98.46&95.65&99.60\\
	red-bell-pepper&846&613&233&44.90& \textbf{61.02}&30.47&44.19&45.92&46.35&98.86&99.24&97.43&97.53&96.57&98.71\\
	red-cabbage&433&321&112&40.81& \textbf{59.46}&35.71&51.72&52.68&44.64&100.00&100.00&100.00&99.18&100.00&99.11\\
	red-potato&1499&1135&364&26.20& 42.78&37.78&32.67&32.22&\textbf{46.70}&98.26&99.65&99.63&100.00&99.26&98.08\\
	redonion&1113&843&270&34.67& \textbf{53.76}&45.88&42.27&46.15&43.33&97.36&97.14&97.53&98.07&98.90&98.15\\
	rutabagas&612&457&155&16.88& 30.47&21.94&28.44&30.32&\textbf{30.97}&96.43&98.81&97.42&98.10&97.42&99.35\\
	squash-acorn&787&591&196&27.78& \textbf{49.07}&53.57&39.29&47.96&35.71&97.22&97.22&98.98&99.31&100.00&96.94\\
	squash-butternut&504&369&135&32.30& 57.65&45.93&38.03&\textbf{61.48}&57.78&93.98&96.81&99.26&97.64&96.29&96.30\\
	squash-spaghetti&625&470&155&27.92& \textbf{54.38}&33.55&30.97&24.52&49.03&90.31&87.29&96.77&96.13&93.55&94.19\\
	squash-yellow&1093&819&274&46.88& 60.13&57.30&53.39&58.03&\textbf{67.88}&98.61&97.57&98.18&96.79&98.54&97.45\\
	sweet-potato&1167&906&261&28.47& 39.58&19.92&31.60&\textbf{41.00}&34.10&95.89&98.26&95.79&95.63&97.70&96.93\\
	tomato&1446&1042&404&37.44& \textbf{55.08}&23.02&37.67&36.39&49.26&92.75&97.45&95.79&92.99&94.06&94.80\\
	turnip&1031&780&251&26.29& 49.02&40.64&54.80&39.84&\textbf{57.37}&97.73&99.62&96.81&98.84&98.01&98.01\\
	vine-tomato&1411&1044&367&63.48& \textbf{78.92}&73.02&70.30&70.03&64.58&97.30&97.30&95.37&97.50&96.46&95.10\\
	whiteonion&1398&1030&368&20.99& 43.88&12.50&40.89&\textbf{44.84}&41.85&96.25&99.22&97.83&97.60&97.28&98.91\\
	yam&1290&979&311&24.92&24.92& 24.92&21.54&26.69&\textbf{31.83}&95.77&99.70&97.75&98.75&98.75&98.07\\
	yellow-bell-pepper&843&640&203&16.80&\textbf{44.11}&37.93&24.49&26.11&30.05&98.54&98.55&98.03&97.88&98.03&95.07\\
	yellow-onion&1262&899&363&17.44&31.77&40.77&24.42&27.55&\textbf{43.53}&99.48&100.00&100.00&99.74&99.45&100.00\\
    \bottomrule
  \end{tabular}
  \end{adjustbox}
  \label{table9}
\end{table*}

\begin{table*}[!h]\centering
  \caption{3D point cloud classification results on Packages subset dataset - 22,604 point clouds from 4121 images spread across 38 classes. The train and test split consist of 17214 and 5390 point clouds, respectively.}
 \begin{adjustbox}{width=\linewidth}
  \begin{tabular}{c|c|c|c|c|c|c|c|c|c|c|c|c|c|c|c}
    \toprule
    \multicolumn{4}{c|}{Input Points}&\multicolumn{6}{c|}{1024 without colors}&\multicolumn{6}{c}{1024 with colors}\\
    \midrule
    $\downarrow$\textbf{Classes }\textbf{Models}$\rightarrow$&\#PCs&Train&Test&\rot{PointNet~\cite{qi2017pointnet}}&\rot{PointNet++~\cite{qi2017pointnet++}}&\rot{DGCNN~\cite{wang2019dynamic}}&\rot{PCT~\cite{guo2021pct}}&\rot{PointMLP~\cite{ma2022rethinking}}&\rot{PointNeXt~\cite{PointNeXt}}&\rot{PointNet~\cite{qi2017pointnet}}&\rot{PointNet++~\cite{qi2017pointnet++}}&\rot{DGCNN~\cite{wang2019dynamic}}&\rot{PCT~\cite{guo2021pct}}&\rot{PointMLP~\cite{ma2022rethinking}}&\rot{PointNeXt~\cite{PointNeXt}}\\
    \midrule
almond-milk&275&210&65&87.65&\textbf{91.52}&87.69&88.04&86.15&90.77&98.96&98.95&98.46&99.02&98.46&98.46\\
apple-juice&429&330&99&79.05&85.95&77.78&77.75&85.86&\textbf{91.92}&99.17&99.17&98.99&99.07&97.98&97.98\\
apple-sauce&664&491&173&84.71&\textbf{96.88}&94.80&87.36&91.91&96.53&100.00&100.00&100.00&100.00&100.00&100.00\\
bagels&784&649&135&57.42&\textbf{72.71}&60.00&59.84&60.00&66.67&98.21&98.80&100.00&97.83&97.78&97.04\\
barbecue-sauce&445&344&101&82.50&\textbf{88.33}&87.13&81.43&88.12&88.12&100.00&100.00&100.00&100.00&100.00&99.01\\
beans&883&648&235&68.38&69.27&70.64&67.19&69.79&\textbf{71.06}&97.57&97.57&98.30&96.97&96.17&97.87\\
beans-green&803&620&183&53.54&\textbf{71.48}&59.02&60.26&61.20&65.03&99.48&100.00&97.81&99.52&99.45&100.00\\
bread&947&709&238&73.32&\textbf{82.41}&82.77&75.14&77.73&78.15&98.26&98.86&98.74&97.48&98.32&97.48\\
cashews&449&336&113&62.73&\textbf{77.78}&65.49&68.75&71.68&75.22&93.75&99.31&99.12&98.13&100.00&98.23\\
cheese&628&496&132&86.81&\textbf{93.75}&90.15&81.25&88.64&89.39&99.31&99.31&100.00&99.65&100.00&100.00\\
chips&416&320&104&52.33&68.83&64.58&48.81&\textbf{76.04}&67.71&86.17&91.33&89.58&81.45&86.46&87.50\\
chocolate-syrup&508&391&117&78.55&\textbf{92.61}&90.60&89.76&90.59&91.45&100.00&100.00&100.00&100.00&100.00&100.00\\
coconut-water&465&353&112&77.45&\textbf{90.74}&84.82&82.89&85.71&90.18&97.92&97.92&99.11&95.52&97.32&94.64\\
coffee&492&395&97&54.50&\textbf{88.00}&75.26&76.91&81.44&84.54&91.50&93.67&92.78&93.99&96.91&95.88\\
corn&738&546&192&60.61&\textbf{73.90}&64.58&71.99&61.98&68.23&99.07&99.07&98.43&97.85&98.96&97.92\\
dip&445&343&102&82.08&83.99&\textbf{87.26}&83.93&82.35&82.35&99.17&98.75&100.00&98.44&99.02&100.00\\
eggs&721&547&174&77.46&82.77&81.61&77.11&\textbf{87.93}&86.78&100.00&100.00&100.00&100.00&100.00&100.00\\
eggs-in-cooler&894&670&224&54.55&\textbf{84.09}&81.70&75.71&83.93&81.25&96.97&95.83&99.11&97.55&99.11&98.66\\
energy-drink&679&512&167&80.07&87.43&89.82&85.83&\textbf{92.22}&91.02&97.40&96.88&99.40&94.69&98.80&95.21\\
gatorade&631&484&147&80.36&86.31&\textbf{91.16}&83.13&89.79&89.12&100.00&99.40&100.00&99.65&100.00&100.00\\
ham&652&484&168&70.83&81.77&83.33&71.59&82.73&\textbf{86.90}&98.70&99.22&99.40&99.43&99.40&97.62\\
ice-tea-gallon&525&393&132&51.74&68.54&62.88&55.56&61.36&\textbf{71.97}&98.47&99.31&100.00&99.65&99.24&99.24\\
mayonnaise&820&613&207&91.52&95.78&\textbf{96.62}&90.48&94.69&93.24&100.00&100.00&91.07&100.00&100.00&100.00\\
milk&643&484&159&61.09&\textbf{74.14}&63.83&72.12&67.02&69.15&84.30&89.29&86.17&88.08&91.49&90.43\\
milk-in-cooler&615&447&168&48.44&66.08&66.07&50.25&63.69&\textbf{69.64}&89.58&90.97&91.07&87.38&87.50&91.67\\
mixed-nuts&429&324&105&50.06&63.94&60.95&55.02&\textbf{70.47}&60.95&99.17&99.17&100.00&100.00&100.00&99.05\\
oreos&757&582&175&61.57&\textbf{88.43}&80.00&78.94&80.00&80.57&99.54&100.00&99.43&99.57&99.43&99.43\\
peanut-butter&610&466&144&70.65&77.28&\textbf{77.78}&74.29&76.39&75.69&97.62&100.00&99.31&98.45&98.61&99.31\\
powerade&517&388&129&89.16&94.38&\textbf{96.12}&94.40&91.47&89.92&99.31&99.31&99.22&100.00&100.00&100.00\\
pretzel&762&587&175&83.48&86.91&78.29&78.57&83.43&\textbf{87.43}&99.48&98.96&96.00&97.77&97.71&97.14\\
ranch&409&294&115&71.23&\textbf{87.50}&67.83&67.76&79.13&81.74&96.53&100.00&100.00&100.00&99.13&98.26\\
sugar&454&342&112&52.08&69.44&66.96&65.37&\textbf{73.21}&69.64&100.00&99.31&100.00&100.00&100.00&99.11\\
tomato-ketchup&613&480&133&77.46&81.91&\textbf{83.46}&77.62&78.95&79.70&98.47&100.00&99.25&99.46&100.00&99.25\\
tropicana-orange-juice&420&317&103&69.40&86.43&87.38&78.70&\textbf{88.35}&\textbf{88.35}&100.00&100.00&100.00&100.00&100.00&100.00\\
vinegar&439&331&108&87.50&\textbf{95.00}&84.26&82.14&80.56&83.33&97.50&96.67&99.07&98.13&98.15&99.07\\
vitamin-water&624&488&136&94.05&\textbf{95.83}&92.65&93.75&95.59&94.12&99.40&99.40&100.00&99.65&100.00&99.26\\
water-gallon&571&442&129&63.59&\textbf{84.07}&79.85&75.65&72.09&75.97&100.00&100.00&100.00&100.00&100.00&99.22\\
whip-cream&723&568&155&93.94&94.53&\textbf{95.48}&93.47&92.26&92.90&100.00&100.00&100.00&100.00&100.00&100.00\\
   \bottomrule
  \end{tabular}
  \end{adjustbox}
  \label{table10}
\end{table*}

\begin{figure*}[!ht]
\centering
\begin{multicols}{2}
    \includegraphics[width=\linewidth,trim={5mm 5mm 5mm 15mm},clip]{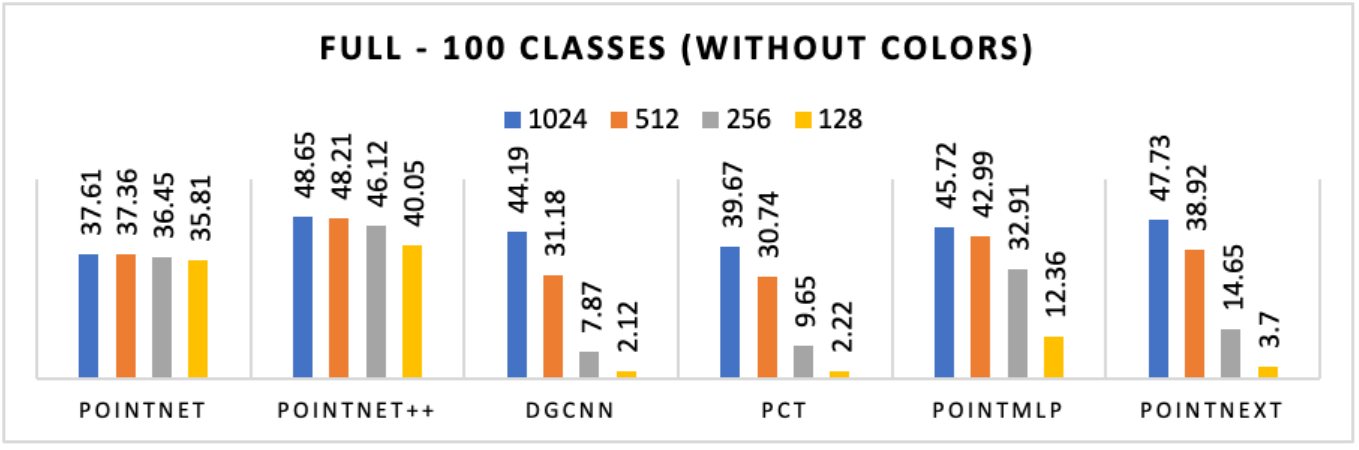}
   (a) Full - without colors\par
    \includegraphics[width=\linewidth,trim={5mm 5mm 5mm 15mm},clip]{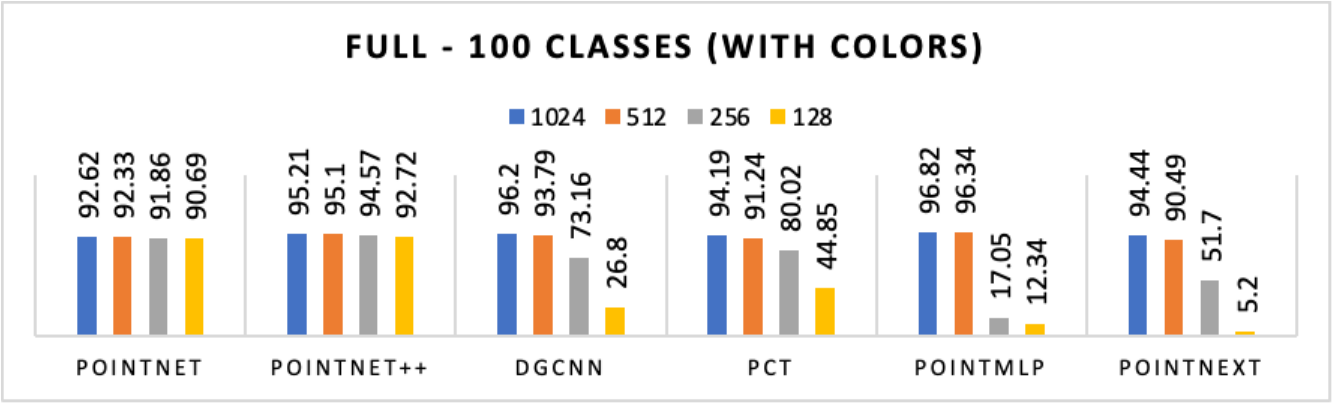}
   (b) Full - with colors\par
\end{multicols}
\begin{multicols}{2}
    \includegraphics[width=\linewidth,trim={5mm 5mm 5mm 25mm},clip]{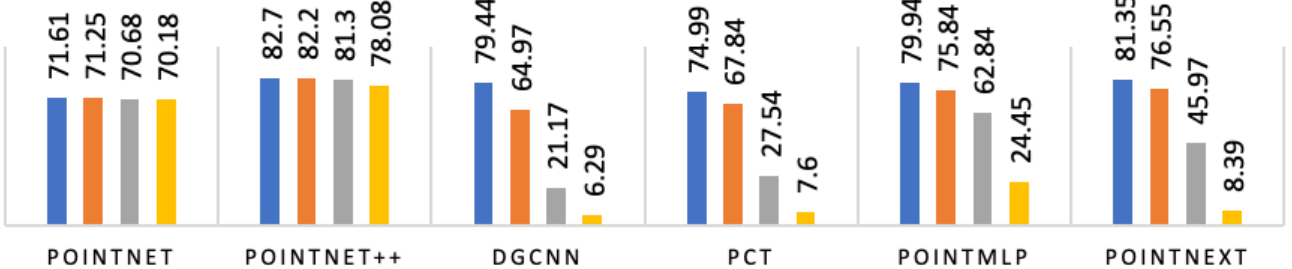}
    (a) Packages - without colors\par
    \includegraphics[width=\linewidth,trim={5mm 5mm 5mm 25mm},clip]{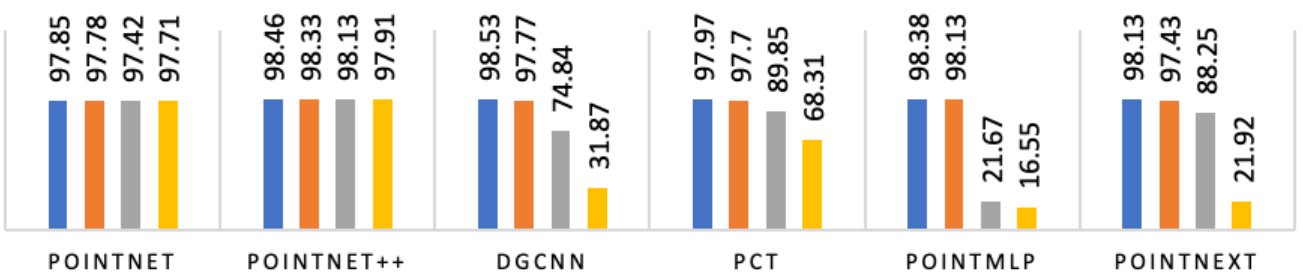}
    (b) Packages - with colors\par
\end{multicols}
\begin{multicols}{2}
    \includegraphics[width=\linewidth,trim={5mm 5mm 5mm 25mm},clip]{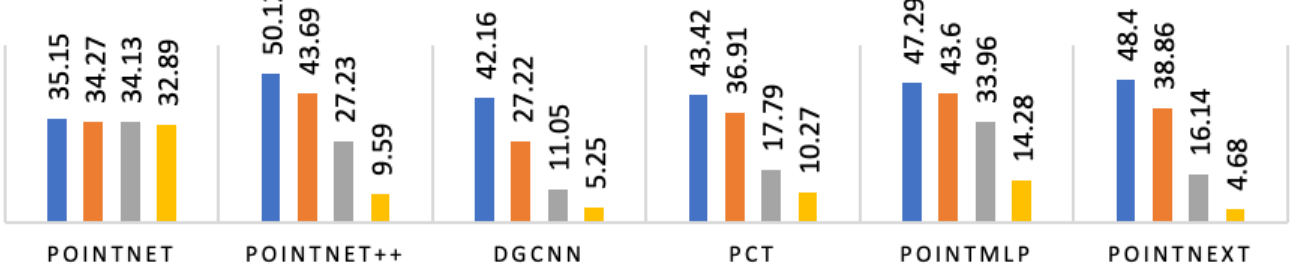}
    (a) Vegetables - without colors\par
    \includegraphics[width=\linewidth,trim={5mm 5mm 5mm 25mm},clip]{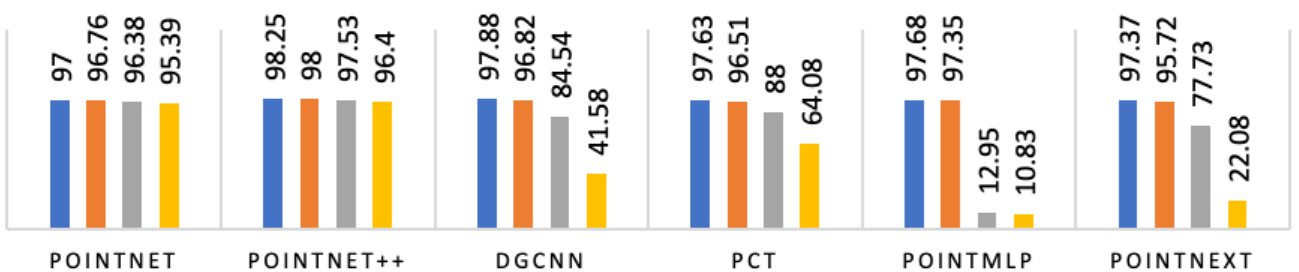}
    (b) Vegetables - with colors\par
\end{multicols}
\begin{multicols}{2}
    \includegraphics[width=\linewidth,trim={5mm 5mm 5mm 25mm},clip]{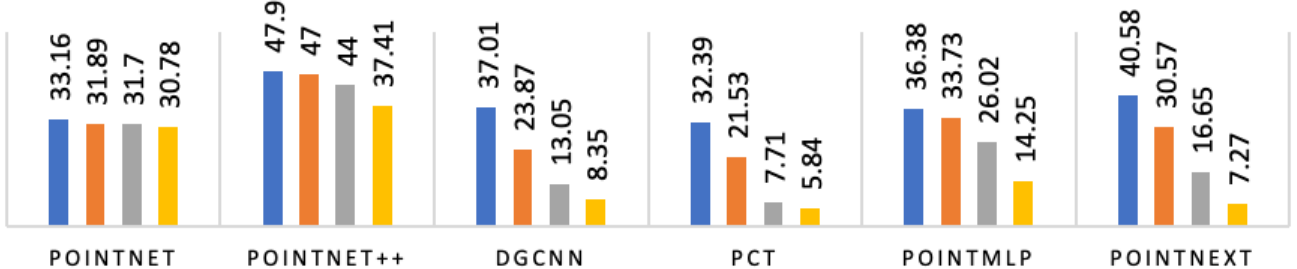}
    (a) Fruits - without colors\par
    \includegraphics[width=\linewidth,trim={5mm 5mm 5mm 25mm},clip]{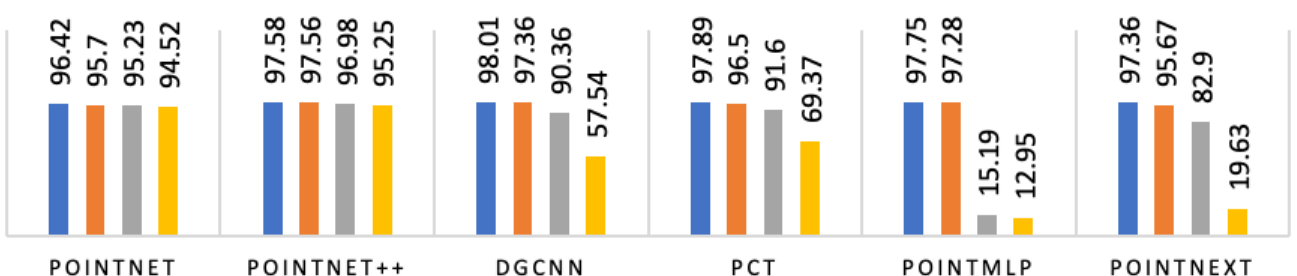}
    (b) Fruits - with colors\par
\end{multicols}
\begin{multicols}{2}
    \includegraphics[width=\linewidth,trim={5mm 5mm 5mm 25mm},clip]{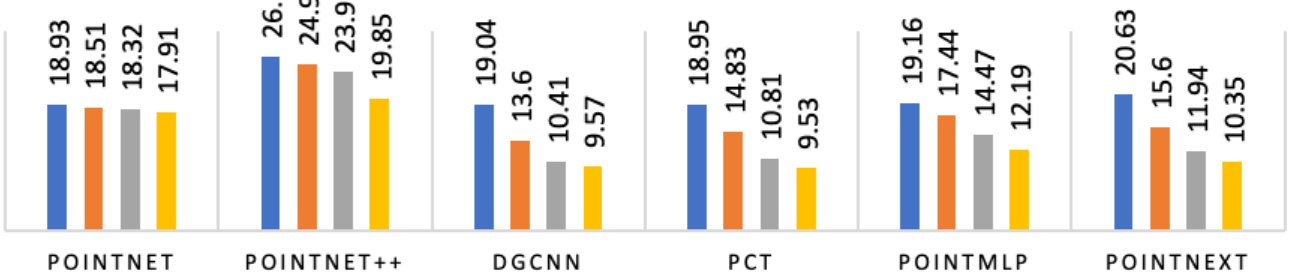}
    (a) Apple10 - without colors\par
    \includegraphics[width=\linewidth,trim={5mm 5mm 5mm 22mm},clip]{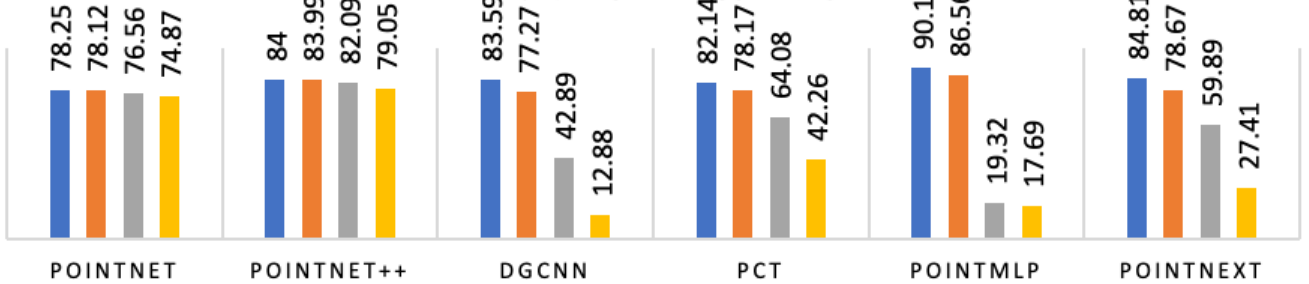}
    (b) Apple10 - with colors\par
\end{multicols}

\caption{Performance of state-of-the-art methods with a different number of input points, i.e., 1024 (blue), 512 (orange), 256 (grey), and 128 (yellow).}
\label{Fig22}
\end{figure*}

\section{3D Point cloud Classification - detailed results}
Figure~\ref{Fig3} shows 3D samples from each of the ten apple classes with colors (a-j) and the respective 3D apples without colors(k-t). While apples with colors are often visibly distinguishable, it is challenging to recognize them without colors. Also, the appearance of apples due to color is quite similar between classes, while the shapes of different apples in the same class vary. For example, in the apple10 subset, eight of the ten apple classes are light red to dark red, posing significant challenges to the models in learning discriminative features. Unlike Fruits and Vegetables, Packages often have a fixed shape for a given class and do not suffer surface deformations for different instances of the same class. 

Several Packages, such as tomato ketchup, ranch, and mayonnaise, are stacked straight and upside down in a rack. In the case of fruits and vegetables, there is no specific order of arrangement. Some stores where the fruits and vegetables are placed in racks suffered from relatively proper illumination compared to objects at the front of the rack leading to variations in the amount of lighting received. 

Table~\ref{table7} shows the class-wise accuracy of the Apple10 subset, both with and without colors. Table~\ref{table8} lists the class-wise accuracy of 24 fruit classes. Since apples without color are hard to distinguish from other apples, we do not include apples with other fruits in this experiment. Tables~\ref{table9} and~\ref{table10} show 3D point cloud classification results of vegetables and packages, respectively. Each table contains the number of point clouds used as train and test samples for each class. The maximum accuracy for each class without colors is highlighted in bold. Each model trained on point clouds with colors achieved 100\% accuracy on multiple classes. Here we list some of our findings from those confusion matrices. From the confusion matrix of each state-of-the-art method on our full dataset without colors, we observe that none of the classes from packages (38 classes) is misclassified as a non-package class (62 classes - fruits and vegetables) or vice-versa. Table~\ref{table8} consists of 3D Fruits (non-apple classes) results in both with and without colors on the six methods~\cite{qi2017pointnet,qi2017pointnet++,wang2019dynamic,ma2022rethinking,guo2021pct, PointNeXt}. Colors as features contribute richly to learning discriminative features. In the absence of colors, fruits with a similar shape, such as watermelon, honeydew-melon, and cantaloupe, are misclassified amongst each other.

Table~\ref{table9} shows 3D Vegetables results with and without colors on each of the six methods. The Vegetables without colors showed improved classification results compared to Fruits without colors. However, a few ambiguous misclassifications are observed. e.g., tomato as red-potato. For example, without colors, a tiny percentage of tomatoes are misclassified as vine-tomato attributed because of the red color of the objects. All six models in our experiments achieve higher classification performance without colors as additional features. Among the six state-of-the-art methods, PointNet++~\cite{qi2017pointnet} achieves the highest classification accuracy on the Packages (without colors).

\section{Ablations}

\subsection{Quality vs. Quantity}
We test the robustness of state-of-the-art models by giving fewer input points, i.e., 512, 256, and 128 points, during testing. This test helps us understand how effectively a given model infers an object's class. Figure~\ref{Fig22} shows the drop in classification accuracy of the six state-of-the-art methods when evaluated with our dataset's five subsets and two variants (with and without colors). Both PointNet~\cite{qi2017pointnet} and PointNet++~\cite{qi2017pointnet++} continued to show robust behavior when evaluated on all five subsets and two variants with 512, 256, and 128 points. However, recent methods such as PointMLP~\cite{ma2022rethinking} and PointNeXt~\cite{PointNeXt} suffered with 256 and 128 input points. 

\subsection{Color is all you need}

From the classification results observed in Tables~\ref{table7},~\ref{table8}, ~\ref{table9}, and~\ref{table10}, it is evident that colors as additional features enrich the appearance of point cloud objects and enhance the classification results by a large margin (some even achieve 100\% accuracy). Also, from the confusion matrix of PointNet++~\cite{qi2017pointnet++} (trained with colors), as shown in figure~\ref{Fig22}b, we observe no major ambiguity issues between the classes. However, figure~\ref{Fig22}a shows that learning discriminative features from the geometry of the objects alone suffered, especially for similar classes, such as apples that are closer in their geometric structure (spherical). We observe similar behavior with geometrically similar grocery objects, such as 1) whiteonion, redonion, and yellowonion, 2) watermelon, cantaloupe, and honeydew-melon, 3) mixed-nuts, cashews, and pretzels, 4) green-bell-pepper, orange-bell-pepper, red-bell-pepper, and yellow-bell-pepper. Despite their similarity in their structure, these classes are better distinguished when colors are used. 

\section{Limitations and Future Work}

Our work is limited by the challenges posed by the sub-components, such as RGB-D to Point cloud conversion, outlier removal, and 3D point cloud classification components. Improvement to each component improves the quality of point clouds captured and methods to classify the point clouds. Annotation of RGB images is manual and time-consuming. One future direction is to use unsupervised learning-based methods in 3D grocery classification.

\section{More Benchmarking}

\subsection{2D Image Patch Classification}

The annotations provide image patches of grocery items, and we use these image patches scaled to the size of 256$\times$256 for 2D image patch classification. For our experiment, we use SwinTransformer~\cite{liu2021swin} model and the classification results are shown in table~\ref{table11}. 3D point clouds with colors classification results of table~\ref{tab2} are superior to 2D image patch classification for Apple10 and Packages subsets as shown in table~\ref{table11}, and the results for other subsets are comparable. 

\begin{table}[!h]\centering
  \caption{2D image classification results using the instance images of the four subsets and the complete dataset on SwinTransformer~\cite{liu2021swin} model.}
  \begin{adjustbox}{width=0.97\linewidth}
  \begin{tabular}{c|c|c|c|c|c}
    \toprule
    \textbf{Models}&\textbf{Apple10}&\textbf{Fruits}&\textbf{Vegetables}&\textbf{Packages}&\textbf{Full}\\
    \midrule
    \#Classes &10&24&28&38&100\\
    \midrule 
    \#Images &1025&2586&3029&4115&10755\\
    \midrule
    Train&772&1944&2264&3103&8083\\
    Test&253&642&765&1012&2672\\
    \midrule
    \#Patches &12905&24682&27707&22604&87898\\
    \midrule
    Train&9706&18406&20720&17214&66032\\
    Test&3199&6276&6987&5390&21866\\
    \midrule
 ST~\cite{liu2021swin}&85.10&98.50&98.40&98.3&98.5\\
    \bottomrule
  \end{tabular}
  \end{adjustbox}
  \label{table11}
\end{table}

\subsection{Few-Shot}

In this section, we elaborate our findings for Point Cloud Few-shot \textit{Baseline} evaluations and \textit{Meta-Learning} tasks explained in Section \ref{sec:FSL}. Table \ref{table3} and Table \ref{table13} demonstrate the few-shot baseline task where we transfer-learn the pre-trained modelnet40\cite{finn2017model} weights of each architecture for $k$-way, $m$-shot few shot learning \cite{ye2023closer}. We observe that PointMLP\cite{ma2022rethinking} performs best on ScanObjectNN\cite{uy2019revisiting} while DCGNN\cite{wang2019dynamic} output performs all other methods on the proposed 3DGrocery63 dataset.

For the meta-learning Task, we follow the settings proposed by ProtoNet~\cite{snell2017prototypical} and construct settings for the point cloud meta-learning task. As explained in Section~\ref{sec:FSL}, we propose 3DGrocery63 as a strong generalization dataset. Table \ref{table4} demonstrates that almost all architectures perform well on weak generalization tasks, but the performance is drastically dropped on proposed strong generalization tasks. This indicates that meta-learns are not robust to data-inductive bias, and a need exists to address this problem by solving real-world scenarios. Meanwhile, Table~\ref{table14} depicts the individual weak generalization results on ScanOBjectNN~\cite{uy2019revisiting}.

\textbf{Note:} We sample 4 episodes of few-shot data for training and 200 validation episodes for 50 epochs. The best-performing weights in the validation set are used to test with 1000 episodes on few-shot data for both weak and strong generalization tasks.

\begin{table}[!h]\centering
  \caption{Quantitative analysis for Few-shot 3D point cloud classification evaluation on ScanObjectNN~\cite{uy2019revisiting} dataset. \textbf{Note:} all models are pre-trained on ModelNet40~\cite{wu20153d}}
  \begin{adjustbox}{width=\linewidth}
  \begin{tabular}{rc|cccc|cccc}
    \hline
   \multirow{2}{*}{\textbf{Model}}&\multirow{2}{*}{\textbf{Weight Init}}&\multicolumn{4}{c|}{\textbf{5-ways}}&\multicolumn{4}{c}{\textbf{10-ways}}\\
   \cline{3-10} 
    &&\multicolumn{2}{c}{\textbf{10-shots}}&\multicolumn{2}{c|}{\textbf{20-shots}}&\multicolumn{2}{c}{\textbf{10-shots}}&\multicolumn{2}{c}{\textbf{20-shots}}\\
   \hline 
    
    \multirow{2}{*}{\textbf{DGCNN}\cite{wang2019dynamic}}&
    Random&67.60&$\pm$8.35&74.46&$\pm$7.38&51.88&$\pm$4.75&61.89&$\pm$5.49\\
    &MN40&79.86&$\pm$6.67&85.10&$\pm$5.76&67.12&$\pm$4.89&76.09&$\pm$4.62\\
    \multirow{2}{*}{\textbf{PointMLP}\cite{ma2022rethinking}}&
    Random&33.20&$\pm$6.11&37.86&$\pm$7.63&20.57&$\pm$3.28&24.41&$\pm$4.25\\
    &MN40&\textbf{81.60}&\textbf{$\pm$6.03}&\textbf{85.78}&\textbf{$\pm$6.38}&\textbf{70.58}&\textbf{$\pm$4.31}&\textbf{76.86}&\textbf{$\pm$3.96}\\
    \multirow{2}{*}{\textbf{PointNet++}\cite{qi2017pointnet++}}&
    Random&67.68&$\pm$8.92&72.98&$\pm$7.46&51.15&$\pm$5.39&58.61&$\pm$4.65\\
    &MN40&77.02&$\pm$6.98&80.86&$\pm$5.81&62.62&$\pm$5.61&69.99&$\pm$4.41\\
    \multirow{2}{*}{\textbf{PointNet}\cite{qi2017pointnet}}&
    Random&69.78&$\pm$8.56&74.76&$\pm$7.65&53.85&$\pm$5.16&61.19&$\pm$5.44\\
    &MN40&76.56&$\pm$8.26&80.34&$\pm$7.48&61.57&$\pm$4.82&70.30&$\pm$5.29\\

    \multirow{2}{*}{\textbf{PCT}\cite{guo2021pct}}&Random&67.73&$\pm$7.72&73.83&$\pm$7.33&56.59&$\pm$4.27&60.02&$\pm$7.55\\
    &MN40&72.76&$\pm$8.73&79.54&$\pm$6.95&61.86&$\pm$3.49&70.14&$\pm$6.03\\

    \multirow{2}{*}{\textbf{PointNeXT}\cite{PointNeXt}}&Random&69.03&$\pm$8.66&74.52&$\pm$7.40&53.33&$\pm$4.70&62.03&$\pm$4.58\\
    &MN40&79.02&$\pm$6.22&83.94&$\pm$5.69&65.64&$\pm$4.01&74.12&$\pm$4.24\\
    
    
    \hline
  \end{tabular}
  \end{adjustbox}
  \label{table13}
\end{table}

\begin{table}[!h]\centering
  \caption{Quantitative analysis for Few-shot 3D point cloud classification on ScanObjectNN\cite{uy2019revisiting} for Weak generalizations evaulation. \textbf{Note:} W1 - Weak 1 (ONLY OBJ split), W2 - Weak 2 (OBJ + BG split), and W3 - Weak 3 (PB75 split).}
  \begin{adjustbox}{width=\linewidth}

  \begin{tabular}{r|c|cccc|cccc}
    \hline
   \multirow{2}{*}{\textbf{Model}}&\multirow{2}{*}{\textbf{Split}}&\multicolumn{4}{c|}{\textbf{5-ways}}&\multicolumn{4}{c}{\textbf{10-ways}}\\
   \cline{3-10} 
    &&\multicolumn{2}{c}{\textbf{5-shots}}&\multicolumn{2}{c|}{\textbf{10-shots}}&\multicolumn{2}{c}{\textbf{5-shots}}&\multicolumn{2}{c}{\textbf{10-shots}}\\
   \hline 
    
   \multirow{3}{*}{\textbf{PointNet}~\cite{qi2017pointnet}}
   &W1&56.32&$\pm$0.64&59.91&$\pm$0.61&39.76&$\pm$0.38&44.08&$\pm$0.36\\

   &W2&54.88&$\pm$0.65&57.44&$\pm$0.61&37.03&$\pm$0.37&41.04&$\pm$0.37\\

   &W3&49.29&$\pm$0.59&51.73&$\pm$0.57&32.40&$\pm$0.36&35.20&$\pm$0.36\\
   \hline
   \multirow{3}{*}{\textbf{PointNet++}~\cite{qi2017pointnet++}}
   &W1&51.35&$\pm$0.61&58.19&$\pm$0.64&38.74&$\pm$0.38&42.54&$\pm$0.36\\

   &W2&52.04&$\pm$0.60&58.13&$\pm$0.63&39.04&$\pm$0.35&41.16&$\pm$0.36\\
   &W3&47.09&$\pm$0.60&51.44&$\pm$0.60&33.30&$\pm$0.35&35.48&$\pm$0.35\\
   
   \hline
   \multirow{3}{*}{\textbf{DGCNN}~\cite{wang2019dynamic}}
   &W1&46.43&$\pm$0.56&52.97&$\pm$0.59&33.24&$\pm$0.34&38.49&$\pm$0.35\\

   &W2&56.47&$\pm$0.65&\textbf{61.26}&$\pm$\textbf{0.64}&41.64&$\pm$0.41&43.87&$\pm$0.38\\

   &W3&51.00&$\pm$0.65&55.14&$\pm$0.61&35.63&$\pm$0.36&37.89&$\pm$0.36\\

   \hline
   \multirow{3}{*}{\textbf{PointMLP}~\cite{ma2022rethinking}}
   &W1&47.67&$\pm$0.63&52.04&$\pm$0.57&34.65&$\pm$0.35&43.75&$\pm$0.38\\

   &W2&48.10&$\pm$0.63&52.94&$\pm$0.60&34.65&$\pm$0.37&43.04&$\pm$0.38\\
   &W3&42.61&$\pm$0.61&45.44&$\pm$0.60&28.98&$\pm$0.32&34.55&$\pm$0.35\\
   
   \hline
   \multirow{3}{*}{\textbf{PointTransformer}~\cite{guo2021pct}}
   &W1&57.94&$\pm$0.64&57.07&$\pm$0.66&42.51&$\pm$0.38&44.77&$\pm$0.37\\

   &W2&59.70&$\pm$0.65&59.48&$\pm$0.66&41.62&$\pm$0.38&\textbf{46.31}&$\pm$\textbf{0.37}\\

   &W3&53.52&$\pm$0.64&53.20&$\pm$0.60&34.86&$\pm$0.36&38.64&$\pm$0.36\\

    \hline
   \multirow{3}{*}{\textbf{PointNeXT}~\cite{PointNeXt}}
   &W1&\textbf{60.17}&$\pm$\textbf{0.63}&\textbf{61.21}&$\pm$\textbf{0.63}&\textbf{43.32}&$\pm$\textbf{0.39}&\textbf{47.68}&$\pm$\textbf{0.37}\\
   &W2&\textbf{60.88}&$\pm$\textbf{0.64}&59.92&$\pm$0.67&\textbf{43.88}&$\pm$\textbf{0.40}&46.17&$\pm$0.38\\
   &W3&\textbf{54.84}&$\pm$\textbf{0.64}&\textbf{55.69}&$\pm$\textbf{0.65}&\textbf{37.90}&$\pm$\textbf{0.38}&\textbf{40.73}&$\pm$\textbf{0.38}\\

   
    \hline
  \end{tabular}
  \end{adjustbox}
  \label{table14}
\end{table}

\end{document}